\documentclass[11pt, oneside]{article}   	
\usepackage{geometry}                		
\usepackage{graphicx}				
\usepackage{amsmath,amssymb}

\usepackage{bm}

\usepackage{hyperref}

\usepackage{xcolor}
\usepackage{synttree}
\usepackage{tikz}
\usepackage{subcaption}
\usepackage{algorithm}
\usepackage{algorithmic}
\usepackage{multirow}
\usepackage{hhline}
\usepackage{mathtools}
\usepackage{booktabs}
\usepackage{stackengine}
\stackMath
\DeclareMathOperator*{\argmax}{argmax}
\DeclareMathOperator*{\bigtimes}{\times}

\usepackage{setspace}
\usepackage{etoolbox}
\AtBeginEnvironment{algorithmic}{\setstretch{1.25}}

\usepackage[normalem]{ulem}

\newcommand{\Hc}{\mathcal{H}}
\newcommand{\Lc}{\mathcal{L}}
\newcommand{\Mc}{\mathcal{M}}
\newcommand{\Nc}{\mathcal{N}}
\newcommand{\Pc}{\mathcal{P}}
\newcommand{\Rc}{\mathcal{R}}
\newcommand{\Tc}{\mathcal{T}}

\newcommand{\Yc}{\mathcal{Y}}
\newcommand{\Xc}{\mathcal{X}}
\newcommand{\Rbb}{\mathbb{R}}
\newcommand{\Ebb}{\mathbb{E}}
\newcommand{\Nbb}{\mathbb{N}}
\newcommand{\Pbb}{\mathbb{P}}

\newcommand{\rank}{\mathrm{rank}}
\DeclareMathOperator{\depth}{depth}
\DeclareMathOperator{\level}{level}

\newtheorem{theorem}{Theorem}[section]
\newtheorem{remark}[theorem]{Remark}
\newtheorem{example}[theorem]{Example}

\usepackage{colortbl}
\definecolor{Gray}{gray}{0.90}

\tikzstyle{active}=[circle,fill=black]

\title{Learning with tree-based tensor formats}
\author{Erwan Grelier\thanks{Centrale Nantes, Laboratoire de Math\'ematiques Jean Leray, CNRS UMR 6629, 
        Joint Laboratory of Marine Technology between Naval Group and Centrale Nantes} \and Anthony Nouy\thanks{Centrale Nantes,
        Laboratoire de Math\'ematiques Jean Leray, CNRS UMR 6629}\;\thanks{Corresponding author, anthony.nouy@ec-nantes.fr} \and Mathilde Chevreuil\thanks{Universit\'e de Nantes, GeM, CNRS UMR 6183}} 
\date{}							

\begin{document}
\maketitle
\begin{abstract}
This paper is concerned with the approximation of high-dimensional functions in a statistical learning setting, using model classes of functions in tree-based tensor format. These are particular classes of rank-structured functions that 
admit explicit and numerically stable representations, parametrized by a tree-structured network of low-order tensors. These nonlinear model classes can be seen as deep neural networks with a sparse architecture related to the tree and non-standard multilinear activation functions.
This paper provides adaptive algorithms for learning with such model classes by empirical risk minimization. The proposed algorithm for learning in a given model class exploits the fact that tree-based tensor formats are multilinear models and recasts the problem of risk minimization over a nonlinear set into a succession of learning problems with linear models. Suitable changes of representation yield numerically stable learning problems and allow to exploit sparsity in the parameters using standard regularization or greedy-type approaches. 
For high-dimensional problems or when only a small data set is available, the selection of a good model class is a critical issue. For a given tree, 
the selection of the tuple of tree-based ranks that minimize the risk is a combinatorial problem. Here, we propose a rank adaptation strategy  which provides in practice a good convergence of the risk as a function of the model class complexity.  
Finding a good tree is also a combinatorial problem, which can be related to the choice of a particular sparse architecture for deep neural networks. 
Here, we propose a stochastic algorithm for minimizing the complexity of the representation of a given function over a class of trees with a given arity, allowing changes in the topology of the tree.
This tree optimization algorithm is then included in a learning scheme that successively adapts the tree and  the corresponding tree-based ranks. 
Contrary to classical learning algorithms for nonlinear model classes, the proposed algorithms are numerically stable, reliable, and require only a low level expertise of the user. Their performances are illustrated with numerical experiments in a supervised learning setting.  
\end{abstract}


\section{Introduction}

The approximation of high-dimensional functions is a typical task in statistics and machine learning. This includes supervised learning problems, where one is interested in the approximation of a random variable $Y$ as a function of many random variables $X = (X_1,\hdots,X_d)$ from samples of $ (X,Y)$. This also includes the estimation of the probability distribution of a high-dimensional random vector $X = (X_1,\hdots,X_d)$ from samples of $X$.
A typical path for solving such problems consists in finding an approximation $v$ that minimizes a risk 
functional 
$$
\Rc(v) = \Ebb(\gamma(v,Z)),
$$ 
with $Z=(X,Y)$ for supervised learning or $Z=X$ for density estimation, and 
where $\gamma$ is a contrast (or loss) function such that $\gamma(v,z)$ is a measure of the error due to the use of the approximation $v$ for a sample $z$ of $Z$. For density estimation, the contrast is chosen such that the risk corresponds to some distance from $v$ to the density of $X$. For supervised learning, the contrast is defined by $\gamma(v,(x,y)) = \ell(y,v(x))$ with $\ell$ a loss function that measures some distance between the observation $y$ and the prediction $v(x)$, for a given sample $(x,y)$ of $(X,Y)$. The function $u$ that minimizes the risk over the set of measurable functions (when it exists) is called the target or oracle function.
In practice, given a training set of samples $\{z_i\}_{i=1}^n$ of $Z$, an approximation $\hat u_M^n$ is obtained by minimizing an empirical risk
$$
\Rc_n(v) = \frac{1}{n} \sum_{i=1}^n \gamma(v,z_i)
$$ 
over a subset of functions $M$ called a model class (or hypothesis set).   
The error (or excess risk) $\Rc(\hat u_M^n) - \Rc(u)$ is the sum of an estimation error 
$ \Rc(\hat u_M^n) -  \Rc( u_M) $ and of an approximation error $\Rc(u_M) - \Rc(u)$. Choosing larger and larger model classes $M$ naturally makes the approximation error decrease but it also makes the variance of the estimation error increase. 
Given a training set, a suitable model class has to be selected in order to make these two errors as low as possible. 

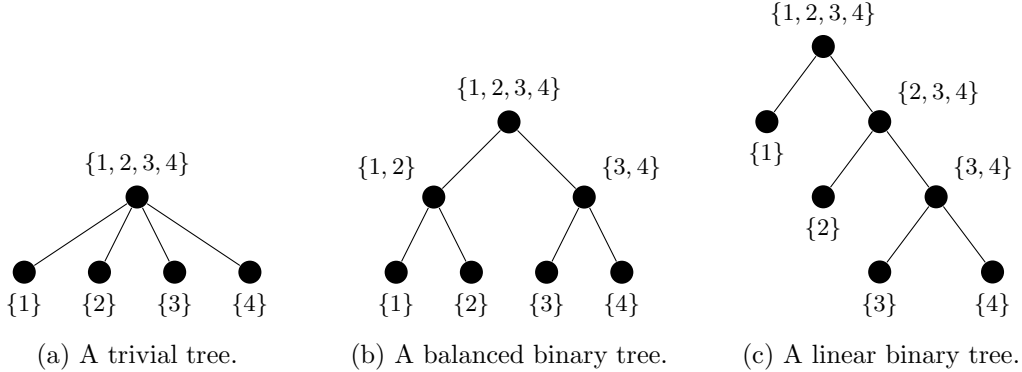
\begin{figure}[h]
		\footnotesize
		\centering
		\begin{subfigure}[b]{.32\textwidth}
			\centering
			\begin{tikzpicture}[level distance = 10mm,sibling distance=10mm]
				\tikzstyle{active}=[circle,fill=black]
				\node [active,label={above:{$\{1,2,3,4\}$}}]  {}
				         child {node [active,label=below:{$\{1\}$}]  {}}
					 child {node [active,label=below:{$\{2\}$}]  {}}
					 child {node [active,label=below:{$\{3\}$}]  {}}
					 child {node [active,label=below:{$\{4\}$}]  {}}
					 ;
			\end{tikzpicture}
			\caption{A trivial tree.}
			\label{fig:tree-trivial}
		\end{subfigure}
		\begin{subfigure}[b]{.32\textwidth}
			\centering
			\begin{tikzpicture}[level distance = 10mm,sibling distance=10mm]
				\tikzstyle{active}=[circle,fill=black]
				\tikzstyle{level 1}=[sibling distance=20mm]
				\tikzstyle{level 2}=[sibling distance=10mm]
				\node [active,label={above:{$\{1,2,3,4\}$}}]  {}
					child {node [active,label=above left:{$\{1,2\}$}]  {}
						child {node [active,label=below:{$\{1\}$}] {}}
						child {node [active,label=below:{$\{2\}$}] {}}
						}
					child {node [active,label=above right:{$\{3,4\}$}] {}
						child {node [active,label=below:{$\{3\}$}] {}}
						child {node [active,label=below:{$\{4\}$}] {}}
						};
			\end{tikzpicture}
			\caption{A balanced binary tree.}
			\label{fig:tree-balanced}
		\end{subfigure}
		\begin{subfigure}[b]{.32\textwidth}
			\centering
			\begin{tikzpicture}[level distance = 10mm,sibling distance=10mm]
				\tikzstyle{active}=[circle,fill=black]
				\tikzstyle{level 1}=[sibling distance=15mm]
				\tikzstyle{level 2}=[sibling distance=15mm]
				\tikzstyle{level 3}=[sibling distance=15mm]
				\node [active,label={above:{$\{1,2,3,4\}$}}]  {}
					child {node [active,label=below:{$\{1\}$}]  {}
					}
					child {node [active,label=above right:{$\{2,3,4\}$}] {}
						child {node [active,label=below:{$\{2\}$}] {}}
						child {node [active,label=above right:{$\{3,4\}$}] {}
							child {node [active,label=below:{$\{3\}$}] {}}
							child {node [active,label=below:{$\{4\}$}] {}}
							}
					};
			\end{tikzpicture}
			\caption{A linear binary tree.}
			\label{fig:tree-linear}
		\end{subfigure}
		\caption{Examples of dimension partition trees over $D=\{1,\hdots,4\}$.}.
		\label{fig:trees}
	\end{figure}

For high dimensional problems (large $d$) or when only a limited number of samples are available (small $n$), the proposed model classes have to exploit low-dimensional structures of the target function $u$. Typical model classes for high dimensional approximation include 
\begin{itemize}
\item additive models $ v_1(x_1) + \hdots + v_d(x_d)$, or more general models with low interactions $ \sum_{\alpha \in T} v_\alpha(x_\alpha)$, where $T$ is a  collection of small subsets $\alpha$ of $\{1,\hdots,d\}$ and the $v_\alpha$ are functions of groups of variables,
\item expansions $\sum_{\lambda \in \Lambda} c_\lambda \psi_\lambda(x)$ on a set of functions $\{\psi_\lambda\}_{\lambda \in \Lambda}$, possibly picked in a larger basis or dictionary of functions (in which a sparse approximation is searched for),
\item multiplicative models $v_1(x_1)  \hdots v_d(x_d)$ or a sum of such models  $ \sum_{i=1}^n v_1^i(x_1)  \hdots v_d^i(x_d) $,
\item ridge models $f(Ax)$, where $A$ is a linear map from $ \Rbb^d$ to $\Rbb^m$  and where $f$ belongs to a  model class of functions of $m$ variables, 
\item projection pursuit models $f_1(w_1^Tx) + \hdots + f_m(w_m^Tx) $ where $w_i \in \Rbb^d$ (which is a ridge model with a function $f$  taken in the class of additive models), a particular case being neural networks $c_1 \sigma(w_1^Tx + b_1) + \hdots + c_m \sigma(w_m^Tx + b_m) $, where $\sigma$ is a given (so called activation) function,
\item more general compositions of functions $f\circ g_1 \circ \hdots \circ g_L(x)$, where the $g_i$ are vector-valued functions whose components are taken in some standard model classes, a particular case being  deep neural networks for which the $g_i$ are ridge functions of the form $g_i(t) = \sigma(A_i t + b_i)$ (with $A_i$ possibly sparse, such as for convolutional or recurrent networks).
\end{itemize}
In this paper, we consider model classes of rank-structured functions. The sum of multiplicative models is a particular case of such classes associated with the canonical notion of rank. Other notions of rank 
lead to classes of functions that are more amenable for numerical computations and approximation tasks. 
For any subset $\alpha$ of $D := \{1,\hdots,d\}$, 
a natural notion of $\alpha$-rank of a function $v(x)$, denoted $\rank_\alpha(v)$, can be defined as the minimal integer
 $r_\alpha$ such that 
$$
v(x) = \sum_{k=1}^{r_\alpha} v^\alpha_k(x_\alpha) v_k^{\alpha^c}(x_{\alpha^c})
$$
for some functions $v^\alpha_k$ and $v_k^{\alpha^c}$ of two complementary groups of variables.
By considering a collection $T$ of subsets of $D$ and a tuple ${{r}} = (r_\alpha)_{\alpha \in T}$, a model class $\Tc^T_{{r}}$ of rank-structured functions can then be defined as 
$$
\Tc^T_{{r}} = \{v \in \Hc : \rank_{\alpha}(v) \le r_\alpha , \alpha \in T\},
$$
where $\Hc$ is some (Banach) tensor space of functions (e.g., $L^2_\mu(\Rbb^d)$ with $\mu$ a product probability measure). 
This paper is concerned with the particular case where 
 $T$ is a \emph{dimension partition tree} over $D$, which corresponds to the model class of 
functions in \emph{tree-based tensor format} \cite{Falco2018SEMA}, a particular class of tensor networks \cite{cichocki2016tensor}. This includes the Tucker format for a trivial tree (Figure~\ref{fig:tree-trivial}), the hierarchical tensor format \cite{hackbusch2009newscheme} for a balanced binary tree (Figure~\ref{fig:tree-balanced}), and the tensor train format \cite{oseledets2009breaking} for a linear tree (Figure~\ref{fig:tree-linear}). 
When  $T$ is a dimension partition tree, the set $\Tc^T_r$ possesses nice topological and geometrical properties \cite{Falco:2012uq,Falco2018SEMA,2018arXiv180911001A,uschmajew2013geometry,2015arXiv150503027F,Falco2018,Holtz:2012fk} and elements of $\Tc^T_{{r}}$ admit explicit and numerically stable representations, that will be exploited in this paper. These representations are parametrized by a tree-structured network of low-order tensors. 
The resulting complexity (i.e., the number of parameters) of the model class is linear in $d$ and polynomial in the ranks, with a polynomial degree depending on the arity of the tree.

Model classes of functions in tree-based tensor format have a high expressive power.  
To some extent, these classes are related to the above listed model classes. 
In particular, additive models have $\alpha$-ranks bounded by $2$ for any $\alpha$, so that they are included in $\Tc^T_{r}$ for any tree $T$ and ${r} \ge 2$.
The sum of $n$ multiplicative models is included in $\Tc^T_{r}$ for any tree $T$ and ${r} \ge n$. Also, sparse approximations on tensor product bases or ridge models can be proved to admit accurate representations in tree-based tensor formats with moderate complexity, under some regularity assumptions 
\cite{hackbusch2012book}.	\par
The model class $\Tc^T_{r}$ can be interpreted as a class of functions that are compositions of multilinear functions, the structure of compositions being given by the tree. For example, with the tree of  Figure \ref{fig:tree-balanced}, a function $v\in \Tc^T_r$ 
has a representation of the form
\begin{equation}
	v(x) = f^{D}(f^{\{1,2\}}(f^{\{1\}}(\bm \Phi^1(x_1)),f^{\{2\}}(\bm \Phi^2(x_2))),f^{\{3,4\}}(f^{\{3\}}(\bm \Phi^3(x_3)),f^{\{4\}}(\bm \Phi^4(x_4)))), \label{compo-balanced}
\end{equation}
where the $f^\alpha$ are $\Rbb^{r_\alpha}$-valued multilinear functions, and $\bm \Phi^\nu(x_\nu)$ is a set of variables associated with a set of functions of the variable $x_\nu$\footnote{$\bm \Phi^\nu(x_\nu) $ could be a classical approximation basis (e.g., polynomials or wavelets) or a feature map associated with a reproducing kernel Hilbert space of functions of $x_\nu$.}. This yields an interpretation of tree-based tensor formats as a particular class of deep neural networks \cite{cohen2016expressive,khrulkov2018expressive}, with multilinear activation functions, a sparse architecture determined by the tree $T$, a depth corresponding to the depth of the tree $T$, and a width at a certain level related to the ranks $r_\alpha$ of the nodes $\alpha$ of that level. 
In this context, a balanced tree is related to convolutional networks, while a linear tree is related to recurrent networks.	
\\\par

The aim of this paper is to provide 
robust algorithms for learning with model classes 
of functions in tree-based tensor format. Contrary to classical learning algorithms for nonlinear model classes, the proposed algorithms are numerically stable, reliable, and require only a low level expertise of the user.
The proposed algorithm for learning in a given model class $\Tc^T_r$ exploits the fact that tree-based tensor formats are multilinear models. Alternating minimization (or block descent) algorithms therefore recast the risk minimization over a nonlinear set  into a succession of learning problems with linear models. For each of these problems,  a suitable change of representation (with orthogonality conditions) yields a numerically stable learning problem, with improved statistical properties. Furthermore, when based on higher-order singular value decompositions, this change of representation leads to a linear model with a hierarchical basis, which allows us to exploit sparsity in the parameters (e.g., using sparsity inducing regularizations or adaptive greedy-type algorithms).
  
The expressive power of tree-based tensor formats is related to the fact that
for any tree $T$, the union of sets $\Tc_{{r}}^T$, ${{r}} \in \Nbb^{\# T}$, is dense in the tensor space $\Hc$, so that any function in $\Hc$ can be approximated at an arbitrary small precision by a sequence of approximations with increasing ranks. However, the selection of an optimal sequence of ranks, i.e., yielding an optimal convergence of the error with respect to the complexity of the model class, is a nontrivial problem. In this paper, we also provide a strategy for rank adaptation, which consists in increasing the $\alpha$-ranks associated with the highest 
truncation errors $$\min_{\rank_\alpha(v) \le r_\alpha} \Rc(v) - \Rc(u),$$
which are in practice estimated by replacing the unknown oracle function $u$ by an 
approximation obtained by a correction of the current approximation with $\alpha$-ranks $r_\alpha$.

Although tree-based tensor formats have a high expressive power for a given tree, the choice of a good tree is crucial in practice. Indeed, the ranks required for reaching a certain error $\epsilon$ may strongly depend on the choice of the tree. More precisely, the required ranks ${r} $ may grow as $\epsilon^{-\gamma}$ with $\gamma$ strongly depending on the tree.  For high-dimensional problems and/or when the sample size is small, finding a good tree therefore becomes a critical issue. This issue has to be related to the choice of a particular sparse architecture for deep neural networks (e.g., convolutional or recurrent networks, with an ordering of the variables which is application-dependent). 
It is known that a function in tree-based tensor format with a linear binary tree and ranks $r = O(R)$ can be represented with a balanced binary tree and ranks  $r=O(R^2)$, while a function in tree-based tensor format with a balanced binary tree and ranks $r=O(R)$ can be represented with a linear binary tree and ranks $r=O(R^{\log_2(d)})$. Also, as will be illustrated in this paper, different trees with the same topology but different ordering of the variables may  yield very different complexities, sometimes dramatically depending on $d$. Therefore, for a given sample, the use of different trees   may yield very different performances in a statistical learning setting. 
Finding an optimal dimension partition tree is a combinatorial problem which is intractable in practice. 
In this paper, we propose a heuristic stochastic algorithm that explores the set of trees with a given arity (allowing changes in the topology of the tree) in order to minimize the complexity of the representation of a given function. It consists in comparing a current representation with a representation associated with another tree,  drawn randomly from a suitable probability distribution over the set of trees with a fixed arity, this distribution taking into account the computational complexity for changing the representation and the potential complexity reduction of the function's representation. This tree optimization procedure is then included in a learning algorithm that successively modifies the tree and adapts the corresponding ranks. 

The outline of the paper is as follows. Section \ref{sec:treeBasedFormats} introduces the model class of functions in tree-based tensor format and describes the key ingredients for the development of learning algorithms (representations with orthogonality conditions, hierarchical representations for sparse approximation, truncations).
Section \ref{sec:learning} presents a learning algorithm to compute an approximation of a function in tree-based tensor format, possibly exploiting sparsity in the parameters, and discusses the problems of validation and model selection.
Section \ref{sec:adaptiveAlgorithms} presents rank and tree adaptation strategies. 
Finally, Section \ref{sec:experiments} presents numerical experiments that demonstrate the performance of the proposed algorithms in a supervised learning setting. 

\section{Tree-based tensor formats}\label{sec:treeBasedFormats}

In this section, we introduce the model class of functions in tree-based tensor format in a typical learning setting. For an introduction to tree-based tensor formats and their applications in other contexts, the reader is referred to the monograph \cite{hackbusch2012book} and recent surveys \cite{KOL09,Khoromskij:2012fk,Grasedyck:2013,nouy:2016_handbook,nouy:2017_morbook,bachmayr2016tensor}.  

\subsection{Tensor spaces of multivariate functions}\label{sec:tensors}
Let $(X_1,\hdots,X_d)$ be a set of independent random variables, where $X_\nu$ is with values in $\Xc_\nu$ and with probability law $\mu_\nu$, $1\le \nu \le d$. Typically, $\Xc_\nu$ is a subset of $ \Rbb$ but the case where $\Xc_\nu \subset \Rbb^{d_\nu}$, $d_\nu>1$, can be considered as well. 
We denote by $X = (X_1,\hdots,X_d)$ the random variable with values in $\Xc = \Xc_1\times \hdots \times \Xc_d$ and with probability law    $\mu = \mu_1\otimes \hdots \otimes \mu_d$, and by $\Ebb(\cdot)$ the mathematical expectation. 

	Let $\Hc_\nu$ be a Hilbert space of functions defined 
	on $\Xc_\nu$, $1\le \nu\le d$, equipped with the inner product $(\cdot,\cdot)_\nu$ and associated norm $\Vert \cdot\Vert_\nu$. 
	The elementary tensor product $u^1\otimes \hdots \otimes u^d$ of functions $u^\nu \in \Hc_\nu$, $1\le \nu\le d$, is identified with a function defined on $\Xc$ such that for $x=(x_1,\hdots,x_d)\in \Xc$,
	$
	(u^1\otimes \hdots \otimes u^d)(x_1,\hdots,x_d) = u^1(x_1)\hdots u^d(x_d).
	$
	The algebraic tensor space $ \Hc =  \Hc_1\otimes \hdots \otimes \Hc_d$  is then identified with the set of functions $u$ which can be written as a finite linear combination of elementary tensors.
	The tensor space $\Hc$ is equipped with the canonical inner product 
	$(\cdot,\cdot)$, first defined for elementary tensors by 
	$(u^1\otimes \hdots \otimes u^d,v^1\otimes \hdots \otimes v^d) = (u^1,v^1)_1\hdots (u^d,v^d)_d,$
	and then extended by linearity to the whole space $\Hc$. We denote by $\Vert \cdot\Vert $ the canonical norm associated with $(\cdot,\cdot)$. 
	\begin{remark}
		If the  $\Hc_\nu$ are infinite dimensional spaces, the tensor space $\Hc$ is a pre-Hilbert space.
		A tensor Hilbert space $\Hc_{\Vert\cdot\Vert}$ is  obtained by the completion of $\Hc$ (in the topology induced by the norm $\Vert\cdot\Vert$). 
		In particular, $L^2_\mu(\Xc)$ can be identified with the completion of the algebraic tensor space $ L^2_{\mu_1}(\Xc_1) \otimes \hdots \otimes L^2_{\mu_d}(\Xc_d)$.
	\end{remark}
	Hereafter, we consider that $\Hc_\nu$ is a finite dimensional subspace of $L^2_{\mu_\nu}(\Xc_\nu)$, equipped with the  norm $\Vert u^\nu \Vert_{\nu}^2 = {\Ebb(u^\nu(X_\nu)^2)}$. Then, $\Hc$ is a subspace of $L^2_{\mu}(\Xc)$, equipped with the canonical norm  $\Vert u \Vert^2 = {\Ebb(u(X)^2)}$. 
	Let $\{\phi^\nu_{i} : i\in I^\nu \}$ be an orthonormal basis of $\Hc_\nu$, and $N_\nu = \dim(\Hc_\nu) = \#I^\nu$. 
	For a multi-index $i = (i_1,\hdots,i_d)  \in   I^1\times \hdots \times I^d := I$, we let  
	$\phi_i = \phi^1_{i_1} \otimes \hdots \otimes \phi^d_{i_d} $. The set of functions $\{\phi_i : i\in I\}$ constitutes an orthonormal basis of $\Hc$. A function $u \in \Hc$ can be written 
	$
		u(x) = \sum_{i \in I} u_i \phi_i(x),
	$
	where the set of coefficients $  (u_i)_{i\in I} \in \Rbb^{I}$ is identified with a tensor $\bm{u}  \in \Rbb^{I^1} \otimes \hdots \otimes \Rbb^{I^d}$, and 
	$$
	\Vert u \Vert^2= {\sum_{i\in I} u_{i}^2} = {\sum_{i_1\in I^1} \hdots \sum_{i_d \in I^d} u_{i_1,\hdots,i_d}^2}
	$$ 
	coincides with the \emph{canonical norm} of $ \bm{u}$, also denoted $\Vert \bm{u} \Vert$. Denoting $$\bm\Phi^\nu(x_\nu) = (\phi_i^{\nu}(x_\nu))_{i \in I^\nu} \in \Rbb^{I^\nu}$$ and $$\bm\Phi(x) = \bm\Phi^1(x_1)\otimes \hdots \otimes \bm\Phi^d(x_d) \in \Rbb^{I^1} \otimes \hdots \otimes \Rbb^{I^d},$$ we have 
	$
		u(x) = \langle \bm\Phi(x)  , \bm{u}\rangle,
	$
	where $\langle \cdot,\cdot \rangle$ is the canonical inner product in $\Rbb^{I^1}\otimes \hdots \otimes \Rbb^{I^d}$. The linear map  $$F : \bm{u} \mapsto \langle\bm\Phi(\cdot),\bm{u}\rangle $$ defines a linear isometry 
	from $\Rbb^{I^1}\otimes \hdots \otimes \Rbb^{I^d}$ to $\Hc$, such that 
	$\Vert F(\bm{u}) \Vert = \Vert\bm{u} \Vert$.
	
	The \emph{canonical rank} of a tensor $u \in \Hc$ is the minimal integer $r$ such that $u$ can be written in the form 
	\begin{align*}
		u(x_1,\hdots,x_d) = \sum_{i=1}^r v_i^1(x_1)\hdots v_i^d(x_d),
	\end{align*} 
	for some $v_i^\nu \in \Hc_\nu$ and $r\in \Nbb$. 
	The set of tensors in $\Hc$ with canonical rank bounded by $r$ is denoted $\Rc_r(\Hc)$. An approximation in $\Rc_r(\Hc)$ is called an approximation in \emph{canonical tensor format}.
	For an order-two tensor ($d=2$), the canonical rank coincides with the classical and unique notion of rank.
	For higher-order tensors ($d>3$),  different notions of rank can be introduced. 

\subsection{Tree-based ranks and corresponding tree-based formats}\label{sec:ranks}
	For a non-empty subset $\alpha$ in $\{1,\hdots,d\}:= D$ and its complementary subset $\alpha^c = D\setminus \alpha$, a tensor $u\in \Hc$ can be identified with an element $\Mc_\alpha(u)$ of the space  of order-two tensors $\Hc_\alpha \otimes \Hc_{\alpha^c}$, where $\Hc_\alpha = \bigotimes_{\nu\in \alpha} \Hc_\nu$. This is equivalent to identifying $u(x)$ with a bivariate function of the complementary groups of variables $x_\alpha = (x_\nu)_{\nu\in \alpha}$ and $x_{\alpha^c} = (x_\nu)_{\nu\in \alpha^c}$ in $x$. The operator $\Mc_\alpha$ is called the $\alpha$-matricization operator. 
The $\alpha$-rank of $u$, denoted $\rank_\alpha(u)$, is the dimension of the minimal subspace $U^{min}_\alpha(u)$, which is the smallest subspace of functions of the variables $x_\alpha$ such that $\Mc_\alpha(u)\in U^{min}_\alpha(v) \otimes \Hc_{\alpha^c}$. By convention, $U^{min}_D(u)=\mathrm{span}\{u\}$ and $\rank_D(u)=1$ if $u\neq 0$ and $0$ if $u=0$. If $\rank_\alpha(u) = r_\alpha$, then $u$ admits the following 
 representation 
	\begin{equation}
	u(x) = \sum_{i=1}^{r_\alpha} u^\alpha_i(x_\alpha) u^{\alpha^c}_i(x_{\alpha^c}), \label{repres-alpha-rank}
	\end{equation}
	for some functions $u^\alpha_i \in \Hc_\alpha$ and $u^{\alpha^c}_i \in \Hc_{\alpha^c}$, and $U^{min}_\alpha(u) = \mathrm{span}\{u_i^\alpha\}_{i=1}^{r_\alpha}$. 
	The set of tensors $u$ in $\Hc$ with $\alpha$-rank bounded by $r_\alpha$ is denoted by $$\Tc_{r_\alpha}^{\{\alpha\}}(\Hc)=\left\{u \in \Hc : \rank_\alpha(u) \le r_\alpha\right\}.$$
	For a collection $T$ of non-empty subsets of $D$, we define the $T$-rank of $u$ as the tuple $ \rank_T(u)=\{\rank_\alpha(u) : \alpha \in T\}$. Then, we define the set of tensors $\Tc_r^T(\Hc)$ with $T$-rank bounded by $r=(r_\alpha)_{\alpha \in T}$ by
	$$
	\Tc_r^T(\Hc) = \left\{u \in \Hc : \rank_T(u) \le r \right\} = \bigcap_{\alpha\in T} \Tc_r^{\{\alpha\}}(\Hc)  .
	$$
	A \emph{dimension partition tree} $T$ is a tree such that 
	(i) all nodes $\alpha \in T$ are non-empty subsets of $D$, (ii) $D$ is the root of $T$, (iii) every node $\alpha \in T$ with $\#\alpha \ge 2$ has at least two children and the set of children of $\alpha$, denoted by $S(\alpha)$, is a non-trivial partition of $\alpha$, and (iv) every node $\alpha$ with $\#\alpha = 1$ has no child and is called a leaf (see examples on Figure \ref{fig:trees}).

	When $T$ is a dimension partition tree, $\Tc_r^T(\Hc)$ is the set of tensors with \emph{tree-based rank} bounded by $r$, and an approximation in $\Tc_r^T(\Hc)$ is called an approximation in \emph{tree-based (or hierarchical) tensor format} \cite{hackbusch2009newscheme,2015arXiv150503027F}. A tree-based rank $r$ is said admissible if the set  $\Tc_{=r}(\Hc) := \{v : \rank_{\alpha}(v) = r_\alpha,\alpha\in T \}$ of tensors with $T$-rank $r$ is non empty. Necessary conditions of admissibility can be found in \cite[Section 2.3]{2015arXiv150503027F}. In particular, $r_D$ has to be less or equal to $1$ for  $\Tc_{=r}^T$ to be non empty, and $\Tc_{=r}^T$ is reduced to $\{0\}$ if $r_D=0$.

	For a dimension partition tree $T$ and a vertex $\alpha \in T$, we denote by $P(\alpha)$ and $A(\alpha)$ the parent and ascendants of $\alpha$, respectively. The level of a node $\alpha$ is denoted by $\level(\alpha)$. The levels are defined such that $\level(D) = 0$ and $\level(\beta) = \level(\alpha) + 1$ for $\beta \in S(\alpha)$. We let $\depth(T) = \max_{\alpha \in T} \level(\alpha)$ be the depth of $T$, and $\Lc(T)$ be the set of leaves of $T$, which are such that $S(\alpha) = \emptyset$ for all $\alpha \in \Lc(T)$.

\paragraph{Representation of tensors in tree-based format.}	
	Let $u \in \Tc_r^T(\Hc)$ having a tree-based rank $r=(r_\alpha)_{\alpha \in T}$. 
   By definition of the minimal subspaces, we have that 
$u \in \bigotimes_{\alpha \in S(D)} U^{min}_\alpha(u)$, and $U^{min}_{\alpha}(u) \subset \bigotimes_{\beta \in S(\alpha)} U^{min}_\beta(u)$ for any $\alpha \in T \setminus \Lc(T)$ (see \cite{Falco:2012uq}). For any $\alpha \in T $, let $\{u^\alpha_{k_\alpha}\}_{k_\alpha=1}^{r_\alpha}$ be a basis of the minimal subspace $U^{min}_\alpha(u)$. For each $\alpha \in T \setminus \Lc(T)$, we let $I^\alpha = \bigtimes_{\beta\in S(\alpha)}\{1,\dots,r_\beta\}$ and we let
 \begin{align}
 \phi^\alpha_{i_\alpha}(x_\alpha) = \prod_{\beta \in S(\alpha)} u_{k_\beta}^\beta(x_\beta), \quad  i_\alpha = (k_\beta)_{\beta \in S(\alpha)} \in I^\alpha,\label{phi-alpha}
 \end{align}
 be a basis of $\bigotimes_{\beta \in S(\alpha)} U^{min}_\beta(u) \subset \Hc_\alpha$.
Therefore, $u$ admits the following representation
\begin{equation*}
		u(x) =  \sum_{i_D \in I^D} C^D_{i_D,1} \phi^D_{i_D}(x)
	\end{equation*}
	where $C^D \in \Rbb^{I^D \times \{1\}} =  \Rbb^{I^D}$ is a tensor of order $\#S(D)$, 
and for any $\alpha \in T \setminus \{ D\}$, the functions $\{u^\alpha_{k_\alpha}\}_{ k_\alpha =1}^{  r_\alpha}$ admit the following representation 
	\begin{equation}\label{eq: uAlpha}
	u^\alpha_{k_\alpha} (x_\alpha) = \sum_{i_\alpha \in I^\alpha} C^\alpha_{i_\alpha,k_\alpha} \phi_{i_\alpha}^{\alpha}(x_\alpha), \quad 1 \leq k_\alpha \leq r_\alpha,
	\end{equation}
	where $C^\alpha \in \Rbb^{I^\alpha \times \{1,\dots ,r_\alpha\}}$. 
The function $u $ can finally be written 
 $u = F(\bm u)$, with $F : \Rbb^I \to \Hc$  the linear isometry introduced in Section \ref{sec:tensors}, and 
 $\bm u \in \Tc^T_r(\Rbb^I)$ the tensor given by 
	\begin{equation}
		 u_{i_1,\hdots,i_d} 
		 =		\sum_{\substack{1 \leq k_\alpha \leq r_\alpha \\ \alpha \in T}} \prod_{\alpha \in T\setminus \Lc(T)} C^\alpha_{(k_\beta)_{\beta \in S(\alpha)},k_\alpha} \prod_{\alpha \in \Lc(T)} C^{\alpha}_{i_\alpha,k_\alpha}, \label{param-u}
	\end{equation}	
with $C^\alpha \in A^\alpha = \Rbb^{I_\alpha \times \{1,\dots,r_\alpha\}}$ for $\alpha \in T$.
 \begin{remark}
	Note that a function $u \in \Hc$ in a tensor format is associated  with a tensor $\bm{u} = F^{-1}(u)$ in $\Rbb^I$ in the same tensor format. In particular,  $u$ is in $\Tc_r^T(\Hc)$ if and only if  $\bm{u} = F^{-1}(u)  $ is in $\Tc_r^T(\Rbb^{I})$. 
	\end{remark}
	
	\paragraph{Interpretation as compositions of multilinear functions.}	 
For a node $\alpha \in T \setminus \Lc(T) $, a tensor $ C^\alpha \in \Rbb^{I_\alpha \times \{1,\dots ,r_\alpha\}} = \Rbb^{\left( \bigtimes_{\beta \in S(\alpha)} \{1,\dots ,r_\beta\}\right) \times \{1,\dots ,r_\alpha\}}$  can be identified with a $\Rbb^{r_\alpha}$-valued multilinear function $f^\alpha : \bigtimes_{\beta \in S(\alpha)} \Rbb^{r_\beta} \rightarrow \Rbb^{r_\alpha}$. Also, for a leaf node $\alpha\in \Lc(T)$, $ C^\alpha \in \Rbb^{I_\alpha \times \{1,\dots ,r_\alpha\}}$ can be identified with a linear function $ f^\alpha : \Rbb^{\#I^\alpha} \rightarrow \Rbb^{r_\alpha}$.
Denoting by $u^\alpha$ the $\Rbb^{r_\alpha}$-valued function defined for $x_\alpha \in \Xc_\alpha$ by 
${u}^\alpha(x_\alpha) = (u^\alpha_1(x_\alpha),\hdots,u^\alpha_{r_\alpha}(x_\alpha))$, we have 
$$
		{u}^\alpha(x_\alpha) = f^\alpha(\bm\Phi^\alpha(x_\alpha)) \quad \text{for } \alpha \in \Lc(T),
		$$
$${u}^\alpha(x_\alpha) = f^\alpha((u^\beta(x_\beta))_{\beta \in S(\alpha)}) \quad \text{for } \alpha \in T\setminus \Lc(T) ,$$
		and finally 
\begin{align*}
			u(x) = u^D(x) = f^D((u^\alpha(x_\alpha))_{\alpha \in S(D)}).
		\end{align*}
		For example, in the case of Figure \ref{fig:tree-balanced}, the tensor $u$ admits the representation
		\begin{align*}
			u(x) = f^D(f^{\{1,2\}}( f^{\{1\}}(\bm\Phi^1(x_1)), f^{\{2\}}(\bm\Phi^2(x_2))), f^{\{3,4\}}( f^{\{3\}}(\bm\Phi^3(x_3)), f^{\{4\}}(\bm\Phi^4(x_4)))),
		\end{align*}
		and in the case of Figure \ref{fig:tree-linear}, it admits the representation
		\begin{align*}
			u(x) =  f^D( f^{\{1\}}(\bm\Phi^1(x_1)), f^{\{2,3,4\}}( f^{\{2\}}(\bm\Phi^2(x_2)), f^{\{3,4\}}( f^{\{3\}}(\bm\Phi^3(x_3)), f^{\{4\}}(\bm\Phi^4(x_4))))).
		\end{align*}

\subsection{Tree-based tensor formats as multilinear models}\label{sec:multilinear-models}

Let us define  the multilinear map $G:\bigtimes_{\alpha \in T} A^\alpha \to \Rbb^{I} $ such that for a given set of tensors $(C^\alpha)_{\alpha \in T}$, 
the tensor $\bm u = G(C^\alpha)_{\alpha \in T}$ is given by \eqref{param-u}. A function 
$u \in \Tc_r^T$ therefore admits the following parametrization 
\begin{align*}
	u = F \circ G((C^\alpha)_{\alpha \in T}),
	\end{align*}
	where $F \circ G$ is a multilinear map.
	We denote by
	$
		\Psi(x) : \bigtimes_{\alpha \in T} A^\alpha \to \Rbb
	$
	the multilinear map such that 
	\begin{align}
		u(x) = \Psi(x)((C^\alpha)_{\alpha \in T}) \label{multilinear_parametrization_composed}
	\end{align}
	provides the evaluation at $x$ of $u = F\circ G((C^\alpha)_{\alpha \in T})$, and defined by	
		\begin{equation*}
		\Psi(x)((C^\alpha)_{\alpha \in T}) = \sum_{\substack{i_\alpha \in I^\alpha \\ \alpha \in \Lc(T)}} \sum_{\substack{1 \leq k_\alpha \leq r_\alpha \\Â \alpha \in T}} \prod_{\alpha \in T\setminus \Lc(T)} C^\alpha_{(k_\beta)_{\beta \in S(\alpha)},k_\alpha} \prod_{\alpha \in \Lc(T)}  C^\alpha_{i_\alpha,k_\alpha} \phi^\alpha_{i_\alpha}(x_\alpha).
	\end{equation*}	
	For a given $\alpha \in T$ and fixed tensors $(C^{\beta})_{\beta \in T \setminus \{\alpha\}}$, 
	the  partial map $\Psi^\alpha(x) : C^\alpha \mapsto \Psi(x)((C^{\beta})_{\beta \in T})$ is a linear map from $A^\alpha$ to $\Rbb$.
	We can write 
	\begin{equation}\label{eq: uFunOfPhiAndW}
	u(x) 
	= \sum_{1 \leq k_\alpha \leq r_\alpha} u^\alpha_{k_\alpha}(x_\alpha) w^\alpha_{k_\alpha}(x_{\alpha^c})
	= \sum_{1 \leq k_\alpha \leq r_\alpha}\sum_{i_\alpha \in I^\alpha} C^\alpha_{i_\alpha,k_\alpha} \phi_{i_\alpha}^{\alpha}(x_\alpha) w^\alpha_{k_\alpha}(x_{\alpha^c}),
	\end{equation}
	where $u^D_1 = u$ and $w_1^D = 1$ when $\alpha = D$, and 
	\begin{equation*}
		w^\alpha_{k_\alpha}(x_{\alpha^c}) = \sum_{\substack{1 \leq k_\beta \leq r_\beta \\Â \beta \in S(\gamma) \setminus \{\alpha\}}} \sum_{1 \leq k_\gamma \leq r_\gamma} C^\gamma_{(k_\beta)_{\beta \in S(\gamma)},k_\gamma} \prod_{\beta \in S(\gamma) \setminus \{\alpha\}} u^\beta_{k_\beta}(x_\beta) w^\gamma_{k_\gamma}(x_{\gamma^c}), 
	\end{equation*}
	with $\gamma = P(\alpha)$, when $\alpha \neq D$. Functions $\phi^\alpha_{i_\alpha}$ depend on the tensors $\{C_\beta : \beta \in D(\alpha)\}$, with $D(\alpha) = \{\beta \in T : \alpha\in A(\beta)\}$ the set of descendants of $\alpha$, while   functions $w^\alpha_{k_\alpha}$ depend on the tensors $\{C_\beta : \beta \in T\setminus (D(\alpha) \cup \{\alpha\})\}$.

	 When equipping $A^\alpha$ with the canonical inner product $\langle \cdot,\cdot\rangle_\alpha$, $\Psi^\alpha(x)$ can be identified with a tensor $\boldsymbol{\Psi}^\alpha(x)$ in $A^\alpha$ such that 
	\begin{align}
		\Psi^\alpha(x)(C^\alpha) = \langle \boldsymbol{\Psi}^\alpha(x), C^\alpha \rangle_\alpha. \label{partial-map-psialpha}
	\end{align}
	Letting $\bm\Phi^\alpha(x_\alpha)= (\phi^\alpha_{i_\alpha}(x_\alpha))_{i_\alpha\in I^\alpha} \in \Rbb^{I^\alpha}$, with $\phi^\alpha_{i_\alpha}$ defined in Section \ref{sec:tensors} for $\alpha \in \Lc(T)$ and by \eqref{phi-alpha} for $\alpha \in T \setminus \Lc(T)$, we can write 
	 $\bm\Psi^\alpha(x) = \bm\Phi^\alpha(x_\alpha)$ if $\alpha = D$ and $\bm\Psi^\alpha(x) = \bm\Phi^\alpha(x_\alpha) \otimes \bm w^\alpha(x_{\alpha^c})$ otherwise, where $\bm w^\alpha(x_{\alpha^c}) = (w^\alpha_i(x_{\alpha^c}))_{i=1}^{r_\alpha} \in \Rbb^{r_\alpha}$.

\subsection{Representation with orthogonality conditions} 
	The parametrization of a function $ u = F\circ G((C^\alpha)_{\alpha \in T})$ in tree-based format $\Tc_r^T(\Hc)$ is not unique. In other terms, the map $G$ is not injective. For a given $\alpha$, orthogonality conditions can be imposed on the parameters $C^\beta$, for all $\beta \neq \alpha$. More precisely, we can impose orthogonality conditions on matricizations of the tensors $C^\beta$ so that we obtain a representation \eqref{eq: uFunOfPhiAndW} where the set of functions 
	\begin{align*}
	\bm \Psi^\alpha(x)	 = \left(\phi_{i_\alpha}^{\alpha}(x_\alpha) w^\alpha_{k_\alpha}(x_{\alpha^c}) \right)_{i_\alpha \in I^\alpha , 1 \leq k_\alpha \leq r_\alpha}
	\end{align*}
in \eqref{partial-map-psialpha}	forms an orthonormal system in $\Hc$. Algorithms \ref{alg: orth} and \ref{alg: alpha-orth} present the procedure for such an orthogonalization of the representation. They use $\beta$-matricizations $\Mc_{\beta}(C^\alpha)$ of tensors $C^\alpha$, defined in Section \ref{sec:ranks} (see also \cite[Section 5.2]{hackbusch2012book}), and notations
	\begin{equation*}
		s_\alpha = 
		\begin{cases}
			\#S(\alpha) & \text{for } \alpha \in T \setminus \Lc(T) \\
			1 & \text{for } \alpha \in \Lc(T),
		\end{cases}
	\end{equation*}
	and $i^\gamma_\alpha$, for $\alpha \in T \setminus \{D\}$ and $\gamma = P(\alpha)$, such that $\alpha$ is the $i^\gamma_\alpha$-th child of $\gamma$. Note that in Algorithm \ref{alg: alpha-orth}, the matrix $\bm  G^\alpha$ in step \ref{alg: step gramian} is the gramian matrix of the set of functions $\{w_{k_\alpha}^\alpha\}_{k_\alpha = 1}^{r_\alpha}$.
	
	\begin{algorithm}[h]\caption{Orthogonalization of the representation of a function ${u} = F\circ G((C^\alpha)_{\alpha \in T})$ in tree-based tensor format.}\label{alg: orth}
		\begin{algorithmic}[1]
			\REQUIRE {parameters} $(C^\alpha)_{\alpha \in T}$ of ${u}$ 
			\ENSURE new parametrization of ${u}$ with orthonormal sets of functions $\{u^\alpha_{k_\alpha}\}_{k_\alpha = 1}^{r_\alpha}$ for all $\alpha \in T \setminus \{D\}$
			\FOR{$\alpha \in T \setminus \{D\}$ by decreasing level}
				\STATE{set $\gamma = P(\alpha)$}
				\STATE{$\Mc_{\{s_\alpha + 1\}}(C^\alpha)^T = \bm{QR}$ (QR factorization)}
				\STATE{$\Mc_{\{s_\alpha + 1\}}(C^\alpha) \leftarrow \bm{Q}^T$}
				\STATE{$\Mc_{\{i^\gamma_\alpha\}}(C^\gamma) \leftarrow \bm{R}\Mc_{\{i^\gamma_\alpha\}}(C^\gamma)$}
			\ENDFOR
		\end{algorithmic}
	\end{algorithm}
	
	\begin{algorithm}[h]\caption{$\alpha$-orthogonalization of a function ${u} = F\circ G((C^\alpha)_{\alpha \in T})$ in tree-based tensor format.}\label{alg: alpha-orth}
		\begin{algorithmic}[1]
			\REQUIRE {parametrization} of ${u} = F\circ G((C^\alpha)_{\alpha \in T})$ 
			\ENSURE new parametrization of ${u}$  with orthonormal functions $\{u^\alpha_{k_\alpha}\}_{k_\alpha = 1}^{r_\alpha}$ and $\{w^\alpha_{k_\alpha}\}_{k_\alpha = 1}^{r_\alpha}$ for a certain $\alpha \in T \setminus \{D\}$
			\STATE{apply Algorithm \ref{alg: orth} to obtain a parametrization of ${u}$  with orthonormal sets of functions $\{u^\alpha_{k_\alpha}\}_{k_\alpha = 1}^{r_\alpha}$ for all $\alpha \in T \setminus \{D\}$}
			\FOR{$\beta \in (A(\alpha) \setminus \{D\}) \cup \{\alpha\}$ by increasing level}
				\STATE{set $\gamma = P(\beta)$}
				\STATE{$B \leftarrow C^\gamma$}
				\IF{$\gamma \neq D$}
					\STATE{$\Mc_{\{s_\gamma + 1\}}(B) \leftarrow \bm G^\gamma \Mc_{\{s_\gamma + 1\}}(B)$}
				\ENDIF
				\STATE{$\bm G^\beta \leftarrow \Mc_{\{i_\beta^\gamma\}}(C^\gamma)\Mc_{\{i_\beta^\gamma\}}(B)^T$}
			\ENDFOR
			\STATE{set $\gamma = P(\alpha)$}
			\STATE{$\bm G^\alpha = \bm{UDU}^T$ (spectral decomposition of the gramian matrix of the set ${\{w_{k_\alpha}^\alpha\}_{k_\alpha = 1}^{r_\alpha}}$) \label{alg: step gramian}}
			\STATE{$\bm L \leftarrow \bm U\sqrt{\bm D}$}
			\STATE{$\Mc_{\{i^\gamma_\alpha\}}(C^\gamma) \leftarrow \bm L^{-1} \Mc_{\{i^\gamma_\alpha\}}(C^\gamma)$}
			\STATE{$\Mc_{\{s_\alpha + 1\}}(C^\alpha) \leftarrow \bm L^T \Mc_{\{s_\alpha + 1\}}(C^\alpha)$}
		\end{algorithmic}
	\end{algorithm}

\subsection{Singular values of higher-order tensors and tensor truncation}\label{sec:singular-values}
	As seen in Section \ref{sec:ranks}, for a non-empty subset $\alpha $ of $D$, a tensor $u \in \Hc$ can be identified with an order-two  tensor $\Mc_\alpha(u)$ in $\Hc_\alpha \otimes \Hc_{\alpha^c}$, with rank $r_\alpha = \rank(\Mc_\alpha(u)) = \rank_\alpha(u).$ Such an order-two tensor admits a singular value decomposition (SVD)
	\begin{equation*}
		u(x) = \sum_{i=1}^{r_\alpha} \sigma_i^\alpha v_i^\alpha(x_\alpha)   v_i^{\alpha^c}(x_{\alpha^c}), 
	\end{equation*}
	where the $ \sigma^\alpha_i$ are the singular values and $v_i^\alpha$ and $v_i^{\alpha^c}$ are the corresponding left and right singular functions respectively, and where the sets of functions $\{v_i^\alpha\}_{i=1}^{r_\alpha}$ and $\{v_i^{\alpha^c}\}_{i=1}^{r_{\alpha}}$ form  orthonormal systems in $\Hc_\alpha$ and $\Hc_{\alpha^c}$ respectively. The set of singular values of $\Mc_\alpha(u)$ is denoted $\Sigma^\alpha(u) = \{\sigma^\alpha_i\}_{i=1}^{r_\alpha}$, and they are called $\alpha$-singular values of the tensor $u$. Singular values are assumed to be sorted in decreasing order, i.e.  $\sigma^\alpha_1 \ge \hdots \ge  \sigma^\alpha_{r_\alpha}$. 
	\\\par
	An element $u_{\alpha,m_\alpha}$  of best approximation of $u$ in the set $\Tc_{m_\alpha}^{\{\alpha\}}(\Hc)$ of tensors with $\alpha$-rank bounded by $m_\alpha$ (with $m_\alpha \le r_\alpha$), such that
	\begin{align}\label{best-approx-alpha-rank}
		\Vert u-u_{\alpha,m_\alpha} \Vert = \min_{v \in \Tc^{\{\alpha\}}_{m_\alpha}(\Hc)}
		\Vert u - v \Vert,
	\end{align}
	is obtained by truncating the SVD of $\Mc_\alpha(u)$ at rank $m_\alpha$,
	$$
		u_{\alpha,m_\alpha}(x) = \sum_{i=1}^{m_\alpha} \sigma_i^\alpha v_i^\alpha(x_\alpha)   v_i^{\alpha^c}(x_{\alpha^c}),
	$$
	and satisfies 
	$$
		\Vert u-u_{\alpha,m_\alpha} \Vert^2 = \sum_{i=m_\alpha+1}^{r_\alpha} (\sigma_i^\alpha)^2.
	$$
	Denoting by $U^{\alpha}_{m_\alpha}$ the subspace of $\Hc_\alpha$ spanned by  the left singular functions $\{v_i^\alpha\}_{i=1}^{m_\alpha}$ of $\Mc_\alpha(u)$, and by $\Pc^{(\alpha)}_{U^{\alpha}_{m_\alpha}} = \Mc_\alpha^{-1} \circ \left( P^{(\alpha)}_{U^\alpha_{m_\alpha}} \otimes \mathrm{id}_{\alpha^c} \right) \circ \Mc_\alpha$ the orthogonal projection from $\Hc$ onto $\left\{v = \Mc_\alpha^{-1}(w)  : w \in U_{m_\alpha}^\alpha \otimes \Hc_{\alpha^c}\right\}$, we have that $u_{\alpha,m_\alpha} = \Pc^{(\alpha)}_{U^{\alpha}_{m_\alpha}} (u).$ 
	\\\par
	When $T$ is a dimension partition tree over $D$, an approximation $u_{m}$ of a tensor $u$ in the set of tensors $\Tc^T_m(\Hc)$ with $T$-rank bounded by $m= (m_\alpha)_{\alpha\in T}$ can be defined by 
	\begin{align}\label{hierarchical-svd}
		u_{m} = \Pc^{(L)}\hdots \Pc^{(1)} (u), \quad \Pc^{(\ell)} = \prod_{\substack{\alpha \in T \\ \level(\alpha)=\ell}} \Pc^{(\alpha)}_{U^{\alpha}_{m_\alpha}},
	\end{align}
	with $1 \leq \ell \leq L = \depth(T)$.
	The approximation $u_m$ defined by \eqref{hierarchical-svd} is one possible variant of a truncated higher-order singular value decomposition (HOSVD) (see \cite[Section 11.4.2]{hackbusch2012book} or \cite{Nouy2019} for active learning algorithms based on higher-order SVD).
	 
	For a tensor $u$ in the same tree-based tensor format (i.e., $u\in \Tc^T_r(\Hc) $ for some $r\ge m$), the approximation $u_{m}$ can be computed efficiently using standard SVD algorithms.
	In the following, for a tensor $u$ with $T$-rank $r$ and for a certain $m \le r$, we denote by $u_m = \mathrm{Truncate}(u;T,m)$ the truncated HOSVD approximation with $T$-rank $m$.
	
	The approximation $u_m$ obtained by truncated HOSVD   is a quasi-best approximation of $u$ in $\Tc^T_m(\Hc)$ satisfying 
	\begin{align}\label{quasi-best}
		\Vert u-u_m\Vert^2 \le {\sum_{\alpha\in T \setminus \{D\}} \Vert u - \Pc^{(\alpha)}_{U^{\alpha}_{m_\alpha}} (u) \Vert^2} \le ({\#T-1})  \min_{v \in \Tc^T_m(\Hc)} \Vert u - v \Vert^2.
	\end{align}
	If for all $\alpha$, the truncation rank $m_\alpha$ is chosen such that $\Vert u - \Pc^{(\alpha)}_{U^{\alpha}_{m_\alpha}} (u) \Vert^2 \le {\epsilon^2}({{\#T-1}})^{-1} \Vert u \Vert^2$, which means 
	\begin{align}\label{hosvd-relative-precision-choice-ranks}
		\sum_{i=m_\alpha+1}^{r_\alpha} (\sigma_i^\alpha)^2 \le \frac{\epsilon^2}{\#T-1} \sum_{i=1}^{r_\alpha} (\sigma_i^\alpha)^2,
	\end{align}
	then $u_m$ provides an approximation of $u$ with relative precision $\epsilon$, i.e.,
	\begin{align}\label{hosvd-relative-precision}
		\Vert u -u_m \Vert \le \epsilon \Vert u \Vert.
	\end{align}
	In the following, for a tensor $u$ and a certain $\epsilon<1$,  we denote by $u_{m(\epsilon)} = \mathrm{Truncate}(u;T,\epsilon)$ the truncated HOSVD approximation of $u$ with $T$-rank $m(\epsilon)$ chosen as the highest tuple satisfying \eqref{hosvd-relative-precision-choice-ranks}, which ensures \eqref{hosvd-relative-precision}. Algorithm \ref{alg: truncate} presents the procedure for applying $\mathrm{Truncate}(u;T,\epsilon)$. It is similar to Algorithm \ref{alg: alpha-orth}, with additional truncation steps.

	\begin{algorithm}[h]\caption{Algorithm of the function $\mathrm{Truncate}(u;T,\epsilon)$.}\label{alg: truncate}
		 
		\begin{algorithmic}[1]
			\REQUIRE parameters $(C^\alpha)_{\alpha \in T}$ of a function $u \in \Tc_r^T(\Hc)$
			\ENSURE approximation $u_m \in \Tc_m^T(\Hc)$ satisfying \eqref{hosvd-relative-precision}
			\STATE{use Algorithm \ref{alg: orth} to obtain orthonormal sets of functions $\{u^\alpha_{k_\alpha}\}_{k_\alpha = 1}^{r_\alpha}$, $\alpha \in T \setminus \{D\}$}
			\FOR{$\alpha \in T \setminus \{D\}$ by increasing level}
				\STATE{set $\gamma = P(\alpha)$}
				\STATE{$B \leftarrow C^\gamma$}
				\IF{$\gamma \neq D$}
					\STATE{$\Mc_{\{s_\gamma + 1\}}(B) \leftarrow \bm G^\gamma \Mc_{\{s_\gamma + 1\}}(B)$}
				\ENDIF
				\STATE{$\bm G^\alpha \leftarrow \Mc_{\{i^\gamma_{\alpha}\}}(C^\gamma)\Mc_{\{i^\gamma_{\alpha}\}}(B)^T$ with eigenvalues $((\sigma^\alpha_i)^2)_{i = 1}^{r_\alpha}$}
			\ENDFOR
			
			\FOR{$\alpha \in T \setminus \{D\}$ by decreasing level}
				\STATE{$\bm G^\alpha \approx \bm{UDU}^T$ such that \eqref{hosvd-relative-precision-choice-ranks} is ensured (truncated spectral decomposition of the gramian matrix of the set of functions ${\{w_{k_\alpha}^\alpha\}_{k_\alpha = 1}^{r_\alpha}}$ at rank $m_\alpha$)}
				\STATE{$\Mc_{\{i^\gamma_\alpha\}}(C^\gamma) \leftarrow \bm U^T \Mc_{\{i^\gamma_\alpha\}}(C^\gamma)$}
				\STATE{$\Mc_{\{s_\alpha + 1\}}(C^\alpha) \leftarrow \bm U^T \Mc_{\{s_\alpha + 1\}}(C^\alpha)$}
			\ENDFOR
		\end{algorithmic}
	
	\end{algorithm}

\subsection{Sparse representations in tree-based tensor formats} \label{sparse_tensors}
	A function $u$ has a sparse representation in a certain tree-based format if it admits a parametrization \eqref{multilinear_parametrization_composed} where 
	parameters $C^\alpha$ contain zero entries.
	When the parameter $C^\alpha$ is an array containing the coefficients of functions on a given basis of functions $\bm\Phi^\alpha(x_\alpha)= (\phi^\alpha_i(x_\alpha))_{i\in I^\alpha}$ (see Section \ref{sec:multilinear-models}), sparsity in $C^\alpha$ means sparsity of the corresponding functions relatively to the basis $\bm\Phi^\alpha(x_\alpha)$. The choice of the bases is crucial for the existence of a sparse representation of a tensor (or of an accurate approximation with a sparse representation).  For leaf nodes $\alpha \in \Lc(T)$, 
typical choices of bases $\bm \Phi^\alpha$  include polynomials, wavelets or other bases for multiresolution analysis, where the set of basis functions can be partitioned into subsets of basis functions with different levels (or resolutions). 
For interior nodes $\alpha \in T\setminus \Lc(T)$, bases $\bm \Phi^\alpha$ introducing sparsity for the representation of a function $u$ can be obtained by using for the bases $\{u_{k_\beta}^\beta\}$ of minimal subspaces $U^{min}_\beta(u)$ the principal components of $\alpha$-matricizations of $u$ obtained with singular value decomposition (ordered by decreasing singular values). Algorithm \ref{alg: truncate} with $\epsilon = 0$ yields such a representation with a natural hierarchy in the bases $\{u_{1}^\beta,\dots,u_{r_\beta}^\beta\}$, which induces a natural hierarchy in the bases $\bm \Phi^\alpha$ of interior nodes that are obtained through tensorization of bases $\{u_{1}^\beta,\dots,u_{r_\beta}^\beta\}$, $\beta\in S(\alpha)$. 
	
	For such bases with a natural hierarchy of the basis functions, we introduce a nested sequence $I_0^\alpha \subset \hdots \subset I_{p}^{\alpha} $ of subsets in $I^\alpha$, with $I_{p}^{\alpha} = I^\alpha$, such that $\{\phi^\alpha_i(x_\alpha)\}_{i\in I^\alpha_\lambda}$ is a basis of a subspace of functions $\Hc_\alpha^{\lambda} \subset \Hc_\alpha$,  with level $\lambda$. This defines a sequence of nested spaces 
	$$
		\Hc^0_\alpha \subset \hdots \subset \Hc^{p}_\alpha \subset \Hc_\alpha.
	$$
			\begin{example}[Hierarchical bases for the leaves $\alpha \in \Lc(T)$]\label{ex: multilevelBases}
			$\,$
			\begin{itemize}
				\item For a univariate polynomial basis, where $\phi^\alpha_i$ is a polynomial of degree $i-1$, we simply take $I^\alpha_{\lambda} = \{1,\hdots,\lambda+1\}$, so that $\Hc_\alpha^\lambda = \Pbb_{\lambda}(\Xc_\alpha)$ is the space of polynomials with degree $\lambda$.
				\item For a multivariate polynomial basis on $\Xc_\alpha \subset \Rbb^{d_\alpha}$, $\Hc_\alpha^\lambda$ can be chosen as the space of polynomials with partial (or total) degree $\lambda$.
				\item For a univariate wavelet (or multiresolution) basis, $\Hc_\alpha^\lambda$ can be taken as the space of functions with resolution $\lambda$.
			\end{itemize}
		\end{example}
	For a tensor $u$ in tree-based tensor format and a node $\alpha \in T$, where $C^\alpha \in \Rbb^{I^\alpha} \otimes \Rbb^{r_\alpha}$ collects the coefficients of the functions $\{u^\alpha_k(x_\alpha)\}_{k = 1}^{r_\alpha}$ on a basis $\bm\Phi^\alpha(x_\alpha)$, we define
	\begin{align*}
		K^\alpha_\lambda = I^\alpha_\lambda \times \{1,\hdots,r_\alpha\},
	\end{align*}
	so that $K^\alpha_0 \subset \hdots \subset K^\alpha_p $ form a nested sequence of subsets in $K^\alpha =I^\alpha \times  \{1,\hdots,r_\alpha\}$, with $K^\alpha_p = K^\alpha$. Therefore, if $C^\alpha = (C^\alpha_{k})_{k \in K^\alpha}$ is such that $C^\alpha_{k} = 0$ for all $k \notin K^\alpha_\lambda$, then all functions $u^\alpha_k(x_\alpha)$ are in the space $\Hc_\alpha^\lambda$, and $u \in \Hc_\alpha^\lambda \otimes \Hc_{\alpha^c}$.
	
	The proposed sequence of candidate patterns $K^\alpha_0 \subset \hdots \subset K^\alpha_p$ for the parameter $C^\alpha$ can be used in learning algorithms using working set strategies for sparse approximation (see Section \ref{sec:exploit-sparsity}).

\section{Statistical learning using tree-based formats}\label{sec:learning}
Let ${Z = (X,Y)}$ be a pair of random variables taking values in $\Xc \times \Yc$. The aim of supervised learning is to construct an approximation of $Y$ as a function of $X$ (predictive model), from a set $S = \{{z_i}\}_{i=1}^n $ of $n$ realizations of ${Z}$. We introduce a loss function $\ell: \Yc\times \Yc \to \Rbb$ such that for a given ${z = (x,y)}$ and a function $v$, $\ell(y,v(x))$ provides a measure of the error between $y$ and the prediction $v(x)$.
Letting $\gamma$ be the contrast function defined as $\gamma(v,(x,y)) = \ell(y,v(x))$, an approximation can be obtained by minimizing the \emph{empirical risk}
\begin{equation*}
	\Rc_n(v) = \frac{1}{n} \sum_{i=1}^n {\gamma(v,z_i)}
\end{equation*}
over a certain subset of functions $v$ (hypothesis set).
The empirical risk is a statistical estimate of the \emph{risk}
$$
	\Rc(v) = \Ebb({\gamma(v,Z)}).
$$
In least-squares regression, the contrast is taken as $\gamma(v,z) = (y - v(x))^2$, and
\begin{equation*}
	\Rc(v) = \Ebb((Y-u(X))^2) + \Ebb((u(X)-v(X))^2) = \Rc(u) + \Vert u-v \Vert^2,
\end{equation*}
where $\Vert \cdot\Vert$ is the $L^2_\mu$-norm (with $\mu$ the probability law of $X$) and $u(X) = \Ebb(Y \vert X)$ is the best approximation of $Y$ by a measurable function of $X$, so that the minimization of the risk corresponds to the minimization of the approximation error $\Vert u-v \Vert^2.$ Other contrast functions could be considered for other purposes in supervised learning  (such as hinge loss or logistic loss for classification).

 In the context of density estimation, $Z = X$ is a random variable that is assumed to have a density $u$ with respect to some measure $\mu$, and $S$ is a set of $n$ samples of $X$. Choosing  the contrast function as $\gamma(v,x) = \|v\|^2 - 2v(x)$, with $\Vert \cdot \Vert$ the natural norm in $L^2_\mu$, leads to
\begin{equation*}
\Rc(v) = \Rc(u) + \|u-v\|^2,
\end{equation*}
so that the minimization of $\Rc(v)$ is equivalent to the minimization of the distance (in $L^2_\mu$ norm) between $v$ and the density $u$.
Choosing $\gamma(v,z) = -\log(v(z))$ leads to
\begin{equation*}
	\Rc(v) = \Rc(u) + D_{\text{KL}}(u \| v),
\end{equation*}
with $D_{\text{KL}}(u \| v)$ the Kullback-Leibler divergence between $u$ and $v$, and the empirical risk minimization corresponds to a maximum likelihood estimation.

\subsection{Empirical risk minimization}
	An approximation of $u$ in a tensor format can be obtained by minimizing the empirical risk  over the set of functions 
	$$
		v(x) = \Psi(x)((C^\alpha)_{\alpha \in T}),
	$$
	where the $(C^\alpha)_{\alpha \in T} \in \bigtimes_{\alpha \in T} A^\alpha$ are the parameters of the representation of $v$ and
	$\Psi(x) : \bigtimes_{\alpha \in T} A^\alpha \to \Rbb $ is a multilinear map depending on the chosen tensor format (see Section \ref{sec:multilinear-models}). This yields the optimization problem
	\begin{equation}\label{min_risk}
		\min_{(C^\alpha)_{\alpha \in T}} \frac{1}{n} \sum_{i=1}^n
		{\gamma(\Psi(\cdot)((C^\alpha)_{\alpha \in T}),z_i)}
	\end{equation}
	over the set of parameters.
	%
	For the solution of the optimization problem \eqref{min_risk}, we use an alternating minimization algorithm which consists in successively solving 
	\begin{equation}\label{min_risk_alternate}
		\min_{C^\alpha} \frac{1}{n} \sum_{i=1}^n 
		{\gamma(\Psi(\cdot)((C^\beta)_{\beta \in T}),z_i)}
	\end{equation}
	for fixed parameters $C^\beta$, $\beta\neq \alpha$. Letting $\Psi^\alpha$ be the {(linear)} partial map defined by \eqref{eq: uFunOfPhiAndW} yields the optimization problem  
	\begin{equation}\label{min_risk_nu}
		\min_{C^\alpha} \frac{1}{n} \sum_{i=1}^n 
		{\gamma(\Psi^\alpha(\cdot)(C^\alpha),z_i)}.
	\end{equation}
	The problem is then reduced to a succession of learning problems with linear models. The algorithm is described in Algorithm \ref{alg:learning_als}, where  step \ref{als-step-estimate-p} consists in solving \eqref{min_risk_nu}.
	
	\begin{algorithm}\caption{Learning algorithm for an approximation in a given subset of low-rank tensors.}\label{alg:learning_als}
		\begin{algorithmic}[1]
			\REQUIRE sample $S = \{(z_i)\}_{i=1}^n$, contrast function, tensor format with a multilinear parametrization $\Psi(\cdot)((C^\alpha)_{\alpha \in T})$, and initial values for the parameters $\{C^\alpha\}_{\alpha \in T}$
			\ENSURE approximation $v(\cdot) = \Psi(\cdot)((C^\alpha)_{\alpha \in T})$
			\WHILE{not converged}
				\FOR{$\alpha \in T$}
					\STATE estimate $C^\alpha$ for fixed parameters $C^\beta$, $\beta\neq \alpha$ (learning problem with a linear model)\label{als-step-estimate-p}
				\ENDFOR
			\ENDWHILE
		\end{algorithmic}
	\end{algorithm}

	
	\paragraph{Regularization (or penalization).}
	The optimization problem \eqref{min_risk}
	 may be replaced by 
	 \begin{equation}\label{min_risk_regul}
		 \min_{(C^\alpha)_{\alpha \in T}}  \frac{1}{n} \sum_{i=1}^n
		 {\gamma(\Psi(\cdot)((C^\alpha)_{\alpha \in T}),z_i)}
		 + \sum_{\alpha \in T} \lambda_\alpha\Omega_\alpha(C^\alpha), 
	\end{equation}
	where $\lambda_\alpha \Omega_\alpha(C^\alpha)$ is a regularization (or penalization) term promoting some properties for the parameter $C^\alpha$ (e.g., sparsity or smoothness of functions associated with the parameter $C^\alpha$). Regularization may be required for stability when only a few training samples are available or for exploiting a prior information on the parameters (with a bayesian point of view).
	\begin{example}
	    Usual regularization functions $ \Omega_\alpha(C^\alpha)$ for promoting sparsity include the $\ell^0$-norm $\Vert C^\alpha \Vert_0$, which is the number of non-zero coefficients in $C^\alpha$, or its convex regularization provided by the $\ell^1$-norm $\Vert C^\alpha \Vert_1$ (see \cite{bach2012optimization}).
	\end{example}
	
	When using an alternating minimization algorithm to solve \eqref{min_risk_regul} (Algorithm  \ref{alg:learning_als}), the problem is reduced to the solution of successive learning problems with linear models (step \ref{als-step-estimate-p} of the algorithm)
	\begin{equation}\label{min_risk_nu_regul}
		\min_{C^\alpha} \frac{1}{n} \sum_{i=1}^n 
		{\gamma(\Psi^\alpha(\cdot)(C^\alpha),z_i)}
		 + \lambda_\alpha \Omega_\alpha(C^\alpha),
	\end{equation}
	 for which efficient algorithms are usually available (for standard {contrast} functions and regularization functionals). Also, standard statistical methods such as cross-validation (see Section \ref{sec:cv}) can be used for the selection of the regularization parameter $\lambda_\alpha$ and of the regularization functional $\Omega_\alpha$ (possibly depending on other parameters).
	

\subsection{Validation and model selection}\label{sec:cv}
	We first recall the principle of validation methods for the estimation of the risk $\Rc( v_S)$ for a function $v_S$  estimated from the 
	sample $S = \{ {z_i}\}_{i=1}^n$ (see, e.g., \cite{arlot2010survey}). If $V$ is a sample independent of $S$,  
	then 
	$$
		\Rc_{V}( v_{S}) = \frac{1}{\#V} \sum_{{z}\in V}  {\gamma(v_S,z)}
	$$
	provides an unbiased estimator of the risk $\Rc( v_S)$. If $S$ is the only available information, the risk can be estimated 
	by the \emph{hold-out estimator}
	$
		\Rc_{V}( v_{S\setminus V}),
	$
	where $V \subset S$ is a {validation sample} contained in the sample $S$, and 
	$ v_{S\setminus V}$ is the model estimated from the sample $S\setminus V$. Also, by introducing a partition of $S$ into $L$ validation samples $V_1,\hdots,V_L$, we define the \emph{cross-validation estimator} of the risk $\Rc( v_S)$ by 
	$$
		\Rc^{CV}( v_S) = \frac{1}{L} \sum_{i=1}^L \Rc_{V_i}( v_{S\setminus V_i}).
	$$
	The case where $L=n$, with $V_i=\{ {z_i}\}$, corresponds to the \emph{leave-one-out (LOO) estimator}
	$$
		\Rc^{LOO}(v_S) = \frac{1}{n} \sum_{i=1}^n {\gamma(v_{-i},z_i)},
	$$
	with $ v_{-i} =  v_{S\setminus \{{z_i}\}}.$ In some cases, cross-validation estimators can be computed efficiently, without computing models $ v_{S\setminus V_i}$ for all $i$.
	\begin{example}[LOO estimator for ordinary least-squares regression]\label{ex:loo-ls}
		Consider the case of least-squares regression, with $\ell(y,y') = (y-y')^2$, and consider a linear model $ v_S(x) = \boldsymbol{\Psi}(x)^T {\bm{a}}_S$, with a given $\boldsymbol{\Psi}(x) \in \Rbb^m$, and  coefficients $ {\bm{a}}_S \in \Rbb^m$ obtained by ordinary least-squares regression, i.e., by minimizing the empirical risk $\Rc_S(\boldsymbol{\Psi}(\cdot)^T {\bm{a}}) = \frac{1}{n} \Vert \bm{y} - \bm{A}\bm{a} \Vert_2^2$ over $\bm{a}\in \Rbb^m$, where $\bm{y} = (y_i)_{i=1}^n \in \Rbb^n$ and $\bm{A} \in \Rbb^{n\times m}$ is the matrix whose $i$-th row is $\boldsymbol{\Psi}(x_i)^T$. In this case, the leave-one-out estimator of the risk is 
		$$
			\Rc^{LOO}( v_S) = \frac{1}{n} \sum_{i=1}^n \left(\frac{y_i - v_S(x_i)}{1-h_{i}}\right)^2,
		$$
		where $h_{i}$ is the diagonal term $\bm{H}_{ii}$ of the matrix $\bm{H}=\bm{A}(\bm{A}^T\bm{A})^{-1}\bm{A}^T$. This estimator only depends on $v_S$ and not on the functions $v_{-i}$, $1 \leq i \leq n$.
	\end{example}
	\begin{remark}[Corrected estimators for ordinary least-squares regression]\label{corrected-estimator}
		When the sample size $n$ is small compared to the number of parameters $m$, several corrected estimators have been proposed. For the case of ordinary least-squares regression with linear models described in Example \ref{ex:loo-ls}, a corrected estimator 
		has been proposed in \cite{chapelle2002model} in the form $$\tilde \Rc^{LOO}( v_S) = \Rc^{LOO}( v_S)\left(1-\frac{m}{n}\right)^{-1}\left(1+\frac 1 n trace(\bm{G}^{-1}\bm{\bar G})\right),$$ where $\bm{\bar G} $ is the Gram matrix $\Ebb(\boldsymbol{\Psi}(X)\boldsymbol{\Psi}(X)^T)$ of $\boldsymbol{\Psi}(X)$ and $\bm G = \frac{1}{n} \sum_{i=1}^n \boldsymbol{\Psi}(x_i)\boldsymbol{\Psi}(x_i)^T = \frac 1 n \bm A^T \bm A$ is the empirical Gram matrix. When $n\to \infty$, $\bm{G}$ converges (almost surely) to $\bm{\bar G}$, so that $trace(\bm{G}^{-1}\bm{\bar G}) $ converges to $m$ and the correction factor converges to $1$.
	\end{remark}

	\paragraph{Model selection.}
		Cross-validation estimators can be used for model selection. 
		Suppose that different functions $ v^\lambda_S$, $\lambda \in \Lambda$, have been estimated from the sample $S$ (e.g., low-rank models with different bases, different ranks or different trees, see Section \ref{sec:adaptiveAlgorithms}). Denoting by $\Rc^{CV}(v_S^\lambda)$ the estimator of the risk for the model $v_S^\lambda$, the optimal model with respect to this estimation of the risk is $v_S^{ \lambda_S}$ with $\lambda_S$ such that
		$$
			\Rc^{CV}(v_S^{\lambda_S}) =  \min_{\lambda \in \Lambda} \Rc^{CV}(v_S^\lambda).
		$$

\subsection{Exploiting sparsity}\label{sec:exploit-sparsity}
	Here we present possible strategies for exploiting sparsity in the parameters $C^\alpha$ of a function in tree-based format, where $C^\alpha = (C^\alpha_k)_{k\in K^\alpha} \in \Rbb^{K^\alpha}$, with $K^\alpha$ a finite set of indices.
	We define the support, or \emph{pattern}, of $C^\alpha $  as $\mathrm{support}(C^\alpha) = \{ k \in K^\alpha : C^\alpha_k \neq 0 \}$. 
	We will first describe how to select a pattern among a set of candidate patterns for $C^\alpha$. Then, we will describe how to determine a set of candidate patterns.
	
	\paragraph{Selection of a pattern from a set of candidate patterns.}
		Let us suppose that we have a collection of candidate patterns $K_\lambda^\nu$, $\lambda\in \Lambda$, for the parameter $C^\alpha$. At step $\alpha$ of the alternating minimization procedure for empirical risk minimization (step  \ref{als-step-estimate-p} of Algorithm \ref{alg:learning_als}),
		instead of solving \eqref{min_risk_nu}, we compute for all $\lambda\in \Lambda$ the solution $C^{\alpha,\lambda}$ of
		\begin{align}\label{ols-pattern}
			\min_{C^\alpha} \frac{1}{n} \sum_{i=1}^n 
			{\gamma(\Psi^\alpha(\cdot)(C^\alpha),z_i)}
			\quad \text{subject to } \mathrm{support}(C^\alpha) \subset K^\alpha_\lambda,
		\end{align}
		which provides a collection of approximations {$v^{\lambda}(\cdot) = {\Psi}^\alpha(\cdot)(C^\alpha)$}, $\lambda\in \Lambda$.   
		Then, cross-validation methods can be used in order to select  a particular solution $C^\alpha = C^{\alpha,\hat \lambda}$, with $\hat \lambda$ minimizing over $\lambda \in \Lambda$ a certain cross-validation estimator of the risk $\Rc(v^\lambda)$ (see Section \ref{sec:cv}). Usually, the pattern of $C^{\alpha,\lambda}$ will coincide with $K^\alpha_\lambda$ but it may be strictly contained in $K^\alpha_\lambda$.
		\begin{remark}
			An equivalent form of problem \eqref{ols-pattern} is given by
			\begin{align}\label{ols-pattern-regulform}
				\min_{C^\alpha} \frac{1}{n} \sum_{i=1}^n 
				{\gamma(\Psi^\alpha(\cdot)(C^\alpha),z_i)}
				 +  \Omega^{\lambda}_\alpha(C^\alpha),
			\end{align}
			where $\Omega^{\lambda}_\alpha$ is the characteristic function of the subset of elements $C^\alpha$ whose support is contained in $K^\alpha_\lambda$, i.e. 
			$\Omega^{\lambda}_\alpha(C^\alpha) = 0$ if $\mathrm{support}(C^\alpha) \subset K^\alpha_\lambda$  and $\Omega^{\lambda}_\alpha(C^\alpha) = +\infty$ if $\mathrm{support}(C^\alpha) \not\subset K^\alpha_\lambda$. This formulation can be seen as a regularized version of the empirical risk minimization problem, where $\lambda$ plays the role of the regularization parameter which can be estimated using cross-validation methods. 
		\end{remark}
	
	\paragraph{Determination of the set of candidate patterns.}
		Let us now discuss how to propose the set of candidate patterns $K_\lambda^\alpha$, $\lambda\in \Lambda$.
		In the case where  $C^\alpha$ collects the coefficients of functions on a hierarchical (or multilevel) basis $\bm\Phi^\alpha(x_\alpha)$, the patterns can be determined \emph{by hand} as a nested sequence of patterns $K_0^\alpha\subset \hdots \subset K_p^\alpha$, where $K_\lambda^\alpha$ is associated with a basis of level $\lambda$ (e.g., for polynomial approximation, $\lambda$ may be the degree of the polynomial space); see Example \ref{ex: multilevelBases}. 
		The set of candidate patterns can also be determined automatically by using a greedy algorithm, such as a matching pursuit algorithm \cite{mallat1993matching}, which  provides a sequence of nested patterns $K_0^\alpha,\hdots,K_{\# K^\alpha -1}^\alpha$. Another approach consists in solving \eqref{min_risk_nu_regul} with a sparsity-inducing penalization for several values of $\lambda = \lambda_\alpha$, therefore leading to a collection of solutions $v^{\lambda}$. The solutions $v^{\lambda}$ may be directly considered as the set of candidate approximations. However, in practice, we  extract the patterns $K^\alpha_{\lambda}$ of the different approximations and re-estimate the coefficients by solving the problem \eqref{ols-pattern} without regularization. This usually provides estimates with better statistical properties and  allows the use of fast procedures for the estimation of cross-validation estimators \cite{belloni2013}.

		\begin{example}
			When using a square loss and a $\ell^1$-norm regularization, and assuming (up to a vectorization) that $C^\alpha\in\Rbb^{m_\alpha}$, with $m_{\alpha} = \#K^\alpha$, \eqref{min_risk_nu_regul} is a LASSO problem for which efficient algorithms are available, such as the LARS algorithm \cite{efron2004least}, which directly provides a set of solutions associated with different patterns (the so-called \emph{regularization path}). 
		\end{example}

\section{Adaptive approximation in tree-based tensor formats} \label{sec:adaptiveAlgorithms}
In this section, we present adaptive learning algorithms for the approximation of functions in tree-based tensor format. 
First, we present an algorithm with rank adaptation for the approximation in tree-based tensor format with a given dimension tree. 
Then, we present a strategy for the adaptive selection of a tree.

\subsection{Rank adaptation}
Here, we propose a learning algorithm for the approximation in tree-based tensor format with rank adaptation. 
The tree $T$ is supposed to be given. 
This algorithm provides a sequence of approximations $u^m \in \Tc_{r^m}^T$  with increasing $T$-ranks $r^m = (r^m_\alpha)_{\alpha\in T}$ (relatively to the partial order on $\Nbb^d$). 
The sequence is defined as follows. 
We start by computing an approximation $u^1$ with $T$-rank $r^1=(1,\hdots,1)$ using Algorithm \ref{alg:learning_als}.
At iteration $m$, given an approximation $u^m $ with $T$-rank $r^m$, we first define the $T$-rank $r^{m+1}$ of the next iterate by
	\begin{align}\label{update-ranks}
		r^{m+1}_\alpha = \begin{cases} r^m_\alpha +1 & \text{if } \alpha \in T_m^\theta, \\
		r^m_\alpha & \text{if } \alpha \notin T_m^\theta,
		\end{cases}
	\end{align}
	where $T_m^\theta$ is a suitably chosen subset of nodes (see below), 
and we use Algorithm \ref{alg:learning_als} for obtaining an approximation  
$u^{m+1}$ in the set $\Tc_{r^{m+1}}^{T}(\Hc)$ of tensors with fixed tree-based rank $r^{m+1}$. Algorithm \ref{alg:rank-optim} presents this learning algorithm in a given tree-based tensor format with rank adaptation, which returns an element of the generated sequence of approximations selected using an estimation of the generalization error. 

	\begin{algorithm}\caption{Learning algorithm in a given tree-based tensor format with rank adaptation.}\label{alg:rank-optim}
		\begin{algorithmic}[1]
			\REQUIRE sample $S = \{(z_i)\}_{i=1}^n$, contrast function, dimension tree $T$ and maximal number of iterations $M$
			\ENSURE approximation $u_r$ in $\Tc_{r}^{T}(\Hc)$ with $T$-rank $r$
			\STATE compute an approximation $u^1$ with $T$-rank $r^1 = (1,\hdots,1)$ using Algorithm \ref{alg:learning_als}
			\FOR{$m=1,\hdots,M-1$}
				\STATE compute $T_m^\theta$ with Algorithm \ref{alg:Tm} and $r^{m+1}$ defined by \eqref{update-ranks} \label{step-selection-Tm}
				\STATE compute an approximation $u^{m+1}$ in $\Tc_{r^{m+1}}^{T}(\Hc)$ using Algorithm \ref{alg:learning_als}\label{alg:rank-optim-learning-step}
				\label{alg:rank-optim-estimate}
			\ENDFOR
			\STATE{select $m^\star = \arg\min_{1\le m \le M} \Rc_{V}(u^{m})$, where $V$ is a validation set independent of $S$, and return $u_r = u^{m^\star}$, with $r=r^{m^\star}$ \label{alg:modelSelectionStep}}
		\end{algorithmic}	
	\end{algorithm}
	
	The crucial ingredient of Algorithm \ref{alg:rank-optim} is the selection of the subset of nodes 
	$T_m^\theta$ at Step \ref{step-selection-Tm}. A natural idea is to select the nodes $\alpha\in T$ associated with the highest truncation errors  $$  \min_{\rank_{\alpha}(v)\le r^m_\alpha} \Rc(v) - \Rc(u) := \eta_\alpha(u,r_\alpha^m)^2,$$ where $u$ is the oracle function. When  $\Rc(v) - \Rc(u) $
	is the square of the distance in $L^2_\mu$-norm between $v$ and $u$ (for least-squares regression or density estimation in $L^2_\mu$), then the truncation errors become $$\eta_\alpha(u,r_\alpha^m)^2 = \min_{\rank_{\alpha}(v)\le r^m_\alpha} \Vert v - u \Vert_{L^2_\mu}^2= \sum_{k > r^m_\alpha} (\sigma_k^\alpha(u))^2 ,$$ where $\sigma_k^\alpha(u)$  are the $\alpha$-singular values of $u$. 
	In practice, these truncation errors are estimated by $\eta_\alpha^2(\tilde u;r^m_\alpha)$ where $\tilde u$ is an approximation obtained by a correction of $u^m$. The algorithm for the selection of $T_m^\theta$ is described in Algorithm \ref{alg:Tm}, and proceeds as follows. 
	We first compute an approximation $\tilde u$ with a rank $ r $ such that $ r^m_\alpha \le  r_\alpha \le r^{m}_\alpha+1$ for all $\alpha\in T$\footnote{An approximation $\tilde u$ with higher rank could be computed in order to improve the estimation of the truncation errors.}. In practice, it is done by first computing a rank-one correction $w$ of $u^m$, which yields an approximation $u^m +w$ with $T$-rank  $ r \ge r^m$, and then by using Algorithm \ref{alg:learning_als}, with $u^m+w$ as an initialization, to compute an approximation $\tilde u$ with rank $ r$. 
	Then, we compute the $\alpha$-singular values $\Sigma^\alpha(\tilde u) = \{\sigma_i^\alpha\}_{i=1}^{r_\alpha}$ of $\tilde u$ for all $\alpha \in T \setminus \{D\}$ (see Section \ref{sec:singular-values}), and 
	 define a subset $\hat T$ of candidate nodes for increasing the ranks as follows:
	\begin{equation} \label{eq:Tbar}
		\hat T = T \setminus \left( \{D\} \cup \{\alpha \in \Lc(T) : r_\alpha = \#I^\alpha\} \cup \{\alpha \in T \setminus \{D\} : \eta_\alpha(\tilde u ; r^m_\alpha) \leq \varepsilon \|\tilde u\|\} \right),
	\end{equation}
	which contains all the nodes of $T$ except the root, the leaf nodes $\alpha$ for which $r_\alpha = \#I^\alpha = N_\alpha$ (necessary condition for the tree-based rank to be admissible, see~\cite[Section 2.3]{2015arXiv150503027F}) and the nodes for which $\eta_\alpha(\tilde u ; r^m_\alpha)$ is smaller than a constant $\varepsilon$ (typically machine precision) multiplied by $\|\tilde u\|$. Then we define 
	\begin{equation}
	T_m^\theta = \{\alpha \in \hat T :  \eta_\alpha(\tilde u ; r^m_\alpha) \ge \theta \max_{\beta \in \hat T} \eta_\beta(\tilde u ; r^m_\beta) \}, \label{Tmtheta}
	\end{equation}
	where $\theta \in [0,1]$ is a parameter controlling the number of nodes to be selected.
	
	Note that since $r_\alpha = r_\alpha^{m}+1$ for all $\alpha \in \hat T$, $\eta_\alpha(\tilde u ; r^m_\alpha) = \sigma_{r_\alpha}^\alpha,$ the minimal $\alpha$-singular value of $\tilde u$. 
	When $\theta = 0$,  $T_m^\theta = \hat T$, which means that $\alpha$-ranks are increased for every $\alpha \in \hat T$, so possibly for every $\alpha \in T\setminus \{D\}$, which is not a desired adaptive strategy. 
	When $\theta = 1$, we increase only the ranks associated with nodes $\alpha$ such that $\sigma_{r_\alpha}^\alpha = \max_{\beta \in \hat T} \sigma_{r_\beta}^\beta $ (usually a single node). This choice results in a slow increase of the ranks and possibly to non admissible ranks. To ensure that $r^{m+1}$ satisfies the necessary conditions of admissibility, we select $\theta$ automatically as the highest value in $[0,\theta^\star]$ such that $r^{m+1}$ is admissible, where $\theta^\star \in [0,1]$ is a user-defined parameter selected in order to increase sufficiently many ranks at each iteration. In the following numerical experiments, $\theta^\star$ is chosen equal to $0.8$.

	\begin{algorithm}\caption{Computation of the subset  of nodes $T_m^\theta$ whose rank is increased.}\label{alg:Tm}
		\begin{algorithmic}[1]
			\REQUIRE sample $S = \{(z_i)\}_{i=1}^n$, approximation $u^m \in \Tc_{r^m}^T$, parameter $\theta^\star$
			\ENSURE subset $T_m^\theta$  
			\STATE{use Algorithm \ref{alg:learning_als} to compute an approximation  $\tilde u$ with ranks $ r_\alpha \in \{r_\alpha^m ,  r_\alpha^m+1 \}$, $\alpha\in T$}\label{step-v}
			\STATE{compute the set of $\alpha$-singular values $\Sigma^\alpha(\tilde u) = \{\sigma_i^\alpha\}_{i=1}^{r_\alpha}$ of $\tilde u$ for all $\alpha \in T$}
			\STATE{compute the subset of candidate nodes $\hat T$ defined by \eqref{eq:Tbar}}
			\STATE{compute the subset $T_m^\theta$ defined by \eqref{Tmtheta} and the corresponding rank $r^{m+1}$ defined by \eqref{update-ranks}, 
			with $\theta$ the highest value in $[0,\theta^\star]$ such that $r^{m+1}$ satisfies the necessary conditions of admissibility in~\cite[Section 2.3]{2015arXiv150503027F}}
		\end{algorithmic}	
	\end{algorithm}
 	
	\begin{remark}
		In practice, we observe that it is important to provide a good initialization for the learning algorithm \ref{alg:learning_als} used at Step \ref{alg:rank-optim-learning-step} of Algorithm \ref{alg:rank-optim} for the computation of $u^{m+1}$. In practice, we take as an initialization the truncation $\mathrm{Truncate}(\tilde u;T,r^{m+1})$ at rank $r^{m+1}$ of the approximation $\tilde u$ computed at Step \ref{step-v} of Algorithm \ref{alg:Tm}.
	\end{remark}

\subsection{Tree adaptation}
\begin{figure}[h]
		\centering
		\footnotesize
		\begin{subfigure}[b]{.32\textwidth}
			\centering
			\begin{tikzpicture}[level distance = 10mm]
				\tikzstyle{active}=[circle,fill=black]
				\tikzstyle{level 1}=[sibling distance=25mm]
				\tikzstyle{level 2}=[sibling distance=10mm]
				\node [active,label={above:{$\{1,2,3,4\}$}}]  {}
				child {node [active,label=above left:{$\{1,2\}$}]  {}
					child {node [active,label=below:{$\{1\}$}] {}}
					child {node [active,label=below:{$\{2\}$}] {}}
				}
				child {node [active,label=above right:{$\{3,4\}$}] {}
					child {node [active,label=below:{$\{3\}$}] {}}
					child {node [active,label=below:{$\{4\}$}] {}}
				};
			\end{tikzpicture}
			\caption{$T$}
			\label{fig:tree-permute-T}
		\end{subfigure}
		\begin{subfigure}[b]{.32\textwidth}
			\centering
			\begin{tikzpicture}[level distance = 10mm]
			\tikzstyle{active}=[circle,fill=black]
			\tikzstyle{level 1}=[sibling distance=20mm]
			\tikzstyle{level 2}=[sibling distance=10mm]
			\node [active,label={above:{$\{1,2,3,4\}$}}]  {}
			child {node [active,label=above left:{$\{1,3\}$}]  {}
				child {node [active,label=below:{$\{1\}$}] {}}
				child {node [active,label=below:{$\{3\}$}] {}}
			}
			child {node [active,label=above right:{$\{2,4\}$}] {}
				child {node [active,label=below:{$\{2\}$}] {}}
				child {node [active,label=below:{$\{4\}$}] {}}
			};
			\end{tikzpicture}
			\caption{$T'$}
			\label{fig:tree-permute-T'}
		\end{subfigure}
		\begin{subfigure}[b]{.32\textwidth}
			\centering
			\begin{tikzpicture}[level distance = 10mm]
			\tikzstyle{active}=[circle,fill=black]
			\tikzstyle{level 1}=[sibling distance=15mm]
			\tikzstyle{level 2}=[sibling distance=15mm]
			\node [active,label={above:{$\{1,2,3,4\}$}}]  {}
			child {node [active,label=above left:{$\{2,3,4\}$}]  {}
				child {node [active,label=above left:{$\{3,4\}$}] {}
					child {node [active,label=below:{$\{3\}$}] {}}
					child {node [active,label=below:{$\{4\}$}] {}}
				   }
				child {node [active,label=below:{$\{2\}$}] {}}
			}
			child {node [active,label=below:{$\{1\}$}]  {}}
;
			\end{tikzpicture}
			\caption{$T''$}
			\label{fig:tree-permute-T''}
		\end{subfigure}
		\caption{Dimension tree $T$ over $D = \{1,2,3,4\}$ (a), tree $T'$ obtained by permuting the nodes $\{2\}$ and $\{3\}$ of $T$ (b), tree $T''$ obtained by permuting nodes $\{1\}$ and $\{3,4\}$ of $T$ (c).}
		
		\label{fig:trees permute}
	\end{figure}
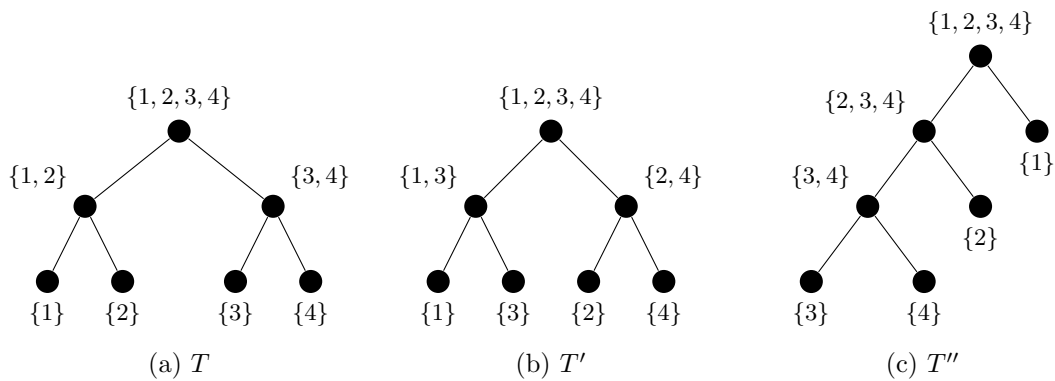

The approximation power of tree-based tensor formats strongly depends on the selection of the dimension tree. 
The number of possible dimension trees being exponential in the dimension $d$, the selection of an optimal tree is intractable in high dimension. 
We here propose a heuristic stochastic algorithm for the optimisation of the tree $T$, within a class of trees having the same arity. The proposed approach consists in comparing trees obtained by successive permutations of nodes drawn randomly according to a suitable probability distribution. Starting from a given tree, the proposed strategy allows the exploration of a very large class of trees obtained by an arbitrary number of permutations of nodes. In particular, starting from a binary tree, the strategy is able to explore the whole set of binary trees over $D$ (including all balanced and linear trees).
For example, a single permutation of nodes allows to go from the tree $T$ of Figure \ref{fig:tree-permute-T} to the tree $T'$ of Figure \ref{fig:tree-permute-T'}, with the same topology,  but also to the tree of Figure \ref{fig:tree-permute-T''} with a different topology.

\subsubsection{Tree optimization for the representation of a given function}
	Let $T$ be a given partition tree and consider a given function $v\in \Hc$.  
	Letting $r = \rank_{T}(v)$, we have that $v \in \Tc^{T}_{r}(\Hc)$ admits a representation	
	\begin{align}\label{eq: tree-based permute}
		v(x) = \sum_{\substack{1 \leq i_\alpha \leq N_\alpha \\ \alpha \in \Lc(T)}} \sum_{\substack{1 \leq k_\alpha \leq r_\alpha \\Â \alpha \in T}} \prod_{\alpha \in T\setminus \Lc(T)} C^\alpha_{(k_\beta)_{\beta \in S(\alpha)},k_\alpha} \prod_{\alpha \in \Lc(T)}  C^\alpha_{i_\alpha,k_\alpha} \phi^\alpha_{i_\alpha}(x_\alpha),
	\end{align}
	with storage complexity\footnote{Note that the possible sparsity of the representation in a given tensor format is not (but could be) taken into account 
	for defining an effective storage complexity.}
\begin{align*}
		C(T,r) = \sum_{\alpha \in T \setminus \Lc(T)} r_\alpha \prod_{\beta \in S(\alpha)} r_\beta + \sum_{\alpha \in \Lc(T)} N_\alpha r_\alpha.
	\end{align*}
	Then, we would like to find a tree $T$ solution of 
	\begin{align}\label{min_permute}
	\min_{T} C(T,\rank_{T}(v)), 
	\end{align}
	which minimizes over a set of dimension trees  the storage complexity for the  function $v$. In practice, when $v$ is an approximation of a target function, we may be only interested in obtaining an approximation of $v$ with a certain precision (e.g., related to an estimation  of the generalization error of $v$) and with minimal storage complexity. Then problem \eqref{min_permute} can be replaced by 
	\begin{align}\label{min_permute_epsilon}
	\min_{T } C(T,\rank^\epsilon_{T}(v)), 
	\end{align}
	with $\rank^\epsilon_T(v) = (\rank^\epsilon_\alpha(v))_{\alpha \in T}$, where the $\epsilon$-rank 
	$\rank^\epsilon_\alpha(v)$ is defined as the minimal integer $r_\alpha^\epsilon$ 
	such that there exists an approximation $v^\epsilon$ with $\rank_\alpha(v^\epsilon) = r^\epsilon_\alpha$ 
	and such that $\Vert v - v^\epsilon \Vert \le \epsilon \Vert v \Vert$.

	For solving \eqref{min_permute}, we propose a stochastic algorithm which successively compares the current tree $T$ with a new tree $\widetilde T$ drawn from a suitable probability distribution over the set of trees, obtained by successive random permutations of nodes, and accepts the tree $\widetilde T$ if it yields a lower storage complexity for $v$ (at relative precision $\epsilon$). The probability distribution, defined below, gives a higher probability to trees $\widetilde T$ presenting the highest potential reduction of storage complexity (by preferably permuting nodes $\alpha$ whose parents have the highest ranks $r_{P(\alpha)}$) but the lowest computational complexity for changing the representation of $v$ from $T$ to $\widetilde T$. 
	
Before presenting the stochastic algorithm	 and how to draw randomly a new tree $\widetilde T$,   we first detail how to change the representation of a function by a permutation of two nodes in a current tree $T$. This will allow us to introduce a notion of computational complexity	 for changes of representations.

\paragraph{Changing the representation by permutations of two nodes.}
	For two nodes $\nu$ and $\mu$ in a tree $T$ such that $\nu \cap \mu = \emptyset$  (i.e., one node is not the ascendant of the other), we denote by $\sigma_{\nu,\mu}$ the map such that $ \sigma_{\nu,\mu}(T)$ is the tree obtained from $T$ by a permutation of nodes $\nu$ and $\mu$. Let $P(\alpha;T)$, $S(\alpha;T)$, $A(\alpha;T)$ and $\level(\alpha;T)$ be the parent, children, ascendants and level of $\alpha$ in $T$, respectively. In the tree $\sigma_{\nu,\mu}(T)$, we have that $P(\nu;\sigma_{\nu,\mu}(T)) = (P(\mu;T) \setminus \mu) \cup \nu$ and $P(\mu;\sigma_{\nu,\mu}(T)) = (P(\nu;T) \setminus \nu) \cup \mu$.
	The map $\sigma_{\nu,\mu}$ only modifies the nodes of $T$ in $(A(\mu;T)\cup A(\nu;T))\setminus (A(\mu;T) \cap A(\nu;T))$, which is the set of all ascendants of $\nu$ or $\mu$ that are not common ascendants of these two nodes. In particular, we have that $ \sigma_{\nu,\mu}(T) = T$ if $P(\mu;T) = P(\nu;T).$ 
	Let \begin{align}\gamma :=\argmax_{\beta \in A(\nu;T) \cup A(\mu;T)} \level(\beta;T) \label{gamma}  \end{align} denote the highest-level common ascendant of $\nu$ and $\mu$ in $T$ and let 
	\begin{align}
	 	T^{\nu,\mu} = (A(\nu;T) \cup A(\mu;T)) \setminus (\{\gamma\} \cup A(\gamma;T)) \label{Tnumu}
	\end{align} be the subset of $T$ containing all the ascendants of $\nu$ and $\mu$ up to $\gamma$, except $\gamma$. The tree $\widetilde T =  \sigma_{\nu,\mu}(T) $ is such that $\widetilde T = (T\setminus T^{\nu,\mu} ) \cup \widetilde T^{\nu,\mu}$.
	
	The representation \eqref{eq: uFunOfPhiAndW} of $v$ can be written
	\begin{align*}
		v(x) = \sum_{\substack{1 \leq k_\beta \leq r_\beta \\ \beta \in S(T^{\nu,\mu})}} \sum_{1 \leq k_\gamma \leq r_\gamma} M^{\gamma}_{(k_\beta)_{\beta \in S(T^{\nu,\mu})},k_\gamma}
		\prod_{\beta \in S(T^{\nu,\mu})} u_{k_\beta}^\beta(x_\beta)
		w^{\gamma}_{k_\gamma}(x_{\gamma^c})
	\end{align*}
	with $S(T^{\nu,\mu}) = \{ \alpha \in T \setminus T^{\nu,\mu} : \alpha \in S(\beta;T), \beta \in T^{\nu,\mu} \}$, and	where
	\begin{align} \label{eq: tensor-tilde-C-gamma}
		M^{\gamma}_{(k_\beta)_{\beta \in S(T^{\nu,\mu})},k_\gamma} = \sum_{\substack{1 \leq k_\alpha \leq r_\alpha \\ \alpha \in T^{\nu,\mu} }} \prod_{\alpha \in T^{\nu,\mu} \cup \{\gamma\}} C^{\alpha}_{(k_\eta)_{\eta \in S(\alpha;T)},k_\alpha}
	\end{align}
	are the components of a tensor $M^{\gamma}$ of order $\#S(T^{\nu,\mu}) + 1$. Assuming that $v$ has a $\gamma$-orthogonal representation \eqref{eq: tree-based permute}, the functions 
	\begin{align*}
		\left\{ \prod_{\beta \in S(T^{\nu,\mu})} u_{k_\beta}^\beta(x_\beta) \right\}_{\substack{1 \leq k_\beta \leq r_\beta \\ \beta \in S(T^{\nu,\mu})}}
		\quad \text{and} \quad 
		\left\{ w^{\gamma}_{k_\gamma}(x_{\gamma^c}) \right\}_{1 \leq k_\gamma \leq r_\gamma}
	\end{align*}
	are orthonormal in $\Hc_\gamma$ and $\Hc_{\gamma^c}$ respectively, and we have $\|M^{\gamma} \| = \| v \|$. Up to a change in the ordering of the children of each node, the expression \eqref{eq: tensor-tilde-C-gamma} can be rewritten, introducing the parents $\gamma^\nu = P(\nu ; T)$ and $\gamma^\mu = P(\mu ; T)$ of $\nu$ and $\mu$ in $T$, as
	\begin{align}\label{eq: tensor-tilde-C-gamma-2}
		M^{\gamma}_{(k_\beta)_{\beta \in S(T^{\nu,\mu})},k_\gamma} = \sum_{\substack{1 \leq k_\alpha \leq r_\alpha \\ \alpha \in T^{\nu,\mu} }} C^{\gamma^\nu}_{(k_\eta)_{\eta \in S(\gamma^\nu;T) \setminus \{\nu\}},k_\nu,k_{\gamma^\nu}}
		C^{\gamma^\mu}_{(k_\eta)_{\eta \in S(\gamma^\mu;T) \setminus \{\mu\}},k_\mu,k_{\gamma^\mu}} \notag &\\
		\prod_{\alpha \in \{\gamma\} \cup (T^{\nu,\mu} \setminus \{ \gamma^\nu,\gamma^\mu \})} 
		C^{\alpha}_{(k_\eta)_{\eta \in S(\alpha;T)},k_\alpha} .&
	\end{align}
	Also, the tensor $M^{\gamma}$ can be identified with
	\begin{align}\label{eq: tensor-tilde-C-gamma-3}
		M^{\gamma}_{(k_\beta)_{\beta \in S(T_{\tilde\sigma}^{\nu,\mu})},k_\gamma} = \sum_{\substack{1 \leq k_\alpha \leq \tilde r_\alpha \\ \alpha \in T^{\nu,\mu}  }} \widetilde C^{\gamma^\nu}_{(k_\eta)_{\eta \in S(\gamma^\nu;T) \setminus \{\nu\}},k_\mu,k_{\gamma^\nu}}
		\widetilde C^{\gamma^\mu}_{(k_\eta)_{\eta \in S(\gamma^\mu;T) \setminus \{\mu\}},k_\nu,k_{\gamma^\mu}} & \notag\\
		\prod_{\alpha \in  \{\gamma\} \cup (T^{\nu,\mu} \setminus \{ \gamma^\nu,\gamma^\mu \})} 
		\widetilde C^{\alpha}_{(k_\eta)_{\eta \in S(\alpha;T)},k_\alpha} &
		,
	\end{align}
	where $\widetilde C^{\gamma^\nu}$ now has an index $k_\mu$, and $\widetilde C^{\gamma^\mu}$ an index $k_\nu$. The representation \eqref{eq: tensor-tilde-C-gamma-3} of $M^{\gamma}$ corresponds to a representation of $v$ in $\Tc^{\widetilde T}_{\tilde r}(\Hc)$ with $\widetilde T = \sigma_{\nu,\mu}(T)$ and $\tilde r_\beta=r_\beta$ for all $\beta \in   T \setminus T^{\nu,\mu}$.
	
	Algorithm \ref{alg: permutation} presents the permutation procedure, showing how to practically obtain an approximation in $\Tc^{\widetilde T}_{ \tilde r}(\Hc)$ of $v \in \Tc^{T}_{r}(\Hc)$ with relative precision $\epsilon$.
	At step \ref{alg:step_SVD}, the singular value decomposition of a matricization of $M^\gamma_{(k_\beta)_{\beta \in S_\gamma},k_\gamma}$ is computed:
	\begin{align}\label{eq: matricization-svd}
		M^\gamma_{(k_\beta)_{\beta \in S_\gamma},k_\gamma} = \sum\limits_{k_\eta = 1}^{\tilde r_\eta} \tilde\sigma_{k_\eta} \tilde a_{(k_\beta)_{\beta \in S_\gamma \setminus S(\eta;\widetilde T)},k_\eta,k_\gamma} \tilde b_{(k_\beta)_{\beta \in S(\eta;\widetilde T)},k_\eta},
	\end{align}
	with singular values $\{\tilde\sigma_{k_\eta}\}_{k_\eta = 1}^{\tilde r_\eta}$. We define $\tilde r_\eta^{\epsilon'}$ as the minimal integer ensuring 
	\begin{align}\label{epsilon-ranks}
	\sum_{k_{\eta}=\tilde r_\eta^{\epsilon'}+1}^{\tilde r_\eta} \tilde \sigma_{k_\eta}^2 \le {\epsilon'}^2 \sum_{k_{\eta}=1}^{\tilde r_\eta} \tilde \sigma_{k_\eta}^2
	\end{align} 
	with ${\epsilon'}^2={\epsilon^2} / {\#T^{\nu,\mu}}$, where $\# T^{\nu,\mu}$ is the number of singular value decompositions required for changing the representation of a function from tree $T$ to tree $\sigma_{\nu,\mu}(T)$. 
	The integers $(\tilde r_\alpha^{\epsilon'})_{\alpha \in \widetilde T}$ correspond to the $\epsilon'$-ranks of $v$ in the format associated with the tree $\sigma_{\nu,\mu}(T)$.
	By truncating \eqref{eq: matricization-svd} at rank $\tilde r^{\epsilon'}_\eta$, we obtain an approximation of $v$ with relative precision $\epsilon^\prime$, ultimately leading to an approximation of $v$ with relative precision $\epsilon$ once the $\# T^{\nu,\mu}$ singular value decompositions are performed.
	
	\begin{algorithm}\caption{Change of representation of a tree-based tensor $v \in \Tc^T_r$ by a permutation of two nodes $\nu$ and $\mu$ in $T$ at precision $\epsilon$.}
		\label{alg: permutation}
	\begin{algorithmic}[1]
			\REQUIRE initial $T$, representation of the tensor $v \in \Tc^{T}_{r}(\Hc)$ with $T$-rank $r$, nodes 
			$\nu$ and $\mu$ to be permuted, such that $\nu \cap \mu = \emptyset$, relative precision $\epsilon$.
		 \ENSURE approximation of the tensor $v$ in $\Tc^{\widetilde T}_{ \tilde r}(\Hc)$ with $\widetilde T = \sigma_{\nu,\mu}(T)$ and 
		$ \widetilde T$-rank $\tilde r$
			\STATE{Compute $\gamma$ defined by \eqref{gamma}  and  $T^{\nu,\mu}$ defined by \eqref{Tnumu}}
			 \STATE{perform a $\gamma$-orthogonalization of the representation of $v$ using Algorithm \ref{alg: alpha-orth}}
			\STATE{$M^\gamma \leftarrow C^\gamma$, $S_\gamma \leftarrow S(\gamma;T)$}
			\FOR{$\eta \in T^{\nu,\mu}$ by increasing level}
				\STATE{$M^\gamma_{(k_\beta)_{\beta \in S_\gamma \setminus \{\eta\}},(k_\beta)_{\beta \in S(\eta;T)},k_\gamma} \leftarrow \sum\limits_{k_{\eta} = 1}^{r_{\eta}} M^\gamma_{(k_\beta)_{\beta \in S_\gamma},k_\gamma} C^{\eta}_{(k_\beta)_{\beta \in S({\eta;T})},k_{\eta}}$}
				\STATE{$S_\gamma \leftarrow (S_\gamma \setminus \{\eta\}) \cup S(\eta;T)$}
			\ENDFOR
			\STATE{$\widetilde T \leftarrow \sigma_{\nu,\mu}(T)$}
			\FOR{$\eta \in \widetilde T^{\nu,\mu}$ by decreasing level}
				\STATE compute the SVD \eqref{eq: matricization-svd} of the matricization $ \Mc_{\{i^\gamma_\beta\}_{\beta \in S(\eta;\widetilde T)}}(M^\gamma)^T$ of $M^\gamma$\label{alg:step_SVD}
				\STATE truncate the SVD at rank $\tilde r^{\epsilon'}_{\eta}$, the minimal integer satisfying \eqref{epsilon-ranks}
				\STATE $\tilde M^\gamma_{(k_\beta)_{\beta \in S_\gamma \setminus S(\eta;\widetilde T)},k_\eta,k_\gamma} \leftarrow \tilde\sigma_{k_\eta} \tilde a_{(k_\beta)_{\beta \in S_\gamma \setminus S(\eta;\widetilde T)},k_\eta,k_\gamma}$, $k_\eta = 1,\hdots,\tilde r^{\epsilon'}_{\eta}$
				\STATE $\widetilde C^\eta_{(k_\beta)_{\beta \in S(\eta;\widetilde T)},k_\eta} \leftarrow \tilde b_{(k_\beta)_{\beta \in S(\eta;\widetilde T)},k_\eta}$, $k_\eta = 1,\hdots,\tilde r^{\epsilon'}_{\eta}$
				\STATE{$\tilde r_\eta \leftarrow \tilde r^{\epsilon'}_{\eta}$}
				\STATE{$M^\gamma \leftarrow \tilde M^\gamma$}
				\STATE{$S_\gamma \leftarrow (S_\gamma \cup \{\eta\}) \setminus S(\eta;\widetilde T)$}
			\ENDFOR
			
			\STATE{$\widetilde C^\gamma \leftarrow M^\gamma$, $\tilde r_\gamma \leftarrow r_\gamma$, $\widetilde C^\beta \leftarrow C^\beta, \tilde r_\beta \leftarrow r_\beta, \forall \beta \in \widetilde T\setminus \widetilde T^{\nu,\mu}$ with $\beta \neq \gamma$}
			\STATE{return the approximation of the tensor $v$ in $ \Tc^{\widetilde T}_{\tilde r}(\Hc)$ given by $\sum\limits_{\substack{1 \leq k_\alpha \leq \tilde r_\alpha \\ \alpha \in \widetilde T}} \prod\limits_{\alpha \in \widetilde T \setminus \Lc(\widetilde T)} \widetilde C^\alpha_{(k_\beta)_{\beta \in S(\alpha;\widetilde T)},k_\alpha} \prod\limits_{\alpha \in \Lc(\widetilde T)} u^\alpha_{k_\alpha} (x_\alpha)$.}
		\end{algorithmic}
	\end{algorithm}

\paragraph{Stochastic algorithm for tree optimization.}
To describe the stochastic algorithm for tree optimization, it remains to detail how to draw a new tree $\widetilde T$ given a current tree $T$.
The tree $\widetilde T$  is defined as
$
\widetilde T = \sigma_m \circ \hdots \circ \sigma_1 (T),
$
where the map $\sigma_i  = \sigma_{\nu_i,\mu_i}$ permutes two nodes of the tree $T_{i-1} = \sigma_{i-1} \circ \hdots \circ \sigma_1(T),$ with $T_0 := T.$
The number of permutations $m$ is first drawn according to the distribution 
\begin{align}\Pbb(m = k) \propto k^{-\gamma_1}, \quad k\in \Nbb^*, \label{draw-m}
\end{align}
with $\gamma_1>0$, which gives a higher probability to low numbers of permutations. 
Then, given $T_{i}$, $\sigma_{i+1} = \sigma_{\nu_{i+1},\mu_{i+1}}$ is defined by first 
 drawing  randomly the node $\nu_{i+1}$ in $T_{i} \setminus \{D\}$ according to the distribution 
 \begin{align}
 \Pbb(\nu_{i+1} = \alpha \vert T_i) \propto \rank_{P(\alpha;T_i)}(v)^{\gamma_2}, \quad \alpha\in T_{i}\setminus \{D\}, \label{draw-first-node}
 \end{align} with $P(\alpha;T_i)$ the parent of $\alpha$ in $T_{i}$ and $\gamma_2>0$. This gives a higher probability to select a node $\nu_{i+1}$ with a high parent's rank (note that $P(\nu_{i+1};T_i)$ will not be in the next tree $T_{i+1}$). Then given the first node $\nu_{i+1}$, we draw the second node $\mu_{i+1}$ in $T_i$ according to the distribution 
 \begin{align}\Pbb(\mu_{i+1} = \alpha \vert T_i , \nu_{i+1}) \propto  
 \begin{cases} 
 d_{T_i}(\nu_{i+1},\alpha)^{-\gamma_3} & \text{if } \alpha \cap \nu_{i+1} = \emptyset\\
 0 & \text{otherwise}\end{cases},\label{draw-second-node}\end{align}
where $\gamma_3>0$ and 
\begin{align}
 d_{T_i}(\nu,\mu) = r_\gamma \prod_{\beta \in S(T_i^{\nu,\mu}) } r_\beta \label{pseudo-metric}
\end{align}
is the storage complexity of the full tensor $M^\gamma$ of order $\#S(T_{i}^{\nu,\mu})+1$ (defined in equation \eqref{eq: tensor-tilde-C-gamma} with $T$ replaced by $T_i$) which is computed 
when changing the representation of the function $v$ from the tree $T_i$ to the tree $ \sigma_{\nu,\mu}(T_i)$.
 This choice of probability distribution therefore gives a higher probability to modifications of the tree with low computational  complexity.
	The stochastic algorithm for solving \eqref{min_permute_epsilon}  is finally given in Algorithm \ref{alg: tree-optim}. 	
	\begin{remark}
The parameters $\gamma_1,\gamma_2,\gamma_3$ of the above probability distributions have an impact on the computational complexity and the ability to try large modifications of the current tree. In practice, we choose $\gamma_1 = \gamma_2 = \gamma_3 = 2$ in the numerical experiments but the choice of these parameters should deserve a deeper analysis.
\end{remark}
	\begin{algorithm}\caption{Tree optimisation algorithm for the representation of a given tensor $v$ at precision $\epsilon$.}\label{alg: tree-optim}
		\begin{algorithmic}[1]
			\REQUIRE initial tree $T$, representation of the tensor $v \in \Tc^{T}_{r}(\Hc)$ with $T$-rank $r$ and storage complexity $C(T,r)$, precision $\epsilon$, number of random trials $N$
			\ENSURE new tree $T^\star$ and approximation $v^\star$ of $v$ in  $\Tc^{T^\star}_{r^ \star}(\Hc)$ with new $T^\star$-rank $r^\star$ and storage complexity $C(T^\star, r^\star) \leq C(T,r)$
			\STATE{$C^\star \leftarrow C(T,r)$, $T^\star \leftarrow T$, $v^\star \leftarrow v$, $\widetilde T \leftarrow T$, $\tilde r \leftarrow r$, $m \leftarrow 0$}
			\STATE{$\Sigma \leftarrow \emptyset$, $newtree = false$}
			\FOR{$k = 1,\hdots,N$}
			\STATE{$m_{\text{old}} \leftarrow m$}
			\STATE{draw randomly $m$ drawn according to \eqref{draw-m}}
			\IF{$m > m_{\text{old}}$ or $newtree$}
				\STATE{compute an approximation $v_0$ of $v$ by applying successively all permutations in $\Sigma$ using Algorithm \ref{alg: permutation} with precision $\epsilon/(m+\#\Sigma)$}
				\STATE{$T_0 \leftarrow T^\star$}
			\ENDIF
			\FOR{$i = 0,\hdots,m$}
				\STATE{draw $\sigma_{i+1} = \sigma_{\nu_{i+1},\mu_{i+1}}$ according to \eqref{draw-first-node} and \eqref{draw-second-node} (with $T$ replaced by $T_{i}$)}
				\STATE{$T_{i+1} \leftarrow \sigma_{i+1}(T_{i})$}
				\STATE{compute an approximation $v_{i+1}$ of $v_{i}$ by applying the permutation $\sigma_{i+1}$ using Algorithm \ref{alg: permutation} with precision $\epsilon/(m+\#\Sigma)$}
			\ENDFOR
			\STATE{$\widetilde T \leftarrow T_m$, $\tilde v \leftarrow v_m$, $\tilde r \leftarrow \rank_{\widetilde T}(\tilde v)$}
			\IF{$C(\widetilde T,\tilde r) < C^\star$}
				\STATE{$newtree = true$}
				\STATE{$T^\star \leftarrow \widetilde T$, $C^\star \leftarrow C(\widetilde T,\tilde r)$, $v^\star \leftarrow \tilde v$}
				\STATE{$\Sigma \leftarrow \Sigma \cup \{\sigma_1,\hdots,\sigma_m\}$ (ordered set)}
				\ELSE{}
				\STATE{$newtree = false$}
			\ENDIF
			\ENDFOR
		\end{algorithmic}	
	\end{algorithm}

\subsubsection{Learning scheme with rank and tree adaptation}
	Algorithm \ref{alg:rank-tree-adaptive} describes a global algorithm for the approximation in tree-based tensor format.	
	It is similar to Algorithm \ref{alg:rank-optim}, with an additional step of tree adaptation using Algorithm \ref{alg: tree-optim}.
	The algorithm stops when reaching a maximum number of iterations $M$, or when at least one of the following stopping criteria is met:
	\begin{align*}
		 \Rc_{V}(u^{m}) &\leq \varepsilon_{\text{goal}} \Rc_V(0),\\
		 \Rc_{V}(u^{m}) &\geq \tau_{\text{overfit}} \min_{1 \leq i \leq m-1} \Rc_{V}(u^{i})
	\end{align*}
	where $V$ is a validation set independent of $S$. For the recovery of functions in the noiseless case, $\varepsilon_{\text{goal}}$ is taken of the order of the machine precision. The second criterion is met once overfitting occurs (in numerical experiments, we take $\varepsilon_{\text{overfit}} = 10$). 
	 
	\begin{algorithm}[h]\caption{Learning scheme with rank and tree adaptation.}\label{alg:rank-tree-adaptive}
	    \begin{algorithmic}[1]
	        \REQUIRE sample $\{(z_i)\}_{i=1}^n$, contrast function, initial tree $T$
	        \ENSURE new tree  $ T$ and approximation $u_r$ in $\Tc_{r}^{ T}(\Hc)$ with $ T$-rank $r$
	        \STATE compute an approximation $u^1$ with $T$-rank $r^1 = (1,\hdots,1)$ using Algorithm \ref{alg:learning_als}
	        \STATE{$m \leftarrow 1$}
	        \WHILE{the stopping criterion is not met}
	        \STATE compute $T_m^\theta$ with Algorithm \ref{alg:Tm} and corresponding $r^{m+1}$ defined by \eqref{update-ranks}
	        \STATE compute an approximation $u^{m+1}$ in $\Tc_{r^{m+1}}^{T}(\Hc)$ using Algorithm \ref{alg:learning_als}
	        \STATE using Algorithm \ref{alg: tree-optim}, search for a new tree $\widetilde T$ to obtain an approximation $\tilde u$ in $\Tc_{\tilde r^{m+1}}^{\widetilde T}(\Hc)$ of $u^{m+1}$ with reduced storage complexity
	        \IF{$C(\widetilde T,\tilde r^{m+1}) < C(T,r^{m+1})$}
	        	\STATE{$u^{m+1} \leftarrow \tilde u$}
	        	\STATE{compute an approximation $\hat u$ in $\Tc_{\tilde r^{m+1}}^{\widetilde T}(\Hc)$ of $u$ using Algorithm \ref{alg:learning_als} with initialization $u^{m+1}$}
				\STATE{$T \leftarrow \widetilde T$, $u^{m+2} \leftarrow \hat u$, $r^{m+2} \leftarrow \tilde r^{m+1}$}
	        	\STATE{$m \leftarrow m+1$}
	        \ENDIF
	        \STATE{$m \leftarrow m+1$}
	        \ENDWHILE
	        \STATE{select $m_{opt} = \arg\min_{1\le j \le m} \Rc_{V}(u^{j})$, where $V$ is a validation set independent of $S$, and return $u_r = u^{m_{opt}}$ \label{alg:modelSelectionStep2}}
	    \end{algorithmic}	
	\end{algorithm}

\clearpage

\section{Numerical experiments}\label{sec:experiments}
This section presents numerical experiments in a supervised learning setting where, for given samples $S = \{(x_i,y_i)\}_{i=1}^n$ of a pair of random variables $(X,Y)$, we try to find an approximation $v(X)$ of $Y$ using tree-based tensor formats. We assume that $Y = u(X) + \varepsilon$  
where $X = (X_1,\hdots,X_d)$ has a known probability law $\mu$ over $\Xc = \Rbb^d$, where $u \in L^2_\mu(\Xc)$, and where the noise $\varepsilon$ is a random variable, independent of $X$, with finite variance. 
In all examples, we consider a least-squares regression setting by choosing a contrast function $\gamma(v,(x,y)) = (v(x)-y)^2$.

We use Algorithm \ref{alg:rank-tree-adaptive} for the approximation in tree-based format $\Tc_r^T(\Hc)$ with adaptation of both the $T$-rank $r$ and the dimension tree $T$ over $D=\{1,\dots,d\}$. For the spaces $\Hc_\nu$ ($\nu \in D$), we choose polynomial spaces $\Pbb_{p}(\Xc^\nu)$ of degree $p$ and we use orthonormal polynomial bases $\{\phi_{i_\nu}^{\nu}\}_{i_\nu \in I^\nu}$ in $L^2_{\mu_{\nu}}$, with $I^\nu = \{0,\hdots,p\}$. We exploit sparsity for the representation of one-dimensional functions in the bases $\{\phi_{i_\nu}^{\nu}\}_{i_\nu \in I^\nu}$, $\nu\in D$, or equivalently, in the tensors $C^\alpha  \in \Rbb^{I_\alpha \times \{1,\hdots,r_\alpha\}}$, with $\alpha \in \Lc(T)$. For the estimation of sparse approximations,  we use the strategy described in Section \ref{sec:exploit-sparsity} which consists in estimating a succession of approximations with increasing polynomial degree, and then in selecting the optimal approximation based on an estimation of the error using corrected leave-one out estimators (see Example \ref{ex:loo-ls}
 and Remark \ref{corrected-estimator}). We also use this cross-validation error estimator as the tolerance $\epsilon$ for the permutations in the tree optimisation Algorithm \ref{alg: tree-optim}. We set $N=100$ in Algorithm \ref{alg: tree-optim} and let the maximum number of iterations in Algorithm \ref{alg:rank-tree-adaptive} be sufficiently high.
 
In all examples, we estimate the generalization ($L^2$) error using a  test sample $S_{\text{test}}$ of size $ 10000$, independent of $S$. The  relative test error $\varepsilon(S_{\text{test}},v)$ associated with the function $v$ and test sample $S_{\text{test}}$ is defined as 
	\begin{equation*}
		\varepsilon(S_{\text{test}},v)^2 = \frac{\sum_{(x,y) \in S_{\text{test}}} (y - v(x))^2}{\sum_{(x,y) \in S_{\text{test}}} y^2}.
	\end{equation*}
For studying the robustness of the algorithm, for each example, we run 10 times the algorithm, each run using a different training sample $S$ and a different test sample $S_{\text{test}}$, and starting from a tree drawn randomly in the set of trees with a given arity. This allows us to provide ranges for the obtained quantities of interest (errors  and complexities). We also present an estimation $\hat \Pbb(T \text{ is optimal})$ of the probability of obtaining an approximation in $\Tc^{T}_{r}(\Hc)$ with a dimension tree $T$ optimal in a sense specified in each example, obtained by counting, out of the 10 runs, how many times the algorithm returned an optimal tree.\footnote{Knowing that the number of times the algorithm returns an optimal tree out of 10 trials follows a binomial distribution of parameters a probability $p$ and the number of trials 10, we can propose a lower bound of  $p$ with confidence level $1-\alpha$ by using the Clopper-Pearson confidence interval. For example, with a level $1 - \alpha = 0.95$, $\hat \Pbb(T \text{ is optimal}) = 0.9$ leads to $\Pbb(p \geq 0.55) = 0.95$, and $\hat \Pbb(T \text{ is optimal}) = 1$ leads to $\Pbb(p \geq 0.69) = 0.95$.}
	
	\subsection{An anisotropic multivariate function}\label{subsec:multivariateFunction}
		We consider the following function in dimension $d = 6$:
		\begin{equation} \tag{\ensuremath{i}} 
		\label{eq:multiVariateFunction}
			u(X) = \frac{1}{(10+2X_1+X_3+2X_4-X_5)^2}
		\end{equation}
		where the random variables $X = (X_1,\hdots,X_6)$ are uniform on $[-1,1]$, and we consider a noise $\varepsilon = 0$. We choose polynomial spaces $\Hc_\nu = \Pbb_{10}(\Xc_\nu)$.		
	We use Algorithm \ref{alg:rank-tree-adaptive} to obtain an approximation of Function \eqref{eq:multiVariateFunction}  in $\Tc^{T}_{r}(\Hc)$, with adapted tree $T$ and tree-based rank $r$.
		
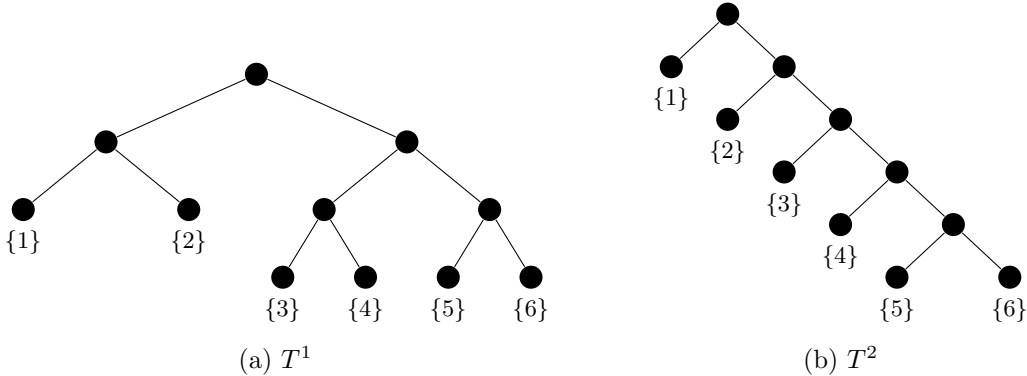
\begin{figure}[h]
		\footnotesize
		\centering
			\begin{subfigure}[b]{.49\textwidth}
				\centering
				\begin{tikzpicture}[level distance = 9mm]
				\tikzstyle{active}=[circle,fill=black]
				\tikzstyle{level 1}=[sibling distance=40mm]
				\tikzstyle{level 2}=[sibling distance=22mm]
				\tikzstyle{level 3}=[sibling distance=11mm]
				\node [active,label={above:{}}]  {}
				child {node [active,label=above left:{}]  {}
					child {node [active,label=below:{$\{1\}$}] {}}
					child {node [active,label=below:{$\{2\}$}] {}}
				}
				child {node [active,label=above right:{}] {}
					child {node [active,label=above left:{}] {}
						child {node [active,label=below:{$\{3\}$}] {}}
						child {node [active,label=below:{$\{4\}$}] {}}
				}
					child {node [active,label=above right:{}] {}
					child {node [active,label=below:{$\{5\}$}] {}}
					child {node [active,label=below:{$\{6\}$}] {}}
				}
				};
				\end{tikzpicture}
				\caption{$T^1$}
				\label{fig:multivariateFunctionTTTTree_0a}
			\end{subfigure}
			\begin{subfigure}[b]{.49\textwidth}
				\centering
				\begin{tikzpicture}[level distance = 7mm]
				\tikzstyle{active}=[circle,fill=black]
				\tikzstyle{level 1}=[sibling distance=15mm]
				\node [active,label={above:{}}] {}
				child {node [active,label=below:{$\{1\}$}] {}
				}
				child {node [active,label=above right:{}] {}
					child {node [active,label=below:{$\{2\}$}] {}}
					child {node [active,label=above right:{}] {}
						child {node [active,label=below:{$\{3\}$}] {}}
						child {node [active,label=above right:{}] {}
							child {node [active,label=below:{$\{4\}$}] {}}
							child {node [active,label=above right:{}] {}
								child {node [active,label=below:{$\{5\}$}] {}}
								child {node [active,label=below:{$\{6\}$}] {}}
							}
						}
					}
				};
				\end{tikzpicture}
				\caption{$T^2$}
				\label{fig:multivariateFunctionTTTTree_0b}
			\end{subfigure}
			\caption{Two particular dimension trees over $D=\{1,\dots,6\}$ yielding two different families of trees obtained by permutations.}
			\label{fig:multivariateFunctionTTTTree_0}
		\end{figure}

		The tree $T$ is selected in the family of trees of arity 2, starting from two different families of trees $T^1_\sigma = \{\sigma(\alpha) : \alpha \in T^1\}$ and $T^2_\sigma = \{\sigma(\alpha) : \alpha \in T^2\}$, with $\sigma$ a permutation of $D$ and where the trees $T^1$ and $T^2$ are represented on Figures  \ref{fig:multivariateFunctionTTTTree_0a} and \ref{fig:multivariateFunctionTTTTree_0b} respectively.
		
		
		Table \ref{tab:multivariateFunction} summarizes the results. We observe that we obtain with high probability a very accurate approximation with only a small training sample, and a fast decrease of the error with the training sample size $n$.
		We also notice that with very high probability, the algorithm finds a tree $T$ containing the node $\{1,3,4,5\}$ (associated with the only variables on which $Y$ depends) and almost only increases the ranks associated with the nodes involving the dimensions $1$, $3$, $4$ and $5$. We also notice that, when the training sample size is large enough, the cross-validation error is a good estimator of the generalization error, enabling model selection in Step \ref{alg:modelSelectionStep} of Algorithm \ref{alg:rank-optim} and Step \ref{alg:modelSelectionStep2} of Algorithm \ref{alg:rank-tree-adaptive}, without the need of an independent test sample.

		\begin{table}[h]
			\centering
			\begin{tabular}{cccccc}\toprule
				$T_\sigma$ & $n$ & $\hat \Pbb(\{1,3,4,5\} \in T)$ & $\varepsilon(S_{\text{test}},v)$ & CV error & $C(T,r)$ \\ \midrule
				\multirow{3}{*}{$T^1_\sigma$}& $10^2$ & $90\%$ & $[2.32 \, 10^{-3} , 8.36 \, 10^{-3}]$ & $[1.08 \, 10^{-4} , 1.34 \, 10^{-3}]$ & $[132 , 255]$\\
				& $10^3$ & $100\%$ & $[6.93 \, 10^{-6} , 4.21 \, 10^{-5}]$ & $[3.11 \, 10^{-8} , 8.49 \, 10^{-7}]$ & $[407 , 1338]$\\
				& $10^4$ & $100\%$ & $[3.13 \, 10^{-8} , 4.47 \, 10^{-6}]$ & $[3.64 \, 10^{-9} , 1.64 \, 10^{-6}]$ & $[376 , 881]$\\ \midrule
				\multirow{3}{*}{$T^2_\sigma$}& $10^2$ & $90\%$ & $[1.18 \, 10^{-3} , 8.23 \, 10^{-3}]$ & $[3.67 \, 10^{-5} , 5.54 \, 10^{-4}]$ & $[132 , 266]$\\
				& $10^3$ & $90\%$ & $[2.23 \, 10^{-6} , 3.73 \, 10^{-5}]$ & $[2.24 \, 10^{-8} , 8.74 \, 10^{-7}]$ & $[374 , 1182]$\\
				& $10^4$ & $100\%$ & $[2.06 \, 10^{-8} , 2.06 \, 10^{-6}]$ & $[1.76 \, 10^{-9} , 1.82 \, 10^{-6}]$ & $[344 , 1403]$\\
				\bottomrule
			\end{tabular}
			\caption{Results for Function \eqref{eq:multiVariateFunction} starting from two families of trees $T^1_\sigma$ and $T^2_\sigma$: training sample size $n$, estimation of the probability of having $\{1,3,4,5\} \in T$, and ranges (over the 10 trials) for the test error, the cross-validation (CV) error estimator and the storage complexity.}
			\label{tab:multivariateFunction}
		\end{table}
				
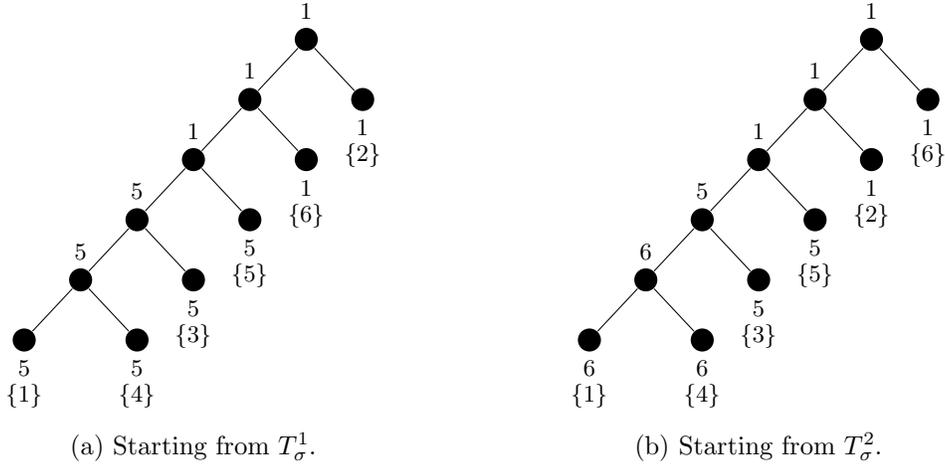
\begin{figure}[H]
			\centering
			\footnotesize
			\begin{subfigure}[b]{.49\textwidth}
				\centering
				\begin{tikzpicture}[level distance = 8mm]
				    \tikzstyle{active}=[circle,fill=black]
				    \tikzstyle{level 1}=[sibling distance=15mm]
				    \node [active,label=above:{$1$}]  {} 
				    child{node [active,label=above:{$1$}] {}
				    	child{node [active,label=above:{$1$}] {}
				    		child{node [active,label=above:{$5$}] {}
				    			child{node [active,label=above:{$5$}] {}
				    				child{node [active,label=below:{$\stackon{\{1\}}{5}$}] {}}
				    				child{node [active,label=below:{$\stackon{\{4\}}{5}$}] {}}}
				    			child{node [active,label=below:{$\stackon{\{3\}}{5}$}] {}}}
				    		child{node [active,label=below:{$\stackon{\{5\}}{5}$}] {}}}
				    	child{node [active,label=below:{$\stackon{\{6\}}{1}$}] {}}}
				    child{node [active,label=below:{$\stackon{\{2\}}{1}$}] {}};
				\end{tikzpicture}
				\caption{Starting from $T^1_\sigma$.}
				\label{fig:multivariateFunctionBinaryTree}
			\end{subfigure}
			\begin{subfigure}[b]{.49\textwidth}
				\centering
			    \begin{tikzpicture}[level distance = 8mm]
				    \tikzstyle{active}=[circle,fill=black]
				    \tikzstyle{level 1}=[sibling distance=15mm]   
				    \node [active,label=above:{$1$}]  {} 
				    child{node [active,label=above:{$1$}] {}
				    	child{node [active,label=above:{$1$}] {}
				    		child{node [active,label=above:{$5$}] {}
				    			child{node [active,label=above:{$6$}] {}
				    				child{node [active,label=below:{$\stackon{\{1\}}{6}$}] {}}
				    				child{node [active,label=below:{$\stackon{\{4\}}{6}$}] {}}}
				    			child{node [active,label=below:{$\stackon{\{3\}}{5}$}] {}}
				    		}
				    		child{node [active,label=below:{$\stackon{\{5\}}{5}$}] {}}}
				    	child{node [active,label=below:{$\stackon{\{2\}}{1}$}] {}}
				    	}
				    child{node [active,label=below:{$\stackon{\{6\}}{1}$}] {}};
			    \end{tikzpicture}
			    \caption{Starting from $T^2_\sigma$.}
			    \label{fig:multivariateFunctionTTTTree}
			\end{subfigure}
			\caption{Examples of trees $T$ obtained using Algorithm \ref{alg:rank-tree-adaptive} on the same training sample of size $10^4$, starting from a tree of the first family (a) or the second family (b). The obtained $\alpha$-ranks are indicated at each node and the dimensions associated with the leaf nodes are displayed in brackets.}
			\label{fig:treesMultivariateFunction}
		\end{figure}

		Figure \ref{fig:treesMultivariateFunction} presents  examples of trees and associated tree-based ranks obtained by the algorithm, starting 
		from a tree of one or the other family of trees. We observe that the algorithm modified the structure of the starting tree in order to reduce the storage complexity of the representation by isolating the dimensions 2 and 6 from the other dimensions. Furthermore, by studying the function \eqref{eq:multiVariateFunction}, we notice that the variables $X_1$ and $X_4$ have the same influence on the output of the function, which explains why they appear grouped in the shown trees. A similar conclusion can be drawn for the variables $X_3$ and $X_5$.
		
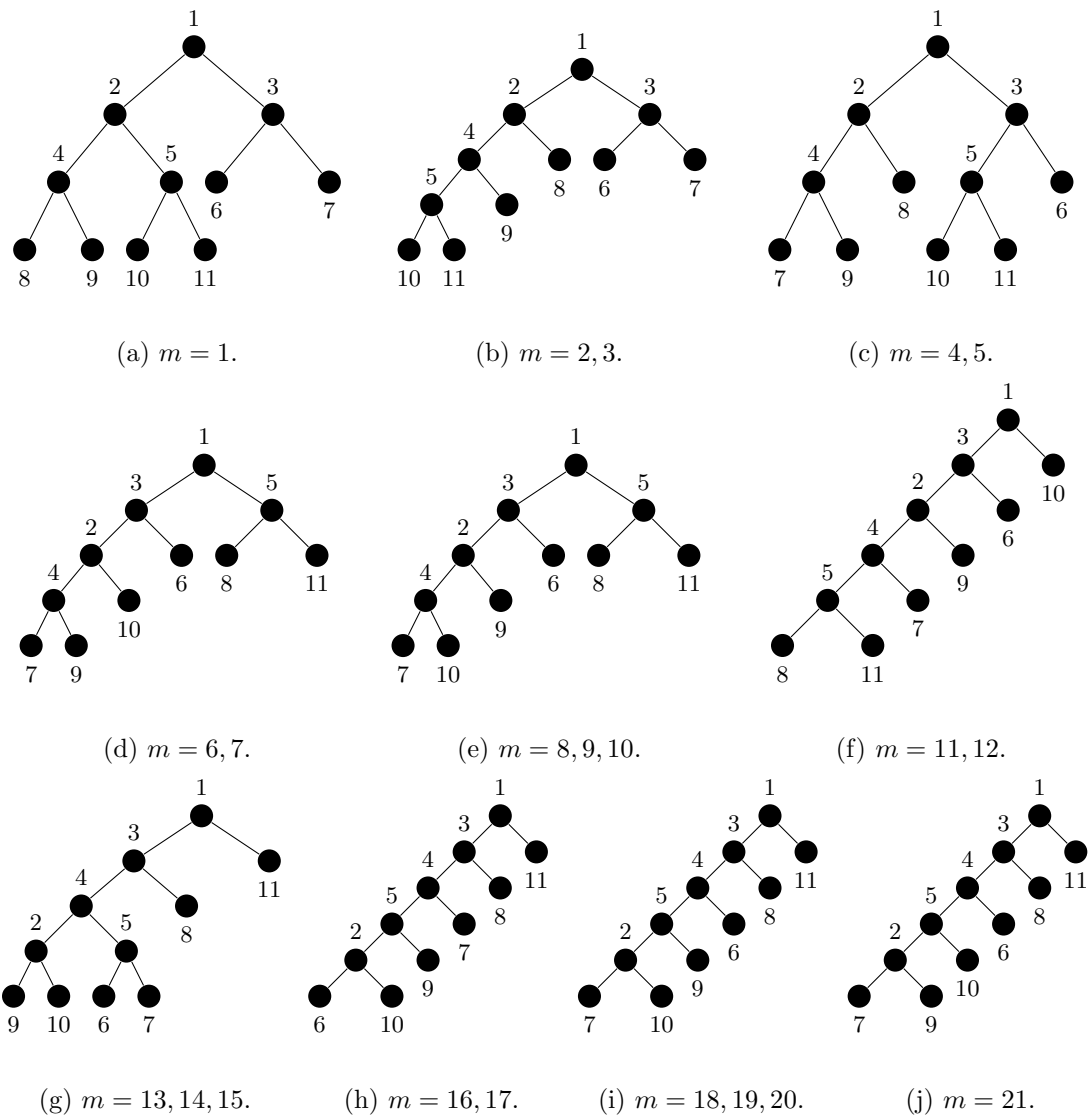
\begin{figure}[H]
			\centering
			\footnotesize
				\begin{subfigure}[b]{.32\textwidth}
					\centering
					\begin{tikzpicture}[scale=0.6]  \tikzstyle{level 1}=[sibling distance=35mm]  \tikzstyle{level 2}=[sibling distance=25mm]  \tikzstyle{level 3}=[sibling distance=15mm]  \node [active,label=above:{$1$}]  {} child{node [active,label=above:{$2$}] {}child{node [active,label=above:{$4$}] {}child{node [active,label=below:{$8$}] {}}child{node [active,label=below:{$9$}] {}}}child{node [active,label=above:{$5$}] {}child{node [active,label=below:{$10$}] {}}child{node [active,label=below:{$11$}] {}}}}child{node [active,label=above:{$3$}] {}child{node [active,label=below:{$6$}] {}}child{node [active,label=below:{$7$}] {}}};\end{tikzpicture}
					\caption{$m = 1$.}
					\label{fig:effective1}
				\end{subfigure}
				\begin{subfigure}[b]{.32\textwidth}
					\centering
					\begin{tikzpicture}[scale=0.4]  \tikzstyle{level 1}=[sibling distance=45mm]  \tikzstyle{level 2}=[sibling distance=30mm]  \tikzstyle{level 3}=[sibling distance=25mm]  \tikzstyle{level 4}=[sibling distance=15mm]  \node [active,label=above:{$1$}]  {} child{node [active,label=above:{$2$}] {}child{node [active,label=above:{$4$}] {}child{node [active,label=above:{$5$}] {}child{node [active,label=below:{$10$}] {}}child{node [active,label=below:{$11$}] {}}}child{node [active,label=below:{$9$}] {}}}child{node [active,label=below:{$8$}] {}}}child{node [active,label=above:{$3$}] {}child{node [active,label=below:{$6$}] {}}child{node [active,label=below:{$7$}] {}}};\end{tikzpicture}
					\caption{$m = 2,3$.}
					\label{fig:effective2}
				\end{subfigure}
				\begin{subfigure}[b]{.32\textwidth}
					\centering
					\begin{tikzpicture}[scale=0.6]  \tikzstyle{level 1}=[sibling distance=35mm]  \tikzstyle{level 2}=[sibling distance=20mm]  \tikzstyle{level 3}=[sibling distance=15mm]  \node [active,label=above:{$1$}]  {} child{node [active,label=above:{$2$}] {}child{node [active,label=above:{$4$}] {}child{node [active,label=below:{$7$}] {}}child{node [active,label=below:{$9$}] {}}}child{node [active,label=below:{$8$}] {}}}child{node [active,label=above:{$3$}] {}child{node [active,label=above:{$5$}] {}child{node [active,label=below:{$10$}] {}}child{node [active,label=below:{$11$}] {}}}child{node [active,label=below:{$6$}] {}}};\end{tikzpicture}
					\caption{$m = 4,5$.}
					\label{fig:effective3}
				\end{subfigure}
				\begin{subfigure}[b]{.32\textwidth}
					\centering
					\begin{tikzpicture}[scale=0.4]  \tikzstyle{level 1}=[sibling distance=45mm]  \tikzstyle{level 2}=[sibling distance=30mm]  \tikzstyle{level 3}=[sibling distance=25mm]  \tikzstyle{level 4}=[sibling distance=15mm]  \node [active,label=above:{$1$}]  {} child{node [active,label=above:{$3$}] {}child{node [active,label=above:{$2$}] {}child{node [active,label=above:{$4$}] {}child{node [active,label=below:{$7$}] {}}child{node [active,label=below:{$9$}] {}}}child{node [active,label=below:{$10$}] {}}}child{node [active,label=below:{$6$}] {}}}child{node [active,label=above:{$5$}] {}child{node [active,label=below:{$8$}] {}}child{node [active,label=below:{$11$}] {}}};\end{tikzpicture}
					\caption{$m = 6,7$.}
					\label{fig:effective4}
				\end{subfigure}
				\begin{subfigure}[b]{.32\textwidth}
					\centering
					\begin{tikzpicture}[scale=0.4]  \tikzstyle{level 1}=[sibling distance=45mm]  \tikzstyle{level 2}=[sibling distance=30mm]  \tikzstyle{level 3}=[sibling distance=25mm]  \tikzstyle{level 4}=[sibling distance=15mm]  \node [active,label=above:{$1$}]  {} child{node [active,label=above:{$3$}] {}child{node [active,label=above:{$2$}] {}child{node [active,label=above:{$4$}] {}child{node [active,label=below:{$7$}] {}}child{node [active,label=below:{$10$}] {}}}child{node [active,label=below:{$9$}] {}}}child{node [active,label=below:{$6$}] {}}}child{node [active,label=above:{$5$}] {}child{node [active,label=below:{$8$}] {}}child{node [active,label=below:{$11$}] {}}};\end{tikzpicture}
					\caption{$m = 8,9,10$.}
					\label{fig:effective5}
				\end{subfigure}
				\begin{subfigure}[b]{.32\textwidth}
					\centering
					\begin{tikzpicture}[scale=0.4]  \tikzstyle{level 1}=[sibling distance=30mm]  \tikzstyle{level 2}=[sibling distance=30mm]  \tikzstyle{level 3}=[sibling distance=30mm]  \tikzstyle{level 4}=[sibling distance=30mm]  \tikzstyle{level 5}=[sibling distance=30mm]  \node [active,label=above:{$1$}]  {} child{node [active,label=above:{$3$}] {}child{node [active,label=above:{$2$}] {}child{node [active,label=above:{$4$}] {}child{node [active,label=above:{$5$}] {}child{node [active,label=below:{$8$}] {}}child{node [active,label=below:{$11$}] {}}}child{node [active,label=below:{$7$}] {}}}child{node [active,label=below:{$9$}] {}}}child{node [active,label=below:{$6$}] {}}}child{node [active,label=below:{$10$}] {}};\end{tikzpicture}
					\caption{$m = 11,12$.}
					\label{fig:effective6}
				\end{subfigure}
				\begin{subfigure}[b]{.26\textwidth}
					\centering
					\begin{tikzpicture}[scale=0.4]  \tikzstyle{level 1}=[sibling distance=45mm]  \tikzstyle{level 2}=[sibling distance=35mm]  \tikzstyle{level 3}=[sibling distance=30mm]  \tikzstyle{level 4}=[sibling distance=15mm]  \node [active,label=above:{$1$}]  {} child{node [active,label=above:{$3$}] {}child{node [active,label=above:{$4$}] {}child{node [active,label=above:{$2$}] {}child{node [active,label=below:{$9$}] {}}child{node [active,label=below:{$10$}] {}}}child{node [active,label=above:{$5$}] {}child{node [active,label=below:{$6$}] {}}child{node [active,label=below:{$7$}] {}}}}child{node [active,label=below:{$8$}] {}}}child{node [active,label=below:{$11$}] {}};\end{tikzpicture}
					\caption{$m = 13,14,15$.}
					\label{fig:effective7}
				\end{subfigure}
				\begin{subfigure}[b]{.23\textwidth}
					\centering
					\begin{tikzpicture}[scale=0.32]  \tikzstyle{level 1}=[sibling distance=30mm]  \tikzstyle{level 2}=[sibling distance=30mm]  \tikzstyle{level 3}=[sibling distance=30mm]  \tikzstyle{level 4}=[sibling distance=30mm]  \tikzstyle{level 5}=[sibling distance=30mm]  \node [active,label=above:{$1$}]  {} child{node [active,label=above:{$3$}] {}child{node [active,label=above:{$4$}] {}child{node [active,label=above:{$5$}] {}child{node [active,label=above:{$2$}] {}child{node [active,label=below:{$6$}] {}}child{node [active,label=below:{$10$}] {}}}child{node [active,label=below:{$9$}] {}}}child{node [active,label=below:{$7$}] {}}}child{node [active,label=below:{$8$}] {}}}child{node [active,label=below:{$11$}] {}};\end{tikzpicture}
					\caption{$m = 16,17$.}
					\label{fig:effective8}
				\end{subfigure}
				\begin{subfigure}[b]{.23\textwidth}
					\centering
					\begin{tikzpicture}[scale=0.32]  \tikzstyle{level 1}=[sibling distance=30mm]  \tikzstyle{level 2}=[sibling distance=30mm]  \tikzstyle{level 3}=[sibling distance=30mm]  \tikzstyle{level 4}=[sibling distance=30mm]  \tikzstyle{level 5}=[sibling distance=30mm]  \node [active,label=above:{$1$}]  {} child{node [active,label=above:{$3$}] {}child{node [active,label=above:{$4$}] {}child{node [active,label=above:{$5$}] {}child{node [active,label=above:{$2$}] {}child{node [active,label=below:{$7$}] {}}child{node [active,label=below:{$10$}] {}}}child{node [active,label=below:{$9$}] {}}}child{node [active,label=below:{$6$}] {}}}child{node [active,label=below:{$8$}] {}}}child{node [active,label=below:{$11$}] {}};\end{tikzpicture}
					\caption{$m = 18,19,20$.}
					\label{fig:effective9}
				\end{subfigure}
				\begin{subfigure}[b]{.23\textwidth}
					\centering
					\begin{tikzpicture}[scale=0.32]  \tikzstyle{level 1}=[sibling distance=30mm]  \tikzstyle{level 2}=[sibling distance=30mm]  \tikzstyle{level 3}=[sibling distance=30mm]  \tikzstyle{level 4}=[sibling distance=30mm]  \tikzstyle{level 5}=[sibling distance=30mm]  \node [active,label=above:{$1$}]  {} child{node [active,label=above:{$3$}] {}child{node [active,label=above:{$4$}] {}child{node [active,label=above:{$5$}] {}child{node [active,label=above:{$2$}] {}child{node [active,label=below:{$7$}] {}}child{node [active,label=below:{$9$}] {}}}child{node [active,label=below:{$10$}] {}}}child{node [active,label=below:{$6$}] {}}}child{node [active,label=below:{$8$}] {}}}child{node [active,label=below:{$11$}] {}};\end{tikzpicture}
					\caption{$m = 21$.}
					\label{fig:effective10}
				\end{subfigure}
			\caption{Dimension trees associated to the iteration number $m$ in the Table \ref{tab:adaptationMultivariateFunction}, with each node numbered. The singletons $\{1\},\{2\},\{3\},\{4\},\{5\},\{6\}$ correspond to the leaf nodes numbered $7,11,10,9,6,8$ respectively.}
			\label{fig:effective}
			\end{figure}

		\paragraph{Illustration of the behavior of Algorithm \ref{alg:rank-tree-adaptive}.}
			Table \ref{tab:adaptationMultivariateFunction} illustrates the behavior of a single run of Algorithm \ref{alg:rank-tree-adaptive} for the construction of an approximation of the function \eqref{eq:multiVariateFunction} using a training sample of size $n = 10000$ and starting from the tree \ref{fig:effective1}. The trees returned by the algorithm at each iteration are displayed in Figure \ref{fig:effective}. We observe that the algorithm presents a fast convergence and yields a very accurate approximation after $21$ iterations (and 9 adaptations of the tree).
			
			\begin{table}[H]
				\centering
					\begin{tabular}{ccccc}\toprule
						$m$ & Tree $T$ & Tree-based rank $r^m$ & $\varepsilon(S_{\text{test}},v)$ & $C(T,r^m)$ \\ \midrule
						$1$ & Fig. \ref{fig:effective1} & $(1, 1, 1, 1, 1, 1, 1, 1, 1, 1, 1)$ & $4.88\, 10^{-2}$ & $71$ \\ \midrule
						$2$ & \multirow{2}{*}{Fig. \ref{fig:effective2}} & $(1, 1, 1, 1, 1, 1, 1, 1, 1, 1, 1)$ & $4.88\, 10^{-2}$ & $71$ \\ 
						$3$ & & $(1, 1, 1, 1, 1, 1, 1, 1, 1, 1, 1)$ & $4.88\, 10^{-2}$ & $71$ \\ \midrule
						$4$ & \multirow{2}{*}{Fig. \ref{fig:effective3}} & $(1, 1, 1, 1, 1, 1, 2, 1, 2, 1, 1)$ & $3.81\, 10^{-2}$ & $96$ \\ 
						$5$ & & $(1, 1, 1, 1, 1, 1, 2, 1, 2, 1, 1)$ & $3.81\, 10^{-2}$ & $96$ \\ \midrule
						$6$ & \multirow{2}{*}{Fig. \ref{fig:effective4}} & $(1, 2, 1, 2, 1, 2, 2, 1, 2, 2, 1)$ & $1.95\, 10^{-3}$ & $132$ \\ 
						$7$ & & $(1, 2, 1, 2, 1, 2, 2, 1, 2, 2, 1)$ & $1.95\, 10^{-3}$ & $132$ \\ \midrule
						$8$ & \multirow{3}{*}{Fig. \ref{fig:effective5}} & $(1, 2, 1, 3, 1, 2, 3, 1, 3, 2, 1)$ & $8.50\, 10^{-4}$ & $174$ \\ 
						$9$ & & $(1, 2, 1, 3, 1, 2, 3, 1, 3, 2, 1)$ & $8.49\, 10^{-4}$ & $174$ \\ 
						$10$ & & $(1, 3, 1, 3, 1, 3, 3, 1, 3, 3, 1)$ & $6.53\, 10^{-5}$ & $219$ \\ \midrule
						$11$ & \multirow{2}{*}{Fig. \ref{fig:effective6}} & $(1, 3, 3, 3, 1, 3, 3, 1, 3, 3, 1)$ & $5.88\, 10^{-5}$ & $227$ \\ 
						$12$ & & $(1, 3, 3, 3, 1, 3, 3, 1, 3, 3, 1)$ & $6.08\, 10^{-5}$ & $227$ \\ \midrule
						$13$ & \multirow{3}{*}{Fig. \ref{fig:effective7}} & $(1, 4, 1, 1, 4, 3, 4, 1, 4, 3, 1)$ & $1.86\, 10^{-5}$ & $290$ \\ 
						$14$ & & $(1, 4, 1, 1, 4, 3, 4, 1, 4, 3, 1)$ & $1.86\, 10^{-5}$ & $290$ \\ 
						$15$ & & $(1, 4, 1, 1, 4, 4, 4, 1, 4, 4, 1)$ & $2.05\, 10^{-6}$ & $344$ \\ \midrule
						$16$ & \multirow{2}{*}{Fig. \ref{fig:effective8}} & $(1, 5, 1, 1, 4, 4, 4, 1, 4, 4, 1)$ & $1.56\, 10^{-6}$ & $376$ \\ 
						$17$ & & $(1, 5, 1, 1, 4, 4, 4, 1, 4, 4, 1)$ & $1.54\, 10^{-6}$ & $376$ \\ \midrule
						$18$ & \multirow{3}{*}{Fig. \ref{fig:effective9}} & $(1, 5, 1, 1, 4, 4, 5, 1, 5, 4, 1)$ & $5.03\, 10^{-7}$ & $438$ \\ 
						$19$ & & $(1, 5, 1, 1, 4, 4, 5, 1, 5, 4, 1)$ & $4.59\, 10^{-7}$ & $438$ \\ 
						$20$ & & $(1, 5, 1, 1, 5, 5, 5, 1, 5, 5, 1)$ & $6.71\, 10^{-8}$ & $519$ \\ \midrule
						$21$ & Fig. \ref{fig:effective10} & $(1, 5, 1, 1, 5, 5, 5, 1, 5, 5, 1)$ & $4.88\, 10^{-8}$ & $519$ \\ 
						\bottomrule
					\end{tabular}
				\caption{Behavior of Algorithm \ref{alg:rank-tree-adaptive} for the approximation of the function \eqref{eq:multiVariateFunction}, with $n = 10000$ and starting from the dimension tree $T^1_\sigma$ of Figure \ref{fig:effective1}. The node numbers can be seen for each tree on Figure \ref{fig:effective}.}
				\label{tab:adaptationMultivariateFunction}
			\end{table}
						
	\subsection{Sum of bivariate functions}\label{subsec:sumOfBivariateFunctions}

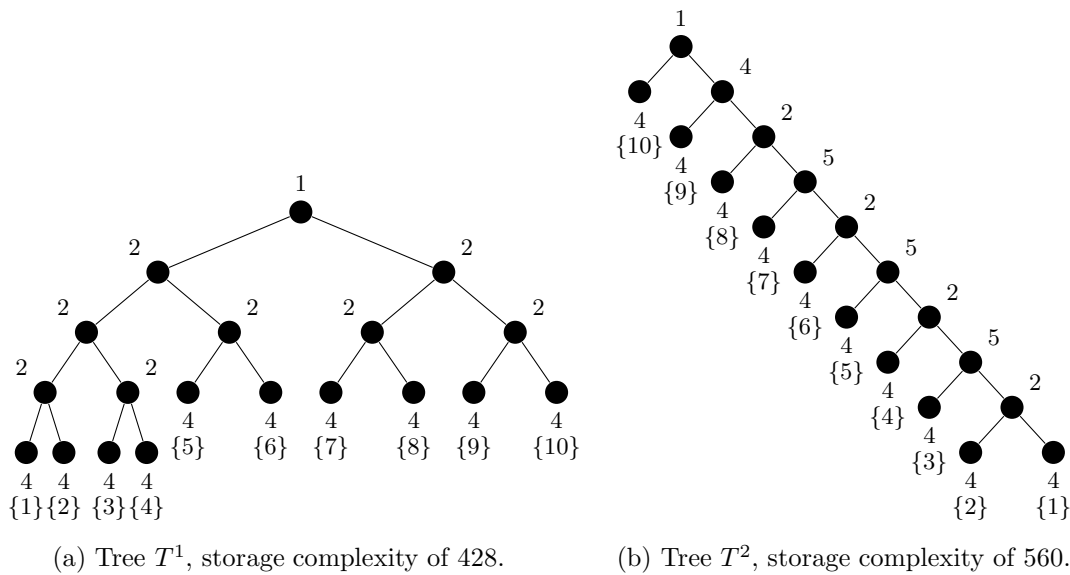
\begin{figure}[h]
			\centering
			\footnotesize
			\begin{subfigure}[b]{.49\textwidth}
				\centering
				\begin{tikzpicture}[level distance = 8mm]
				\tikzstyle{active}=[circle,fill=black]
				\tikzstyle{level 1}=[sibling distance=38mm]
				\tikzstyle{level 2}=[sibling distance=19mm]
				\tikzstyle{level 3}=[sibling distance=11mm]
				\tikzstyle{level 4}=[sibling distance=5mm]
				\node [active,label={above:{$1$}}]  {}
				child {node [active,label=above left:{$2$}]  {}
					child {node [active,label=above left:{$2$}] {}
						child {node [active,label=above left:{$2$}] {}
							child {node [active,label=below:{$\stackon{\{1\}}{4}$}] {}}
							child {node [active,label=below:{$\stackon{\{2\}}{4}$}] {}}
						}
						child {node [active,label=above right:{$2$}] {}
							child {node [active,label=below:{$\stackon{\{3\}}{4}$}] {}}
							child {node [active,label=below:{$\stackon{\{4\}}{4}$}] {}}
						}
					}
					child {node [active,label=above right:{$2$}] {}
						child {node [active,label=below:{$\stackon{\{5\}}{4}$}] {}}
						child {node [active,label=below:{$\stackon{\{6\}}{4}$}] {}}
					}
				}
				child {node [active,label=above right:{$2$}]  {}
					child {node [active,label=above left:{$2$}] {}
						child {node [active,label=below:{$\stackon{\{7\}}{4}$}] {}}
						child {node [active,label=below:{$\stackon{\{8\}}{4}$}] {}}
					}
					child {node [active,label=above right:{$2$}] {}
						child {node [active,label=below:{$\stackon{\{9\}}{4}$}] {}}
						child {node [active,label=below:{$\stackon{\{10\}}{4}$}] {}}
					}
				};
				\end{tikzpicture}
				\caption{Tree $T^1$, storage complexity of $428$.}
				\label{fig:sumOfBivariateFunctionsBinaryTree}
			\end{subfigure}
			\begin{subfigure}[b]{.49\textwidth}
				\centering
				\begin{tikzpicture}[level distance = 6mm]
				\tikzstyle{level 1}=[sibling distance=11mm]
				\tikzstyle{active}=[circle,fill=black]
				\node [active,label={above:{$1$}}]  {}
				child {node [active,label=below:{$\stackon{\{10\}}{4}$}]  {}}
				child {node [active,label=above right:{$4$}]  {}
					child {node [active,label=below:{$\stackon{\{9\}}{4}$}]  {}}
					child {node [active,label=above right:{$2$}]  {}
						child {node [active,label=below:{$\stackon{\{8\}}{4}$}]  {}}
						child {node [active,label=above right:{$5$}]  {}
							child {node [active,label=below:{$\stackon{\{7\}}{4}$}]  {}}
							child {node [active,label=above right:{$2$}]  {}
								child {node [active,label=below:{$\stackon{\{6\}}{4}$}]  {}}
								child {node [active,label=above right:{$5$}]  {}
									child {node [active,label=below:{$\stackon{\{5\}}{4}$}]  {}}
									child {node [active,label=above right:{$2$}]  {}
										child {node [active,label=below:{$\stackon{\{4\}}{4}$}]  {}}
										child {node [active,label=above right:{$5$}]  {}
											child {node [active,label=below:{$\stackon{\{3\}}{4}$}]  {}}
											child {node [active,label=above right:{$2$}]  {}
												child {node [active,label=below:{$\stackon{\{2\}}{4}$}]  {}}
												child {node [active,label=below:{$\stackon{\{1\}}{4}$}]  {}}	
											}
										}
									}
								}
							}
						}
					}
				};
				\end{tikzpicture}
				\caption{Tree $T^2$, storage complexity of $560$.}
				\label{fig:sumOfBivariateFunctionsLinearTree}
			\end{subfigure}
			\caption{Exact representations of Function \eqref{eq:sumOfBivariateFunctions} using two different dimension trees: the dimensions associated with the leaf nodes are displayed in brackets and the $\alpha$-ranks are indicated at each node.}
			\label{fig:sumOfBivariateFunctionsTrees}
		\end{figure}

		We here consider a sum of bivariate functions
		\begin{equation} \tag{\ensuremath{ii}}
		\label{eq:sumOfBivariateFunctions}
			u(X)= g(X_1,X_2) + g(X_3,X_4) + \hdots + g(X_{d-1},X_d),
		\end{equation}
		with $g(X_{\nu-1},X_{\nu})=\sum_{i=0}^m X_{\nu-1}^i X_{\nu}^i$ and where the random variables $X_\nu$ are independent and uniform on $[-1,1]$. 
		The problem addressed here is the recovery of the function using few samples and using Algorithm \ref{alg:rank-tree-adaptive} with tree adaptation. 
		We choose $d=10$ and $m=3$ and we consider $\Hc = \bigotimes_{\nu=1}^d \Pbb_{5}(\Xc^\nu)$, so that $u\in \Hc$ (no discretization errors). Figures \ref{fig:sumOfBivariateFunctionsBinaryTree} and \ref{fig:sumOfBivariateFunctionsLinearTree} present the tree-based ranks and storage complexity for the exact representation of the function using two different dimension trees $T^1$ (balanced tree) and $T^2$ (linear tree). We observe that the function can be represented in both formats with a moderate storage complexity. When running the algorithm with tree adaptation, the tree $T$ is selected in the family of trees of arity 2, starting from two different families of trees $T^1_\sigma = \{\sigma(\alpha) : \alpha \in T^1\}$ and $T^2_\sigma = \{\sigma(\alpha) : \alpha \in T^2\}$, with $\sigma$ a permutation of $D$.
		

		Given the structure of the function \eqref{eq:sumOfBivariateFunctions}, a tree $T$ will be said to be optimal if $\{k,k+1\} \in T$ for all odd $k \in \{1,\hdots,d-1\}$. As we will see below, this optimal tree is recovered with high probability by the proposed algorithm when large enough training sets are used.
		
	\paragraph{Illustration in the noiseless case.} 
	Table \ref{tab:sumOfBivariateFunctions} summarizes the obtained results in the noiseless case $\varepsilon = 0$. 
	We first observe that with a training sample large enough, with high probability, the algorithm is able to recover the function $u$ at machine precision with an optimal tree, this probability being higher when we use the family of balanced trees rather than the family of linear trees.
	For $n = 10^4$, all the obtained approximations use an optimal tree, which can be shown to lead to the smallest storage complexity for an exact representation of the function $u$.
		
	We also notice the importance of the tree adaptation: the highest obtained errors correspond to non-optimal trees, whereas the machine precision errors are obtained with optimal trees.

	\begin{table}[h]
			\centering
			\begin{tabular}{ccccccc}\toprule
				$T_\sigma$ & $n$ & $\hat \Pbb(T \text{ is optimal})$ & $\varepsilon(S_{\text{test}},v)$ & CV error & $C(T,r)$ \\ \midrule
				\multirow{3}{*}{$T^1_\sigma$}& $5 \, 10^2$ & $50\%$ & $[4.23 \, 10^{-15} , 1.80 \, 10^{-1}]$ & $[7.75 \, 10^{-16} , 1.64 \, 10^{-1}]$ & $[84 , 921]$\\
				& $10^3$ & $100\%$ & $[6.64 \, 10^{-16} , 9.60 \, 10^{-15}]$ & $[5.91 \, 10^{-16} , 1.84 \, 10^{-15}]$ & $[428 , 673]$\\
				& $10^4$ & $100\%$ & $[5.34 \, 10^{-16} , 1.18 \, 10^{-15}]$ & $[5.24 \, 10^{-16} , 1.18 \, 10^{-15}]$ & $[428 , 428]$\\ \midrule
				\multirow{3}{*}{$T^2_\sigma$}& $5 \, 10^2$ & $70\%$ & $[5.83 \, 10^{-15} , 1.94 \, 10^{-1}]$ & $[8.87 \, 10^{-16} , 1.88 \, 10^{-1}]$ & $[69 , 1114]$ \\
				& $10^3$ & $90\%$ & $[7.72 \, 10^{-16} , 2.43 \, 10^{-2}]$ & $[6.61 \, 10^{-16} , 1.87 \, 10^{-2}]$ & $[357 , 515]$\\
				& $10^4$ & $100\%$ & $[5.59 \, 10^{-16} , 1.74 \, 10^{-15}]$ & $[5.55 \, 10^{-16} , 1.75 \, 10^{-15}]$ & $[428 , 428]$ \\ 
				\bottomrule
			\end{tabular}
			\caption{Results for the function \eqref{eq:sumOfBivariateFunctions}: training sample size $n$, starting from two families of trees $T^1_\sigma$ and $T^2_\sigma$, estimation of the probability of obtaining an optimal tree and ranges (over the 10 trials) for the test error, the cross-validation (CV) error estimator and the storage complexity.}
			\label{tab:sumOfBivariateFunctions}
		\end{table}

		Table \ref{tab:sumOfBivariateFunctionsNoTreeAdaptation} shows the influence of the tree adaptation for a training sample size of $n = 1000$. We notice that the tree adaptation enables the recovery of the function \eqref{eq:sumOfBivariateFunctions}, whereas without tree adaptation, the obtained error is high, not going below $10^{-3}$. Even if there exists an exact representation of the function $u$ whatever the chosen tree, this example shows that the tree adaptation is essential when only a few samples are available.

		\begin{table}[h]
			\centering
			\begin{tabular}{ccccc}\toprule
				$T_\sigma$ & Tree adaptation & $\varepsilon(S_{\text{test}},v)$ & CV error & $C(T,r)$ \\ \midrule
				\multirow{2}{*}{$T^1_\sigma$}& Yes & $[6.64 \, 10^{-16} , 9.60 \, 10^{-15}]$ & $[5.91 \, 10^{-16} , 1.84 \, 10^{-15}]$ & $[428 , 673]$\\
				& No & $[9.31 \, 10^{-3} , 1.25 \, 10^{-1}]$ & $[4.68 \, 10^{-3} , 1.13 \, 10^{-1}]$ & $[184 , 786]$\\ \midrule
				\multirow{2}{*}{$T^2_\sigma$}& Yes & $[7.72 \, 10^{-16} , 2.43 \, 10^{-2}]$ & $[6.61 \, 10^{-16} , 1.87 \, 10^{-2}]$ & $[357 , 515]$\\
				& No & $[9.89 \, 10^{-3} , 9.69 \, 10^{-2}]$ & $[4.77 \, 10^{-3} , 8.55 \, 10^{-2}]$ & $[221 , 728]$ \\
				\bottomrule
			\end{tabular}
			\caption{Results for the function \eqref{eq:sumOfBivariateFunctions} with $n = 1000$, with and without tree adaptation and starting from the tree $T^1_\sigma$ or $T^2_\sigma$ with a random initial permutation $\sigma$: ranges (over the 10 trials) for the test error, the cross-validation (CV) error and the storage complexity.}
			\label{tab:sumOfBivariateFunctionsNoTreeAdaptation}
		\end{table}
	
	\paragraph{Illustration in the noisy case.}
	We now consider a noisy case $Y = u(X) + \varepsilon$ where $\varepsilon \sim \Nc(0,\zeta^2)$ is independent of $X$, and a training sample $S = \{ (x_i,y_i)\}_{i=1}^n$ with 
$y_i =  u(x_i) + \varepsilon_i$, where  the $\varepsilon_i$ are i.i.d. realizations of $\varepsilon$.
	
	Table \ref{tab:sumOfBivariateFunctionsNoisyCase} shows the obtained results for different training sample sizes and standard deviations of the noise. We observe that, except with the smallest sample and largest noise standard deviations, the algorithm is able to recover with high probability an optimal tree grouping consecutive input variables. We also notice that the cross-validation error is not always a good estimator of the generalization error for noisy observations. This can be an issue for  model selection: at the last step of Algorithm \ref{alg:rank-tree-adaptive}, the selected model is the one leading to the smallest  risk, estimated using a validation set $V$ independent of the training set S. In many practical cases, one does not have access to a validation set, and must then rely on other estimators of the generalization error, such as a cross-validation estimator. Hence, the model selected using a cross-validation estimator might not be the one minimizing the generalization error.

	\begin{table}[h]
		\centering
		\begin{tabular}{cccccc}\toprule
			$n$ & $\zeta$ & $\hat \Pbb(T \text{ is optimal})$ & $\varepsilon(S_{\text{test}},v)$ & CV error & $C(T,r)$ \\ \midrule
			\multirow{3}{*}{$5 \, 10^2$} & $10^{-1}$ & $50\%$ & $[3.26 \, 10^{-2} , 9.78 \, 10^{-2}]$ & $[2.30 \, 10^{-2} , 9.06 \, 10^{-2}]$ & $[142 , 271]$\\
			 & $10^{-2}$ & $80\%$ & $[3.64 \, 10^{-3} , 1.35 \, 10^{-2}]$ & $[1.35 \, 10^{-3} , 6.25 \, 10^{-3}]$ & $[298 , 518]$\\
			 & $10^{-3}$ & $80\%$ & $[1.27 \, 10^{-4} , 1.52 \, 10^{-1}]$ & $[1.34 \, 10^{-4} , 1.16 \, 10^{-1}]$ & $[114 , 515]$\\ \midrule
			 \multirow{3}{*}{$10^3$} & $10^{-1}$ & $30\%$ & $[1.00 \, 10^{-2} , 1.23 \, 10^{-1}]$ & $[1.65 \, 10^{-2} , 9.43 \, 10^{-2}]$ & $[117 , 399]$\\
			 & $10^{-2}$ & $80\%$ & $[7.52 \, 10^{-4} , 6.19 \, 10^{-3}]$ & $[1.55 \, 10^{-3} , 3.75 \, 10^{-3}]$ & $[428 , 546]$\\
			 & $10^{-3}$ & $100\%$ & $[7.63 \, 10^{-5} , 1.04 \, 10^{-4}]$ & $[1.62 \, 10^{-4} , 1.73 \, 10^{-4}]$ & $[428 , 468]$\\ \midrule
			 \multirow{3}{*}{$10^4$} & $10^{-1}$ & $90\%$ & $[2.17 \, 10^{-3} , 1.65 \, 10^{-2}]$ & $[1.70 \, 10^{-2} , 1.90 \, 10^{-2}]$ & $[282 , 631]$\\
			 & $10^{-2}$ & $90\%$ & $[2.08 \, 10^{-4} , 2.31 \, 10^{-4}]$ & $[1.71 \, 10^{-3} , 1.79 \, 10^{-3}]$ & $[428 , 527]$\\
			 & $10^{-3}$ & $100\%$ & $[2.09 \, 10^{-5} , 2.40 \, 10^{-5}]$ & $[1.71 \, 10^{-4} , 1.79 \, 10^{-4}]$ & $[428 , 528]$\\
			\bottomrule
		\end{tabular}
		\caption{Results for the function \eqref{eq:sumOfBivariateFunctions} in the noisy case: training sample size $n$, standard deviation of the noise, estimation of the probability of obtaining an optimal tree and ranges (over the 10 trials) for the test error, the cross-validation (CV) error estimator and the storage complexity.}
		\label{tab:sumOfBivariateFunctionsNoisyCase}
	\end{table}

	We recall that the risk  $\Rc(v) = \Rc(u) + \|u-v\|^2$, with $\Rc(u) = \Ebb((Y - u(X))^2) = \Ebb(\epsilon^2) = \zeta^2$. Then $\Rc(v)$ is the sum of the squared approximation error $\Vert u-v\Vert^2$ and of the variance of the noise. Table \ref{tab:sumOfBivariateFunctionsNoisyCase2} presents, for different values of $n$ and $\zeta$, the ranges (over the 10 trials) for the estimated squared approximation error defined by $\|u-v\|^2_{S_{\text{test}}} = \frac{1}{\#S_{\text{test}}} \sum_{(x,y) \in S_{\text{test}}} (u(x) - v(x))^2$, with a sample $S_{\text{test}}$ independent of $S$. We note that the algorithm is robust with respect to noise and yields (with high probability) an approximation error which is below the noise level, and we clearly observe (as expected) that the approximation error decreases with $n$, whatever the noise level. 
		
		\begin{table}[h]
		\centering
		\begin{tabular}{ccc}\toprule
			$n$ & $\zeta^2$ & $\|u-v\|^2_{S_{\text{test}}}$ \\ \midrule
			\multirow{3}{*}{$5 \, 10^2$} & $10^{-2}$ & $[3.40 \, 10^{-2} , 3.09 \, 10^{-1}]$ \\
			& $10^{-4}$ & $[4.29 \, 10^{-4} , 5.93 \, 10^{-3}]$ \\
			& $10^{-6}$ & $[5.23 \, 10^{-7} , 7.44 \, 10^{-1}]$ \\ \midrule
			\multirow{3}{*}{$10^3$} & $10^{-2}$ & $[3.23 \, 10^{-3} , 4.89 \, 10^{-1}]$ \\
			& $10^{-4}$ & $[1.81 \, 10^{-5} , 1.23 \, 10^{-3}]$ \\
			& $10^{-6}$ & $[1.87 \, 10^{-7} , 3.50 \, 10^{-7}]$ \\ \midrule
			\multirow{3}{*}{$10^4$} & $10^{-2}$ & $[1.52 \, 10^{-4} , 8.77 \, 10^{-3}]$ \\
			& $10^{-4}$ & $[1.38 \, 10^{-6} , 1.72 \, 10^{-6}]$ \\
			& $10^{-6}$ & $[1.42 \, 10^{-8} , 1.85 \, 10^{-8}]$ \\
			\bottomrule
		\end{tabular}
		\caption{Results for the function \eqref{eq:sumOfBivariateFunctions} in the noisy case: training sample size $n$, variance of the noise and ranges (over the 10 trials) for the estimated approximation error.}
		\label{tab:sumOfBivariateFunctionsNoisyCase2}
	\end{table}

	\subsection{Function of a sum of bivariate functions}\label{subsec:functionOfASumOfBivariateFunctions}
		We here consider the approximation of the function
		\begin{equation} \tag{\ensuremath{iii}}
		\label{eq:functionOfASumOfBivariateFunctions}
			u(X) = \log(1+(g(X_1,X_2),\hdots,g(X_{d-1},X_d))^2)
		\end{equation}
		where $d = 10$, the $X_i$ are independent and uniform on $[-1,1]$, we consider the noiseless case $\varepsilon = 0$ and the function $g$ is defined in Subsection \ref{subsec:sumOfBivariateFunctions}.  We use approximation spaces $\Hc_\nu =  \Pbb_{10}(\Xc^\nu)$ and the algorithm \ref{alg:rank-tree-adaptive}, with an adaptive selection of the tree $T$ in the family of trees of arity 2, each run starting from two different families of trees $T^1_\sigma = \{\sigma(\alpha) : \alpha \in T^1\}$ and $T^2_\sigma = \{\sigma(\alpha) : \alpha \in T^2\}$, with $\sigma$ a permutation of $D$ and where $T^1$ and $T^2$ are visible in Figure \ref{fig:sumOfBivariateFunctionsTrees}.
				
		Table \ref{tab:functionOfASumOfBivariateFunctions} shows that with high probability, the algorithm yields a very accurate approximation with a small sample size and that the accuracy increases with the sample size. Also, we observe that the algorithm yields the expected optimal tree (as defined in Subsection \ref{subsec:sumOfBivariateFunctions}) with high probability. Decreasing further the error would require an increase of the sample size and of the degree of the polynomial spaces. 
				
		
		\begin{table}[h]
			\centering
			\begin{tabular}{ccccccc}\toprule
				$T_\sigma$ & $n$ & $\hat \Pbb(T \text{ is optimal})$ & $\varepsilon(S_{\text{test}},v)$ & CV error & $C(T,r)$ \\ \midrule
				\multirow{3}{*}{$T^1_\sigma$}& $5 \, 10^2$ & $80\%$ & $[4.19 \, 10^{-3} , 5.57 \, 10^{-2}]$ & $[2.15 \, 10^{-3} , 5.00 \, 10^{-2}]$ & $[219 , 573]$ \\
				& $10^3$ & $100\%$ & $[8.77 \, 10^{-5} , 2.04 \, 10^{-2}]$ & $[8.79 \, 10^{-6} , 1.53 \, 10^{-2}]$ & $[277 , 1417]$ \\
				& $10^4$ & $90\%$ & $[1.11 \, 10^{-5} , 1.99 \, 10^{-2}]$ & $[6.50 \, 10^{-6} , 1.68 \, 10^{-2}]$ & $[277 , 1834]$ \\ \midrule
				\multirow{3}{*}{$T^2_\sigma$}& $5 \, 10^2$ & $70\%$ & $[1.01 \, 10^{-3} , 7.09 \, 10^{-2}]$ & $[1.00 \, 10^{-4} , 5.02 \, 10^{-2}]$ & $[211 , 1289]$ \\
				& $10^3$ & $90\%$ & $[8.34 \, 10^{-5} , 5.29 \, 10^{-2}]$ & $[1.23 \, 10^{-5} , 4.87 \, 10^{-2}]$ & $[277 , 1566]$ \\
				& $10^4$ & $100\%$ & $[1.15 \, 10^{-5} , 1.98 \, 10^{-2}]$ & $[8.65 \, 10^{-6} , 1.64 \, 10^{-2}]$ & $[277 , 1290]$ \\
				\bottomrule
			\end{tabular}
			\caption{Results for the function \eqref{eq:functionOfASumOfBivariateFunctions}: training sample size $n$, estimation of the probability of obtaining an optimal tree and ranges (over the 10 trials) for the test error, the cross-validation (CV) error estimator and the storage complexity.}
			\label{tab:functionOfASumOfBivariateFunctions}
		\end{table}

	\subsection{Another sum of bivariate functions}\label{subsec:anotherSumOfBivariateFunctions}
		We here consider another sum of bivariate functions
		\begin{equation} \tag{\ensuremath{iv}}
		\label{eq:anotherSumOfBivariateFunctions}
		u(X)= g(X_1,X_2) + g(X_2,X_3) + g(X_3,X_4) + \hdots + g(X_{d-1},X_d),
		\end{equation}
		where $d=16$, $\varepsilon = 0$, the $X_i$ are independent and uniform on $[-1,1]$ and the function $g$ is defined in Subsection \ref{subsec:sumOfBivariateFunctions}. We consider $\Hc = \bigotimes_{\nu=1}^d \Pbb_{5}(\Xc^\nu)$ so that $u\in \Hc$ (no discretization errors). For such a function, the linear tree $T^1 = \{\{1\},\hdots,\{d\},\{1,2\},\hdots,\{1,\hdots,d\}\}$ seems to be a natural choice, although it is not obvious that it is the optimal one.
 		The algorithm \ref{alg:rank-tree-adaptive} is run several times starting from $T^1$ or random permutations $T^1_\sigma$ of $T^1$. Table \ref{tab:anotherSumOfBivariateFunctions} shows the obtained results with a training sample size $n = 10^4$, with or without tree adaptation. It first illustrates that, without tree adaptation, choosing a linear tree $T^1$ leads to a recovery of the function whereas choosing a tree $T^1_\sigma$ with $\sigma$ randomly drawn leads to a poor approximation of the function. However, with tree adaptation, the algorithm recovers the function at machine precision, whatever the starting permutation $\sigma$ (over the 10 trials).
 		
\begin{figure}[H]
 			\centering
 			\footnotesize
 			\begin{subfigure}[b]{.49\textwidth}
 				\centering
 				\begin{tikzpicture}[scale=0.35,level distance = 15mm, outer sep = -0.4mm]
 					\tikzstyle{level 1}=[sibling distance=60mm]
 					\tikzstyle{level 2}=[sibling distance=40mm]
 					\tikzstyle{level 3}=[sibling distance=25mm]
 					\tikzstyle{level 4}=[sibling distance=20mm]
 					\node [active,label=above:{}]  {} 
					child{node [active,label=above:{}] {}
					child{node [active,label=above:{}] {}
					child{node [active,label=above:{}] {}
					child{node [active,outer sep = 0.1mm,label=below:{$\{1\}$}] {}}
					child{node [active,outer sep = 0.1mm,label=below:{$\{2\}$}] {}}}
					child{node [active,outer sep = 0.1mm,label=below:{$\{3\}$}] {}}}
					child{node [active,label=above:{}] {}
					child{node [active,outer sep = 0.1mm,label=below:{$\{5\}$}] {}}
					child{node [active,label=above:{}] {}
					child{node [active,label=below left:{$\{6\}$}] {}}
					child{node [active,label=above:{}] {}
					child{node [active,label=below left:{$\{7\}$}] {}}
					child{node [active,label=above:{}] {}
					child{node [active,label=below left:{$\{8\}$}] {}}
					child{node [active,label=above:{}] {}
					child{node [active,label=below left:{$\{9\}$}] {}}
					child{node [active,label=above:{}] {}
					child{node [active,label=below left:{$\{10\}$}] {}}
					child{node [active,label=above:{}] {}
					child{node [active,label=below left:{$\{11\}$}] {}}
					child{node [active,label=above:{}] {}
					child{node [active,label=below left:{$\{12\}$}] {}}
					child{node [active,label=above:{}] {}
					child{node [active,label=below left:{$\{13\}$}] {}}
					child{node [active,label=above:{}] {}
					child{node [active,label=below left:{$\{14\}$}] {}}
					child{node [active,label=above:{}] {}
					child{node [active,outer sep = 0.1mm,label=below:{$\{15\}$}] {}}
					child{node [active,outer sep = 0.1mm,label=below:{$\{16\}$}] {}}}
					}}}}}}}}}}}
					child{node [active,outer sep = 0.1mm,label=below:{$\{4\}$}] {}};
 				\end{tikzpicture}
 			\end{subfigure}
 			\begin{subfigure}[b]{.49\textwidth}
 				\centering
 				\begin{tikzpicture}[scale=0.35,level distance = 15mm, outer sep = -0.4mm]  
 					\tikzstyle{level 1}=[sibling distance=60mm]
 					\tikzstyle{level 2}=[sibling distance=40mm]
 					\tikzstyle{level 3}=[sibling distance=25mm]
 					\tikzstyle{level 4}=[sibling distance=20mm]
					\node [active,label=above:{}]  {} 
					child{node [active,label=above:{}] {}
					child{node [active,label=above:{}] {}
					child{node [active,label=above:{}] {}
					child{node [active,label=above:{}] {}
					child{node [active,label=above:{}] {}
					child{node [active,label=above:{}] {}
					child{node [active,label=above:{}] {}
					child{node [active,label=above:{}] {}
					child{node [active,label=above:{}] {}
					child{node [active,outer sep = 0.1mm,label=below:{$\{16\}$}] {}}
					child{node [active,outer sep = 0.1mm,label=below:{$\{15\}$}] {}}}
					child{node [active,label=below right:{$\{14\}$}] {}}}
					child{node [active,label=below right:{$\{13\}$}] {}}}
					child{node [active,label=below right:{$\{12\}$}] {}}}
					child{node [active,label=below right:{$\{11\}$}] {}}}
					child{node [active,label=below right:{$\{10\}$}] {}}}
					child{node [active,label=below right:{$\{9\}$}] {}}}
					child{node [active,outer sep = 0.1mm,label=below:{$\{8\}$}] {}}}
					child{node [active,label=above:{}] {}
					child{node [active,outer sep = 0.1mm,label=below:{$\{6\}$}] {}}
					child{node [active,label=above:{}] {}
					child{node [active,label=below left:{$\{5\}$}] {}}
					child{node [active,label=above:{}] {}
					child{node [active,label=below left:{$\{4\}$}] {}}
					child{node [active,label=above:{}] {}
					child{node [active,label=below left:{$\{3\}$}] {}}
					child{node [active,label=above:{}] {}
					child{node [active,outer sep = 0.1mm,label=below:{$\{2\}$}] {}}
					child{node [active,outer sep = 0.1mm,outer sep = 0.1mm,label=below:{$\{1\}$}] {}}}}}}}}
					child{node [active,outer sep = 0.1mm,label=below:{$\{7\}$}] {}};
				\end{tikzpicture}
 			\end{subfigure}
 			\caption{Examples of final trees obtained when running Algorithm \ref{alg:rank-tree-adaptive} with a training sample size $n = 10^4$, starting from two different random permutations of $T^1$.}
 			\label{fig:anotherSumOfBivariateFunctions_trees}
 		\end{figure}
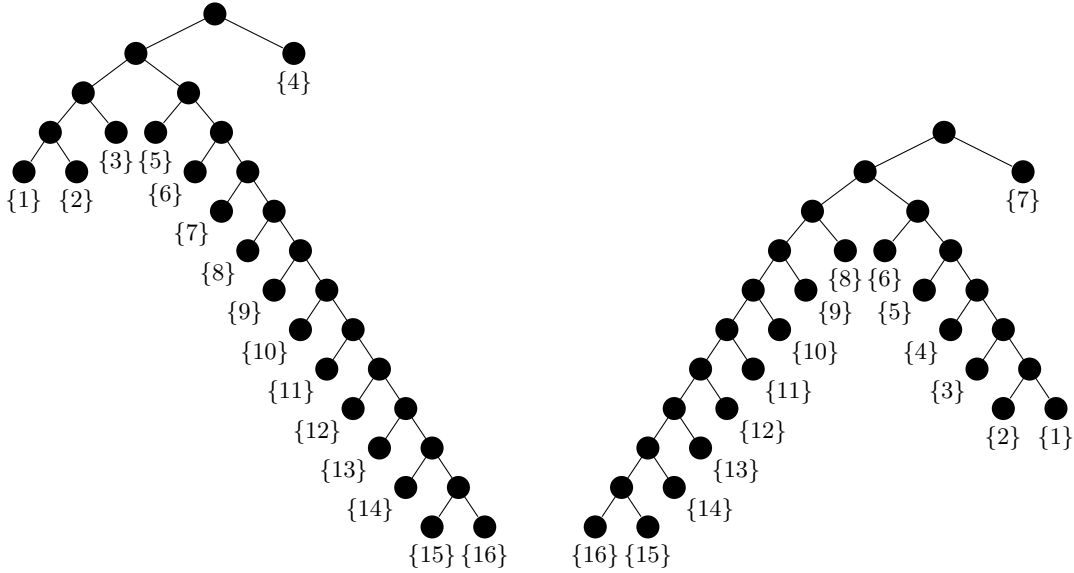

 		Figure \ref{fig:anotherSumOfBivariateFunctions_trees} shows examples of final trees obtained when running Algorithm \ref{alg:rank-tree-adaptive}, starting from two different random permutations of $T^1$. We notice that the algorithm returns non obvious trees, selected in the family of trees of arity 2, whose nodes contain consecutive variables.
	 	
	 	\begin{table}[H]
	 	\centering
		 	\begin{tabular}{cccccc}\toprule
		 		$T_\sigma$ & Tree adaptation & $\varepsilon(S_{\text{test}},v)$ & CV error & $C(T,r)$ \\ \midrule
		 		\multirow{2}{*}{$T^1_{id}$} & false & $[2.28 \, 10^{-15} , 1.98 \, 10^{-14}]$ & $[1.67 \, 10^{-15} , 9.87 \, 10^{-15}]$ & $[1760 , 3131]$ \\
		 		& true & $[3.79 \, 10^{-15} , 2.09 \, 10^{-14}]$ & $[2.30 \, 10^{-15} , 1.26 \, 10^{-14}]$ & $[1800 , 2974]$ \\ \midrule
		 		\multirow{2}{*}{$T^1_{\sigma}$} & false & $[4.48 \, 10^{-3} , 5.06 \, 10^{-3}]$ & $[2.49 \, 10^{-3} , 3.14 \, 10^{-3}]$ & $[4779 , 5490]$ \\
		 		& true & $[4.81 \, 10^{-15} , 3.35 \, 10^{-14}]$ & $[2.88 \, 10^{-15} , 1.59 \, 10^{-14}]$ & $[1791 , 2428]$ \\ \bottomrule
		 	\end{tabular}
		 	\caption{Results for the function \eqref{eq:anotherSumOfBivariateFunctions} for a training sample size $n=10^4$, with or without tree adaptation, and starting from $T^1 = T^1_{id}$ or random permutations $T^1_\sigma$ of $T_1$: ranges (over the 10 trials) for the test error, the cross-validation (CV) error estimator and the storage complexity.}
		 	\label{tab:anotherSumOfBivariateFunctions}
	 	\end{table}
 
	\subsection{Compositions of functions}\label{subsec:compositionOfFunctions}
\begin{figure}[h]
			\centering
			\begin{tikzpicture}[level distance = 10mm]
			\tikzstyle{level 1}=[sibling distance=40mm]
			\tikzstyle{level 2}=[sibling distance=20mm]
			\tikzstyle{level 3}=[sibling distance=10mm]
			\node   {$h$}
			child {node   {$h$}
				child {node  {$h$}
					child {node  {$X_1$}}
					child {node  {$X_2$}}
				}
				child {node  {$h$}
					child {node  {$X_3$}}
					child {node  {$X_4$}}
				}
			}
			child {node  {$h$}
				child {node  {$h$}
					child {node  {$X_5$}}
					child {node  {$X_6$}}
				}
				child {node  {$h$}
					child {node {$X_7$}}
					child {node {$X_8$}}
				}
			};
			\end{tikzpicture}
			\caption{Schematic representation of the function \eqref{eq:compositionOfFunctions}.}\label{fig:compositionOfFunctions}

			\end{figure}
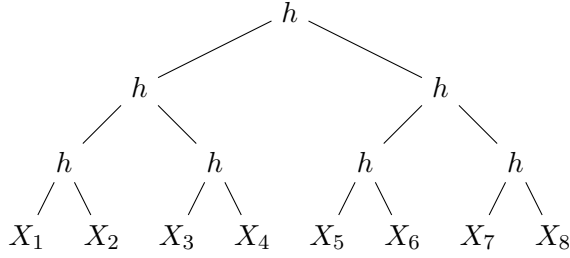

		We consider the approximation of 
		\begin{equation} \tag{\ensuremath{v}}
		\label{eq:compositionOfFunctions}
			u(X) = h(h(h(X_1,X_2),h(X_3,X_4)),h(h(X_5,X_6),h(X_7,X_8))),
		\end{equation}
		where $h$ is a bivariate function and where the $d=8$ random variables $X_1,\dots,X_8$ are independent and uniform on $[-1,1]$. The noise level is set to $\varepsilon = 0$. The function $u$ is obtained by tree-structured compositions of the function $h$, illustrated on Figure \ref{fig:compositionOfFunctions}, where each interior node of the tree corresponds to the application of $h$ to the outputs of its children nodes. 
		
		Here we choose 
		$h(t,s) =  9^{-1}{(2+ts)^2}$. Therefore,  $u$ is a polynomial function of degree $8$. Then, we use approximation spaces $\Hc_\nu =  \Pbb_{8}(\Xc_\nu)$, so that function $u$ belongs to $\Hc$ and could (in principle) be recovered exactly for any choice of tree with a sufficiently high rank.
		
\begin{figure}[h]
			\centering
			\footnotesize
			\begin{subfigure}[b]{.49\textwidth}
				\centering
				\begin{tikzpicture}[level distance = 8mm]
				\tikzstyle{active}=[circle,fill=black]
				\tikzstyle{level 1}=[sibling distance=38mm]
				\tikzstyle{level 2}=[sibling distance=19mm]
				\tikzstyle{level 3}=[sibling distance=11mm]
				\tikzstyle{level 4}=[sibling distance=5mm]
					\node [active,label={above:{}}]  {}
					child {node [active,label=above left:{}]  {}
						child {node [active,label=above left:{}] {}
							child {node [active,label=below:{{${\{1\}}$}}] {}}
							child {node [active,label=below:{{${\{2\}}$}}] {}}
						}
						child {node [active,label=above right:{}] {}
							child {node [active,label=below:{{${\{3\}}$}}] {}}
							child {node [active,label=below:{{${\{4\}}$}}] {}}
						}
					}
					child {node [active,label=above right:{}] {}
						child {node [active,label=above left:{}] {}
							child {node [active,label=below:{{${\{5\}}$}}] {}}
							child {node [active,label=below:{{${\{6\}}$}}] {}}
						}
						child {node [active,label=above right:{}] {}
							child {node [active,label=below:{{${\{7\}}$}}] {}}
							child {node [active,label=below:{{${\{8\}}$}}] {}}
						}
					};
				\end{tikzpicture}
				\caption{Tree $T^1$.}
				\label{composition-T1}
			\end{subfigure}			
			\begin{subfigure}[b]{.49\textwidth}
				\centering
				\begin{tikzpicture}[level distance = 6mm]
				\tikzstyle{level 1}=[sibling distance=11mm]
				\tikzstyle{active}=[circle,fill=black]
				\node [active,label={above:{}}]  {}
				child {node [active,label=below:{${\{8\}}$}]  {}}
						child {node [active,label=above right:{}]  {}
							child {node [active,label=below:{${\{7\}}$}]  {}}
							child {node [active,label=above right:{}]  {}
								child {node [active,label=below:{${\{6\}}$}]  {}}
								child {node [active,label=above right:{}]  {}
									child {node [active,label=below:{${\{5\}}$}]  {}}
									child {node [active,label=above right:{}]  {}
										child {node [active,label=below:{${\{4\}}$}]  {}}
										child {node [active,label=above right:{}]  {}
											child {node [active,label=below:{${\{3\}}$}]  {}}
											child {node [active,label=above right:{}]  {}
												child {node [active,label=below:{${\{2\}}$}]  {}}
												child {node [active,label=below:{${\{1\}}$}]  {}}	
											}
										}
									}
								}
							}
						};
				\end{tikzpicture}
				\caption{Tree $T^2$.}
				\label{composition-T2}
			\end{subfigure}
			 \caption{Two different dimension trees $T^1$ and $T^2$.}
			\label{fig:compositionOfFunctions_trees}
		\end{figure}
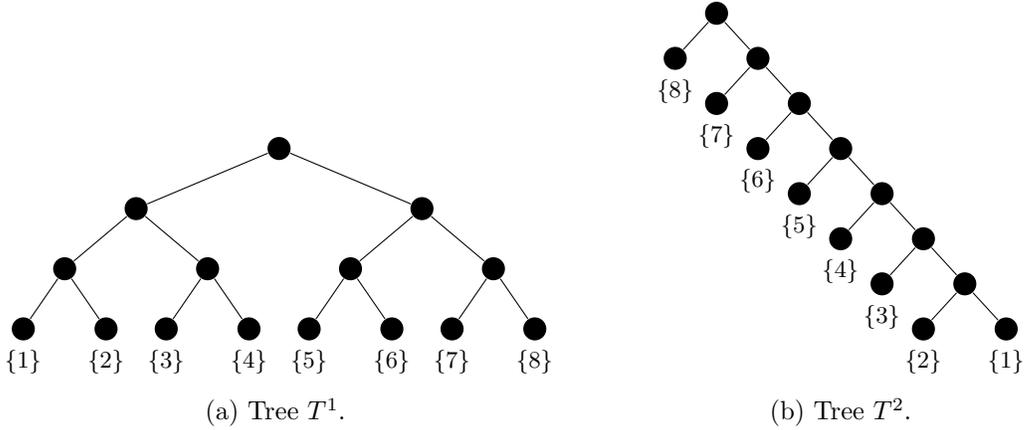

There is a natural dimension tree $T^1$ associated with this function, illustrated on Figure \ref{composition-T1}. Using this tree $T^1$,  the function can be exactly represented in the format $\Tc^{T^1}_{r}(\Hc)$ with a tree-based rank $r = (r_\alpha)_{\alpha \in T^1}$ such that $r_\alpha = 1+2^{\level({\alpha})}$ for $\alpha\neq D$, and a storage complexity of $2427$. Although the function can be exactly represented with any choice of tree, the ranks could be dramatically high for bad choices of tree. For example, when considering the tree $T^1_\sigma$ obtained by applying the permutation $\sigma = (8,1,6,4,7,2,3,5)$ to $T^1$, we obtain a representation with ranks more than $1000$ (at level $1$) and a storage complexity greater than $9.10^6$ for a representation with relative precision $10^{-14}$. 

For the application of Algorithm \ref{alg:rank-tree-adaptive} with tree adaptation, we start from trees belonging to two families $T^1_\sigma$ and $T^2_\sigma$, respectively obtained by permutations of the trees $T^1$ and $T^2$ shown on Figures \ref{composition-T1} and \ref{composition-T2}. Each run of the algorithm starts with a random permutation $\sigma$. Recall that different trees $T_\sigma$ may yield the same tree-based format, if they coincide as elements of $2^{2^D}$. For example, for $T = T^1_\sigma$ with $\sigma = (7,8,6,5,4,3,1,2)$, $T = T^1$. Then, we will say the algorithm finds an optimal tree if it yields a permutation such that $T = T^1$ as elements of $2^{2^D}$.

Table \ref{tab:compositionOfFunctions} summarizes the obtained results.
We observe that the algorithm is able to recover with high probability an optimal tree, even when starting from the tree $T^2_\sigma$ that does not coincide with the natural tree structure of the function $u$.
	
\begin{figure}
		\footnotesize
			\centering
			\begin{tikzpicture}[level distance = 8mm]
			\tikzstyle{active}=[circle,fill=black]
			\tikzstyle{level 1}=[sibling distance=38mm]
			\tikzstyle{level 2}=[sibling distance=19mm]
			\tikzstyle{level 3}=[sibling distance=11mm]
			\tikzstyle{level 4}=[sibling distance=5mm]
			\node [active,label={above:{$1$}}]  {}
			child {node [active,label=above left:{$3$}]  {}
				child {node [active,label=above left:{$5$}] {}
					child {node [active,label=below:{{$\stackon{\{1\}}{9}$}}] {}}
					child {node [active,label=below:{{$\stackon{\{2\}}{9}$}}] {}}
				}
				child {node [active,label=above right:{$5$}] {}
					child {node [active,label=below:{{$\stackon{\{3\}}{9}$}}] {}}
					child {node [active,label=below:{{$\stackon{\{4\}}{9}$}}] {}}
				}
			}
			child {node [active,label=above right:{$3$}] {}
				child {node [active,label=above left:{$5$}] {}
					child {node [active,label=below:{{$\stackon{\{5\}}{9}$}}] {}}
					child {node [active,label=below:{{$\stackon{\{6\}}{9}$}}] {}}
				}
				child {node [active,label=above right:{$5$}] {}
					child {node [active,label=below:{{$\stackon{\{7\}}{9}$}}] {}}
					child {node [active,label=below:{{$\stackon{\{8\}}{9}$}}] {}}
				}
			};
			\end{tikzpicture}
			\caption{$\alpha$-ranks and dimensions associated with leaf nodes of an approximation of the function \eqref{eq:compositionOfFunctions} obtained using Algorithm \ref{alg:rank-tree-adaptive} with $n=10^5$, starting from a tree $T^2_\sigma$. The obtained tree is $T^1$.}
			\label{fig:compositionOfFunctionsRanks}
		\end{figure}
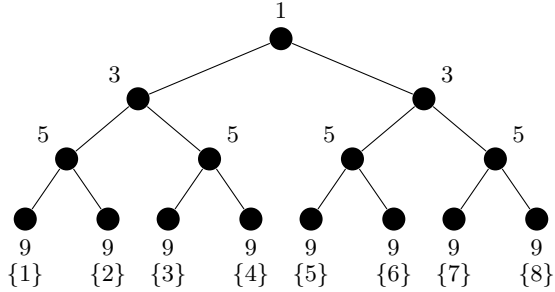

With high probability, the algorithm yields an approximation with very low error, even with a small sample size $n$. With a training sample large enough, the algorithm is able to recover the function $u$ at machine precision. Figure \ref{fig:compositionOfFunctionsRanks} shows the $\alpha$-ranks of an approximation computed by the algorithm with $n=10^5$ with a generalization error at machine precision.

		\begin{table}
			\centering

			\begin{tabular}{cccccc}\toprule
				$T_\sigma$ & $n$ & $\hat \Pbb(T = T^1)$ & $\varepsilon(S_{\text{test}},v)$ & CV error & $C(T,r)$ \\ \midrule
				\multirow{3}{*}{$T^1_\sigma$}& $10^3$ & $90\%$ & $[1.55 \, 10^{-5} , 1.32 \, 10^{-4}]$ & $[8.46 \, 10^{-7} , 3.38 \, 10^{-5}]$ & $[529 , 1121]$\\
				& $10^4$ & $100\%$ & $[1.04 \, 10^{-8} , 6.80 \, 10^{-6}]$ & $[4.34 \, 10^{-11} , 4.91 \, 10^{-6}]$ & $[593 , 2688]$\\
				& $10^5$ & $100\%$ & $[3.29 \, 10^{-15} , 1.80 \, 10^{-4}]$ & $[1.74 \, 10^{-15} , 1.96 \, 10^{-4}]$ & $[342 , 2800]$\\ 
				\midrule
				\multirow{3}{*}{$T^2_\sigma$}& $10^3$ & $90\%$ & $[1.75 \, 10^{-5} , 1.75 \, 10^{-4}]$ & $[1.01 \, 10^{-6} , 8.71 \, 10^{-5}]$ & $[360 , 1062]$ \\
				& $10^4$ & $90\%$ & $[2.15 \, 10^{-8} , 4.10 \, 10^{-3}]$ & $[1.21 \, 10^{-9} , 4.26 \, 10^{-3}]$ & $[185 , 2741]$\\
				& $10^5$ & $100\%$ & $[4.67 \, 10^{-15} , 8.92 \, 10^{-3}]$ & $[2.29 \, 10^{-15} , 6.83 \, 10^{-3}]$ & $[163 , 2594]$ \\ 
				\bottomrule
			\end{tabular}
			\caption{Results for the function \eqref{eq:compositionOfFunctions}: training sample size $n$, estimation of the probability of obtaining $T^1$ or $T^2$ and ranges (over the 10 trials) for the test error, the cross-validation (CV) error estimator and the storage complexity.}
			\label{tab:compositionOfFunctions}
		\end{table}
	
		\paragraph{Illustration of the behavior of Algorithm \ref{alg:rank-tree-adaptive}.}
			Table \ref{tab:adaptationCompositionsOfFunctions} illustrates the behavior of Algorithm \ref{alg:rank-tree-adaptive} when using a training sample of size $n = 10^5$ and starting from a tree $T^2_\sigma$ shown in Figure \ref{fig:poggio1}. The adapted trees at each iteration $m$ are displayed in Figure \ref{fig:compositionsOfFunctions}. We observe that the algorithm recovers the function with a high accuracy after 25 iterations and 7 adaptations of the tree.

		\begin{table}[H]
			\centering
			\begin{tabular}{ccccc}\toprule
				$m$ & Tree $T$ & Tree-based rank $r^m$ & $\varepsilon(S_{\text{test}},v)$ & $C(T,r^m)$ \\ \midrule
				$1$ & Fig. \ref{fig:poggio1} & $(1, 1, 1, 1, 1, 1, 1, 1, 1, 1, 1, 1, 1, 1, 1)$ & $3.38 \, 10^{-2}$ & $79$ \\ \midrule
				$2$ & \multirow{2}{*}{Fig. \ref{fig:poggio2}} & $(1, 1, 2, 1, 1, 1, 1, 1, 1, 1, 1, 1, 2, 1, 1)$ & $2.95 \, 10^{-2}$ & $100$ \\ 
				$3$ & & $(1, 1, 2, 1, 1, 1, 1, 1, 1, 1, 1, 1, 2, 1, 1)$ & $2.95 \, 10^{-2}$ & $100$ \\ \midrule
				$4$ & \multirow{2}{*}{Fig. \ref{fig:poggio3}} & $(1, 1, 2, 1, 2, 1, 1, 1, 2, 1, 1, 1, 2, 1, 1)$ & $2.45 \, 10^{-2}$ & $121$ \\ 
				$5$ & & $(1, 1, 2, 1, 2, 1, 1, 1, 2, 1, 1, 1, 2, 1, 1)$ & $2.45 \, 10^{-2}$ & $121$ \\ \midrule
				$6$ & \multirow{3}{*}{Fig. \ref{fig:poggio4}} & $(1, 1, 2, 1, 2, 1, 1, 1, 2, 1, 2, 1, 2, 2, 1)$ & $1.85 \, 10^{-2}$ & $142$ \\ 
				$7$ & & $(1, 1, 2, 1, 2, 1, 1, 1, 2, 1, 2, 1, 2, 2, 1)$ & $1.85 \, 10^{-2}$ & $142$ \\ 
				$8$ & & $(1, 1, 2, 1, 2, 1, 2, 1, 2, 1, 2, 1, 2, 2, 2)$ & $8.97 \, 10^{-3}$ & $163$ \\ \midrule
				$9$ & \multirow{2}{*}{Fig. \ref{fig:poggio5}} & $(1, 2, 2, 2, 2, 2, 2, 2, 2, 1, 2, 2, 2, 2, 2)$ & $9.54 \, 10^{-3}$ & $188$ \\ 
				$10$ & & $(1, 2, 2, 2, 2, 2, 2, 2, 2, 1, 2, 2, 2, 2, 2)$ & $8.89 \, 10^{-3}$ & $188$ \\ \midrule
				$11$ & \multirow{2}{*}{Fig. \ref{fig:poggio6}} & $(1, 2, 2, 2, 2, 2, 2, 2, 2, 1, 2, 2, 2, 2, 2)$ & $9.47 \, 10^{-3}$ & $188$ \\ 
				$12$ & & $(1, 2, 2, 2, 2, 2, 2, 2, 2, 1, 2, 2, 2, 2, 2)$ & $8.87 \, 10^{-3}$ & $188$ \\ \midrule
				$13$ & \multirow{3}{*}{Fig. \ref{fig:poggio7}} & $(1, 2, 2, 2, 2, 2, 2, 2, 2, 1, 2, 2, 2, 2, 2)$ & $5.22 \, 10^{-3}$ & $188$ \\ 
				$14$ & & $(1, 2, 2, 2, 2, 2, 2, 2, 2, 1, 2, 2, 2, 2, 2)$ & $3.97 \, 10^{-3}$ & $188$ \\ 
				$15$ & & $(1, 2, 3, 2, 3, 2, 3, 2, 3, 2, 3, 2, 3, 3, 3)$ & $1.55 \, 10^{-4}$ & $308$ \\ \midrule
				$16$ & \multirow{10}{*}{Fig. \ref{fig:poggio8}} & $(1, 3, 3, 3, 3, 2, 3, 3, 3, 2, 3, 3, 3, 3, 3)$ & $1.18 \, 10^{-4}$ & $364$ \\ 
				$17$ & & $(1, 3, 3, 3, 3, 2, 3, 3, 3, 2, 3, 3, 3, 3, 3)$ & $1.18 \, 10^{-4}$ & $364$ \\ 
				$18$ & & $(1, 3, 4, 3, 4, 2, 4, 3, 4, 2, 4, 3, 4, 4, 4)$ & $6.65 \, 10^{-6}$ & $520$ \\ 
				$19$ & & $(1, 3, 5, 3, 5, 3, 5, 3, 5, 3, 5, 3, 5, 5, 5)$ & $1.19 \, 10^{-6}$ & $723$ \\ 
				$20$ & & $(1, 4, 5, 4, 5, 3, 5, 4, 5, 3, 5, 4, 5, 5, 5)$ & $1.72 \, 10^{-7}$ & $865$ \\ 
				$21$ & & $(1, 4, 6, 4, 6, 3, 6, 4, 6, 3, 6, 4, 6, 6, 6)$ & $1.47 \, 10^{-8}$ & $1113$ \\ 
				$22$ & & $(1, 5, 6, 5, 6, 3, 6, 5, 6, 3, 6, 5, 6, 6, 6)$ & $7.02 \, 10^{-9}$ & $1311$ \\ 
				$23$ & & $(1, 5, 7, 5, 7, 3, 7, 5, 7, 3, 7, 5, 7, 7, 7)$ & $1.27 \, 10^{-10}$ & $1643$ \\ 
				$24$ & & $(1, 5, 8, 5, 8, 3, 8, 5, 8, 3, 8, 5, 8, 8, 8)$ & $3.87 \, 10^{-12}$ & $2015$ \\ 
				$25$ & & $(1, 5, 9, 5, 9, 3, 9, 5, 9, 3, 9, 5, 9, 9, 9)$ & $2.95 \, 10^{-14}$ & $2427$ \\
				\bottomrule
			\end{tabular}
			\caption{Behavior of Algorithm \ref{alg:rank-tree-adaptive} for the approximation of the function \eqref{eq:compositionOfFunctions}, with $n = 10^5$ and an initial dimension tree shown in Figure \ref{fig:poggio1}.}
			\label{tab:adaptationCompositionsOfFunctions}
		\end{table}

		\begin{figure}[H]
		\footnotesize
				\centering
				\begin{subfigure}[b]{.49\textwidth}
					\centering
					\begin{tikzpicture}[scale=0.3,level distance = 12mm, inner sep=1.1mm]  \tikzstyle{level 1}=[sibling distance=30mm]  \tikzstyle{level 2}=[sibling distance=30mm]  \tikzstyle{level 3}=[sibling distance=30mm]  \tikzstyle{level 4}=[sibling distance=30mm]  \tikzstyle{level 5}=[sibling distance=30mm]  \tikzstyle{level 6}=[sibling distance=30mm]  \tikzstyle{level 7}=[sibling distance=30mm]  \node [active,label=above:{$1$}]  {} child{node [active,label=above:{$2$}] {}child{node [active,label=above:{$4$}] {}child{node [active,label=above:{$6$}] {}child{node [active,label=above:{$8$}] {}child{node [active,label=above:{$10$}] {}child{node [active,label=above:{$12$}] {}child{node [active,label=below:{$14$}] {}}child{node [active,label=below:{$15$}] {}}}child{node [active,label=below:{$13$}] {}}}child{node [active,label=below:{$11$}] {}}}child{node [active,label=below:{$9$}] {}}}child{node [active,label=below:{$7$}] {}}}child{node [active,label=below:{$5$}] {}}}child{node [active,label=below:{$3$}] {}};\end{tikzpicture}
					\caption{$m = 1$.}
					\label{fig:poggio1}
				\end{subfigure}
				\begin{subfigure}[b]{.49\textwidth}
					\centering
					\begin{tikzpicture}[scale=0.4,level distance = 12mm, inner sep=1.1mm]  \tikzstyle{level 1}=[sibling distance=50mm]  \tikzstyle{level 2}=[sibling distance=25mm]  \tikzstyle{level 3}=[sibling distance=30mm]  \tikzstyle{level 4}=[sibling distance=35mm]  \tikzstyle{level 5}=[sibling distance=20mm]  \node [active,label=above:{$1$}]  {} child{node [active,label=above:{$2$}] {}child{node [active,label=above:{$6$}] {}child{node [active,label=above:{$10$}] {}child{node [active,label=above:{$4$}] {}child{node [active,label=below:{$3$}] {}}child{node [active,label=below:{$13$}] {}}}child{node [active,label=above:{$12$}] {}child{node [active,label=below:{$5$}] {}}child{node [active,label=below:{$15$}] {}}}}child{node [active,label=below:{$14$}] {}}}child{node [active,label=below:{$7$}] {}}}child{node [active,label=above:{$8$}] {}child{node [active,label=below:{$9$}] {}}child{node [active,label=below:{$11$}] {}}};\end{tikzpicture}
					\caption{$m = 2,3$.}
					\label{fig:poggio2}
				\end{subfigure}
				\begin{subfigure}[b]{.49\textwidth}
					\centering
					\begin{tikzpicture}[scale=0.3,level distance = 15mm, inner sep=1.1mm]  
\tikzstyle{level 1}=[sibling distance=95mm]  
\tikzstyle{level 2}=[sibling distance=60mm]  
\tikzstyle{level 3}=[sibling distance=35mm]  
\tikzstyle{level 4}=[sibling distance=20mm]
\node [active,label=above:{$1$}]  {} child{node [active,label=above:{$6$}] {}child{node [active,label=above:{$2$}] {}child{node [active,label=below:{$7$}] {}}child{node [active,label=below:{$11$}] {}}}child{node [active,label=above:{$10$}] {}child{node [active,label=above:{$4$}] {}child{node [active,label=below:{$3$}] {}}child{node [active,label=below:{$13$}] {}}}child{node [active,label=above:{$12$}] {}child{node [active,label=below:{$14$}] {}}child{node [active,label=below:{$15$}] {}}}}}child{node [active,label=above:{$8$}] {}child{node [active,label=below:{$5$}] {}}child{node [active,label=below:{$9$}] {}}};\end{tikzpicture}
					\caption{$m = 4,5$.}
					\label{fig:poggio3}
				\end{subfigure}
				\begin{subfigure}[b]{.49\textwidth}
					\centering
					\begin{tikzpicture}[scale=0.3,level distance = 15mm, inner sep=1.1mm]  
\tikzstyle{level 1}=[sibling distance=95mm]  
\tikzstyle{level 2}=[sibling distance=60mm]  
\tikzstyle{level 3}=[sibling distance=35mm]  
\tikzstyle{level 4}=[sibling distance=20mm]
\node [active,label=above:{$1$}]  {} child{node [active,label=above:{$6$}] {}child{node [active,label=above:{$2$}] {}child{node [active,label=below:{$11$}] {}}child{node [active,label=below:{$14$}] {}}}child{node [active,label=above:{$10$}] {}child{node [active,label=above:{$4$}] {}child{node [active,label=below:{$3$}] {}}child{node [active,label=below:{$13$}] {}}}child{node [active,label=above:{$12$}] {}child{node [active,label=below:{$7$}] {}}child{node [active,label=below:{$15$}] {}}}}}child{node [active,label=above:{$8$}] {}child{node [active,label=below:{$5$}] {}}child{node [active,label=below:{$9$}] {}}};\end{tikzpicture}
					\caption{$m = 6,7,8$.}
					\label{fig:poggio4}
				\end{subfigure}
				\begin{subfigure}[b]{.49\textwidth}
					\centering
					\begin{tikzpicture}[scale=0.3,level distance = 15mm, inner sep=1.1mm]  
\tikzstyle{level 1}=[sibling distance=95mm]  
\tikzstyle{level 2}=[sibling distance=60mm]  
\tikzstyle{level 3}=[sibling distance=35mm]  
\tikzstyle{level 4}=[sibling distance=20mm] 
\node [active,label=above:{$1$}]  {} child{node [active,label=above:{$6$}] {}child{node [active,label=above:{$10$}] {}child{node [active,label=above:{$2$}] {}child{node [active,label=below:{$11$}] {}}child{node [active,label=below:{$14$}] {}}}child{node [active,label=above:{$4$}] {}child{node [active,label=below:{$3$}] {}}child{node [active,label=below:{$13$}] {}}}}child{node [active,label=above:{$12$}] {}child{node [active,label=below:{$7$}] {}}child{node [active,label=below:{$15$}] {}}}}child{node [active,label=above:{$8$}] {}child{node [active,label=below:{$5$}] {}}child{node [active,label=below:{$9$}] {}}};\end{tikzpicture}
					\caption{$m = 9,10$.}
					\label{fig:poggio5}
				\end{subfigure}
				\begin{subfigure}[b]{.49\textwidth}
					\centering
					\begin{tikzpicture}[scale=0.3,level distance = 15mm, inner sep=1.1mm]  
\tikzstyle{level 1}=[sibling distance=95mm]  
\tikzstyle{level 2}=[sibling distance=60mm]  
\tikzstyle{level 3}=[sibling distance=35mm]  
\tikzstyle{level 4}=[sibling distance=20mm]  
\node [active,label=above:{$1$}]  {} child{node [active,label=above:{$6$}] {}child{node [active,label=above:{$2$}] {}child{node [active,label=below:{$11$}] {}}child{node [active,label=below:{$14$}] {}}}child{node [active,label=above:{$10$}] {}child{node [active,label=above:{$4$}] {}child{node [active,label=below:{$3$}] {}}child{node [active,label=below:{$13$}] {}}}child{node [active,label=above:{$12$}] {}child{node [active,label=below:{$7$}] {}}child{node [active,label=below:{$15$}] {}}}}}child{node [active,label=above:{$8$}] {}child{node [active,label=below:{$5$}] {}}child{node [active,label=below:{$9$}] {}}};\end{tikzpicture}
					\caption{$m = 11,12$.}
					\label{fig:poggio6}
				\end{subfigure}
				\begin{subfigure}[b]{.49\textwidth}
					\centering
					\begin{tikzpicture}[scale=0.3,level distance = 15mm, inner sep=1.1mm]  
\tikzstyle{level 1}=[sibling distance=95mm]  
\tikzstyle{level 2}=[sibling distance=60mm]  
\tikzstyle{level 3}=[sibling distance=35mm]  
\tikzstyle{level 4}=[sibling distance=20mm]  
\node [active,label=above:{$1$}]  {} child{node [active,label=above:{$6$}] {}child{node [active,label=above:{$4$}] {}child{node [active,label=below:{$3$}] {}}child{node [active,label=below:{$13$}] {}}}child{node [active,label=above:{$10$}] {}child{node [active,label=above:{$2$}] {}child{node [active,label=below:{$11$}] {}}child{node [active,label=below:{$14$}] {}}}child{node [active,label=above:{$12$}] {}child{node [active,label=below:{$7$}] {}}child{node [active,label=below:{$15$}] {}}}}}child{node [active,label=above:{$8$}] {}child{node [active,label=below:{$5$}] {}}child{node [active,label=below:{$9$}] {}}};\end{tikzpicture}
					\caption{$m = 13,14,15$.}
					\label{fig:poggio7}
				\end{subfigure}
				\begin{subfigure}[b]{.49\textwidth}
					\centering
					\begin{tikzpicture}[scale=0.3,level distance = 15mm, inner sep=1.1mm]  
\tikzstyle{level 1}=[sibling distance=90mm]  
\tikzstyle{level 2}=[sibling distance=55mm]  
\tikzstyle{level 3}=[sibling distance=20mm]  
\node [active,label=above:{$1$}]  {} child{node [active,label=above:{$6$}] {}child{node [active,label=above:{$4$}] {}child{node [active,label=below:{$3$}] {}}child{node [active,label=below:{$13$}] {}}}child{node [active,label=above:{$8$}] {}child{node [active,label=below:{$5$}] {}}child{node [active,label=below:{$9$}] {}}}}child{node [active,label=above:{$10$}] {}child{node [active,label=above:{$2$}] {}child{node [active,label=below:{$11$}] {}}child{node [active,label=below:{$14$}] {}}}child{node [active,label=above:{$12$}] {}child{node [active,label=below:{$7$}] {}}child{node [active,label=below:{$15$}] {}}}};\end{tikzpicture}
					\caption{$m = 16,\hdots,25$.}
					\label{fig:poggio8}
				\end{subfigure}
				\caption{Dimension trees associated to the iteration number $m$ in Table \ref{tab:adaptationCompositionsOfFunctions}, with each node numbered. The singletons $\{1\},\{2\},\{3\},\{4\},\{5\},\{6\},\{7\},\{8\}$ correspond to the leaf nodes numbered $9,5,3,13,11,14,7,15$ respectively.}
				\label{fig:compositionsOfFunctions}
		\end{figure}

\section{Conclusion}
	We have proposed algorithms for learning high-dimensional functions using model classes of functions in tree-based tensor formats, with an adaptive selection of trees and corresponding tree-based ranks. The selection of an optimal tree-based rank is a combinatorial problem. The proposed rank adaptation strategy provides approximations that outperform approximations with uniform tree-based ranks. The selection of an optimal tree is also a combinatorial problem. The proposed stochastic algorithm for tree adaptation is able to explore the set of possible trees with given arity and to recover with high probability (at least in our numerical experiments) the optimal tree for the representation of a given function.
	
	
	A larger probability of recovering the optimal tree could be obtained by running algorithm \ref{alg:rank-tree-adaptive} $M$ times, each starting from a different tree (of same arity), and by retaining the approximation giving the best result: if the probability of obtaining the optimal tree out of one trial is $p$, the probability of obtaining the optimal tree out of $M$ trials by selecting the best approximation among them is $1-(1-p)^M$.
	
	The probability to find an optimal tree could be further improved by proposing better probability distributions over the set of possible trees, possibly based on other definitions of the complexity. Also, a more general algorithm for tree adaptation should allow modifications of the arity of the tree. In particular, it could make possible the transition from a binary tree to a trivial tree, and by exploiting sparsity in tensor representations, it could create a bridge between sparse and low-rank tensor approximations.

	As expected, the quality of the approximation improves with the training sample size $n$. However, computational costs increase with $n$. In the case of large data sets, variants of the proposed algorithm using subsamples of the training sample could be proposed, in the spirit of stochastic gradient methods.
	
	Finally, the proposed adaptive algorithms provide numerous approximations, each associated with different trees and different tree-based ranks. 
	This calls for a robust approach for model selection or aggregation that does not rely on statistical estimations of the generalization error, either using an independent test sample (not used to train the model) or a cross-validation estimator (which may be a bad estimator if $n$ is small and/or if the observations are noisy).

\subsection*{Acknowledgements}
The authors acknowledge the French National Research Agency for its financial support within the project ANR CHORUS MONU-0005 and 
the Joint Laboratory of Marine Technology between Naval Group and Centrale Nantes for the financial support with the project Eval PI.
\bibliographystyle{plain}

\end{document}